\documentclass[sn-basic]{sn-jnl}%

\usepackage{graphicx}%
\usepackage{multirow}%
\usepackage{amsmath,amssymb,amsfonts}%
\usepackage{amsthm}%
\usepackage{mathrsfs}%
\usepackage[title]{appendix}%
\usepackage{xcolor}%
\usepackage{textcomp}%
\usepackage{manyfoot}%
\usepackage{booktabs}%
\usepackage{algorithm}%
\usepackage{algorithmicx}%
\usepackage{algpseudocode}%
\usepackage{listings}%
\theoremstyle{thmstyleone}%
\newtheorem{theorem}{Theorem}%
\newtheorem{corollary}{Corollary}%
\newtheorem{assumption}{Assumption}%

\theoremstyle{thmstyletwo}%

\theoremstyle{thmstylethree}%
\newtheorem{definition}{Definition}%

\usepackage{url}            %
\usepackage{nicefrac}       %
\usepackage[dvipsnames]{xcolor}         %

\usepackage{tikz}
\usepackage[nofill,markings]{hf-tikz}
\usetikzlibrary{arrows.meta,positioning,calc}
\usepackage{bm}
\usetikzlibrary{automata,arrows}
\usepackage{subcaption}
\usepackage{wrapfig}
\usepackage{cutwin}
\usepackage{mathtools}
\usepackage{cleveref}
\usepackage{enumitem}
\usepackage{thmtools, thm-restate}

\usepackage[most]{tcolorbox}  %

\def\Zpv{Z^+}
\def\Zps{Z^+}
\def\Zmv{Z^-}
\def\Zms{Z^-}
\def\Xv{X}
\def\Xs{X}  
\def\Yv{Y}
  
\def\Zv{Z}
\def\Zs{Z}  
\def\Sv{S}
 
\def\gG{{\mathcal{G}}}
\newcommand{\Indep}{\mathop{\perp\!\!\!\perp}\nolimits}

\tikzset{
  pfeil/.style={thick,stealth-},
  beschr/.style={remember picture,overlay,font=\small}}

\begin{document}

\title{Regression-Based Estimation of Causal Effects in the Presence of Selection Bias and Confounding}

\author[1,2]{\fnm{Marlies} \sur{Hafer} 
}
    \email{marlies.hafer@tu-dortmund.de}

\author*[1,2]{\fnm{Alexander} \sur{Marx} 
}
\email{alexander.marx@tu-dortmund.de}

\affil[1]{\orgname{TU Dortmund University}, \orgdiv{Department of Statistics}, \city{Dortmund}, \country{Germany}}

\affil[2]{\orgname{Research Center Trustworthy Data Science and Security, University Alliance Ruhr}, \city{Dortmund}, \state{North Rhine-Westphalia}, \country{Germany}}

\abstract{We consider the problem of estimating the expected causal effect $E[\Yv|do(\Xv)]$ for a target variable $\Yv$ when treatment $\Xv$ is set by intervention, focusing on continuous random variables. In settings without selection bias or confounding, $E[\Yv|do(\Xv)] = E[\Yv|\Xv]$, which can be estimated using standard regression methods. However, regression fails when systematic missingness induced by selection bias, or confounding distorts the data. Proxy variables unaffected by the selection process can, under certain constraints, be used to correct for selection bias to recover $E[\Yv|\Xv]$, and hence $E[\Yv|do(\Xv)]$, reliably. When data is additionally affected by confounding, recovering the causal effect from selection-biased data is more challenging and requires access to proxies to both correct for confounding and for the selection mechanism. Assuming access to such proxies from external unbiased observational data, we derive theoretical conditions ensuring identifiability and recoverability of causal effects. We further introduce a linear two-step regression estimator (TSR), which can be extended through adding non-linear basis functions, capable of exploiting proxy variables to adjust for selection bias while accounting for confounding. We show that TSR is consistent with previous estimators when confounding is absent, but achieves a lower variance. Extensive simulation studies validate TSR’s correctness for scenarios that include both selection bias and confounding with proxy variables.}

\keywords{Causality, Causal Inference, Selection Bias, Regression
\footnotetext[0]{\tiny \emph{The underlying manuscript was accepted for publication after peer review.
This author-updated version contains an
additional post-acceptance paragraph in Appendix A.2. The Version of Record is
available online at: \href{https://doi.org/10.1007/s10994-026-07054-6}{https://doi.org/10.1007/s10994-026-07054-6}.}}
}

\clearpage
\pagenumbering{arabic} %
\setcounter{page}{1} 

\addtocontents{toc}{\protect\setcounter{tocdepth}{-1}}

\maketitle

\section{Introduction}
\label{sec:introduction}
Recovering causal effects under selection bias is a fundamental challenge in empirical research. Specifically, we aim to estimate $E[\Yv \mid do(\Xv)]$, the causal effect of setting a continuous treatment $\Xv$ by intervention, denoted by Pearl's do-operator~\citep{pearl:09:causality}, on a continuous target variable $\Yv$ from observational data that may be affected by selection mechanisms and confounding. Selection bias arises when the observed data fails to accurately represent the population due to preferential exclusion or conditioning on a collider, while confounding distorts the true causal relationships through (unobserved) common causes. Both phenomena are pervasive in real-world datasets and, if left unadjusted, can give rise to misleading conclusions.

Selection bias is a critical challenge in many real-world domains, including medicine~\citep{berkson:46:paradox}, economics, and machine learning, with recent examples highlighting its role in COVID-19 research~\citep{herbert:20:collider-bias-covid}, %
cancer progression modeling~\citep{schill:24:bias-cancer-progression}, and fairness~\citep{wang:21:fairness-selection}. %
As a running example, consider that in loan risk assessment, banks may wish to isolate the causal effect of income ($\Xv$) on loan default ($\Yv$) from other risk factors. Naturally, such a dataset only includes cases where loans have been issued (\( \Sv = 1 \)), introducing \emph{selection bias}, as illustrated in \Cref{fig:datasetsandrunning}, implying that the conditional expectation in the selected sample is biased such that $E[Y|X] \neq E[Y|X,S=1]$. Furthermore, unobserved factors like financial literacy of an individual may act as confounders that simultaneously influence income and loan default rates inducing a \emph{confounding bias} implying that the causal effect $E[Y|do(X)]$ differs from $E[Y|X]$. Without proper adjustment for these biases, estimates of risk factors will be unreliable or even contradictory.

While \emph{identifiability} of causal effects 
under confounding has been extensively studied, e.g., through Pearl's do-calculus~\citep{pearl:09:causality}, or %
instrumental variable methods~\citep{imbens:94:iv}, 
\emph{recoverability} from selection-biased data has received comparatively less attention. Standard approaches aim to transform the data, reducing differences between selected and unobserved data conditioning on the propensity score~\citep{austin:11:PSreducingConf,rosenbaum:83:PSinobs}. In contrast, we focus on exploiting unbiased observational data for a subset of variables (proxies and treatment) to identify and recover causal effects. %
Building upon Pearl's do calculus~\citep{pearl:09:causality}, various identification results to ensure recoverability of causal effects from selection-biased data (s-recoverability) have been explored~\citep{pearl:12:solution,correa:18:GeneralizedAdj,jung:24:UnifiedCA,mohan:21:GraphicalModels}.
Further, \citet{boeken:23:privileged-info} emphasize the importance of proposing practical estimators alongside identification results. They introduced regression-based methods to recover the conditional expectation $E[\Yv|\Xv]$ for continuous targets assuming access to a proxy $\Zv$ for the selection variable, which renders the target $\Yv$ independent of the selection variable $\Sv$ when conditioning on $\{\Xv,\Zv\}$, i.e., $\Yv \Indep \Sv \mid \{ \Xv, \Zv \}$. Akin to other works on recoverability~\citep{correa:18:GeneralizedAdj}, they assume access to external data for $\Xv, \Zv$ unaffected by selection %
(cf.~\Cref{fig:datasetsandrunning}). In this work, our goal is to derive a practical two-step estimator for continuous data that %
recovers causal effects from selection biased data. Thereby, we extend previous identification results and analyze the bias and variance of the proposed estimator. In the most general case, our results cover the graph shown in \Cref{fig:datasetsandrunning}, in which we can recover the causal effect of income on loan default by using the covariates ``spendings'', etc.~as a proxy for the selection variable. To correct for the unobserved confounder ``financial literacy'' in our two-step estimator, we leverage the information about the job type of an individual. In summary, our contributions are the following:
\begin{itemize}
    \item Based on Assumption~\ref{ass:new}, we show that for continuous random variables $E[Y|do(X)]$ is identifiable and s-recoverable (Theorem~\ref{theorem:neutheoretisch}) given access to external data for $X$ and proxy $Z$ unaffected by the selection mechanism in \Cref{sec:theory}.
    \item From this theoretical result, we derive a linear Two-Step Regression estimator (TSR) that empirically recovers the causal-effect under Gaussianity in \Cref{sec:our-estimator}.
    \item In \Cref{sec:biasvariance}, we analyze the bias and variance of TSR under linearity for the case in which confounding is absent, noting that TSR is more efficient than repeated regression discussed in prior work.
    \item We confirm those results, as well as the admissibility and usability of our estimator considering ordinary least squares (OLS) and ridge estimation in simulation studies and evaluate violations of our assumptions in \Cref{sec:experiments}. We further provide an application based on semi-synthetic data that supports our previous results.
\end{itemize}

\begin{figure}[t]
\hspace{0.15cm}
\centering
  \hfill%
  \begin{minipage}[t]{0.45\linewidth}
    \centering
    \vspace{-1.6cm}
    \scalebox{0.75}{
\begin{tikzpicture}
    \small
    \node[draw,shape=rectangle] (X) at (0, 0) {income};
    \node[draw,shape=rectangle] (Y) at (4.5/1.2, 0) {loan default};
    \node[draw,shape=rectangle,accepting] (S) at (0, -4.5/1.2) {loan issued};
    \node[draw,shape=rectangle, align=center, text width=1.7cm] (Z^-) at (4.5/1.2, -4.5/1.2) {spendings, loans, loan amount};
    \node[draw,shape=rectangle] (Z^+) at (1.75/1.2, -0.5/1.2) {job};
    \node[draw,shape=rectangle, align=center, text width=1.1cm, fill=gray!20] (U) at (3.5/1.2, -1.5/1.2) {financial literacy};

    \draw[->] (X) -- (Y);
    \draw[->] (X) -- (S);
    \draw[->] (X) -- (Z^-);
    \draw[->] (Z^-) -- (Y);
    \draw[->] (Z^-) -- (S);
    \draw[->] (Z^+) -- (X);
    \draw[->] (U) -- (Y);
    \draw[->] (U) -- (Z^-);
    \draw[->] (U) -- (Z^+);
    
    \end{tikzpicture}}
\end{minipage}
  \hfill%
\begin{minipage}[t]{0.25\linewidth}
    \centering
\scalebox{0.7}{
\begin{tabular}{cccc}
\textbf{$S$} & \textbf{$X$} & \textbf{$Z$} & \textbf{$Y$} \\
\hline
$1$ & $x_1$ & $z_1$ & $y_1$ \\
$1$ & $x_2$ & $z_2$ & $y_2$ \\
$\vdots$ & $\vdots$ & $\vdots$ & $\vdots$ \\ 
$1$ & $x_m$ & $z_m$ & $y_m$ \\
\textcolor{gray!80}{$0$} & \textcolor{gray!80}{$x_{m+1}$} & \textcolor{gray!80}{$z_{m+1}$} & \textcolor{gray!80}{$y_{m+1}$} \\
\textcolor{gray!80}{$\vdots$} & \textcolor{gray!80}{$\vdots$} & \textcolor{gray!80}{$\vdots$} & \textcolor{gray!80}{$\vdots$} \\ 
\textcolor{gray!80}{$0$} & \textcolor{gray!80}{$x_n$} & \textcolor{gray!80}{$z_n$} & \textcolor{gray!80}{$y_n$} \\
\end{tabular}}
\vspace{0.08cm}
\caption*{$\mathcal{S}$}

\caption*{}
  \end{minipage}
  \hfill%
  \begin{minipage}[t]{0.25\linewidth}
    \centering
    \vspace{-1.4cm}
    \scalebox{0.7}{
\begin{tabular}{cc}
 \textbf{$X$} & \textbf{$Z$} \\
\hline
 $x_1'$ & $z_1'$ \\
 $x_2'$ & $z_2'$ \\
 $\vdots$ & $\vdots$ \\ 
 $x_{n'}'$ & $z_{n'}'$ \\
\end{tabular}}
\vspace{1.1cm}
\caption*{$\mathcal{D}$}
  \end{minipage}
\caption{Example graph covered by Assumption~\ref{ass:new}. When estimating the causal effect of income $\Xv$ on loan default $\Yv$, the node ``loan issued`` represents the selection variable $\Sv$, ``financial literacy`` is an unobserved confounder, and all other nodes serve as proxies ($\Zv$). Dataset $\mathcal{S}$ is sampled from $P(X, Z, X | S = 1)$, i.e., the dataset is affected by a selection mechanism, where the gray rows are not observed. $\mathcal{D}$ contains unbiased data ($\mathcal{S} \cap \mathcal{D} = \emptyset$) of observations over $(X,Z)$. Given data from both sources, our estimator leverages $\mathcal{D}$ which includes proxies $Z^- \subseteq Z$ (``spendings, etc'') satisfying $S \Indep Y \mid \{ X, Z^- \}$ to adjust for selection bias and simultaneously corrects for confounding by adjustment based on $Z^+ \subseteq Z$ (``job''). In some situations, $\mathcal{D}$ and $\mathcal{S}$ may overlap.}
\label{fig:datasetsandrunning}
\end{figure}
Before diving into the details of our method, we discuss prior work.

\section{Related Work}
\label{sec:relatedwork}

Identifiability under confounding has been extensively studied through Pearl's do-calculus~\citep{pearl:09:causality}, or in the regression context through instrumental variable methods~\citep{imbens:94:iv}, that are applied in specific settings where confounders between treatment and target are unobserved. A data-driven variable decomposition ($D^2VD$) that jointly optimizes separation of variables into confounders and adjustment variables to handle confounding %
is proposed by \citet{kuang:17:treatment}. %
Further, researchers have employed proxy-based regression models through methods such as generalized least squares~\citep{louizos:17:deeplatentvariablemodels,Miao:18:identifyingcausaleffectswithproxyvariables} or VAEs~\citep{liu:25:proximal-CI}, to adjust for confounding, however, not investigating biased data.

Complementary to identifiably, recovering from selection bias (s-recoverability) has been studied by multiple authors. 
\cite{bareinboim:12:controlling} derived a complete condition indicating feasibility of recoverability of the odds ratio (OR) from selection biased data and offered a method enabling to recover other effect measures than OR from selection bias using instrumental variables. \citet{pearl:12:solution} and \citet{bareinboim:14:recovering-selection} considered the fundamental problem of the identifiability of $P(Y\mid X)$ and $P(Y \mid do(X))$ based on data potentially underlying a selection bias. These results have been extended to data fusion \citep{bareinboim:16:data-fusion}, and \citet{correa:17:CausalEI} established complete conditions for s-recoverability if no  external data is necessary and for the setting in which all proxy variables are observed externally. Moreover, \citet{correa:18:GeneralizedAdj} cover the case in which a subset of the proxy variables is observed externally. 
\citet{forre:20:causal-calculus} extended the backdoor and selection-backdoor criteria to structural causal models with cycles, and \citet{chen:24:modeling} introduced a conditioning operation enabling principled modeling of selection bias akin to confounding, focusing on $M$ structures. Beyond these closely related works, others have explored similar scenarios but require experimental or interventional data~\citep{singh:22:kernelmethodslongterm}.
Building on \citet{bareinboim:14:recovering-selection}, we derive a modified set of assumptions to guarantee s-recoverability of the causal effect in graphs as shown in \Cref{fig:datasetsandrunning}---a detailed comparison is provided in \Cref{sec:appassumptions}. Another distinction to the works above is that we derive a practical estimator for continuous data.

Next, we discuss estimation methods that aim to address selection bias which can arise through systematic missingness or preferential inclusion of data points, which in some settings can even be detected from observational data~\citep{dorst:20:detecting-and-mitigating,kaltenpoth2023identifying}. Fundamental achievements of research about missing data settings are given in \citet{dempster:77:EMalgorithm} on the EM algorithm for Maximum Likelihood estimation, \citet{heckman:79:SpecError} on correcting for selection bias in linear regression and \citet{rosenbaum:84:SubclasProp} on bias reduction through subclassification on the propensity score for binary treatments. Further, \citet{mohan:21:GraphicalModels} 
derived a consistent estimation method in missing data problems, however, not focusing on continuous data.  Other approaches aim to transform the data, eliminating differences between selected and unobserved data, e.g., by conditioning on the propensity score so that covariate distributions of selected and unobserved data become similar~\citep{austin:11:PSreducingConf,rosenbaum:83:PSinobs}.
Balancing diagnostics are needed to assess whether the weighting balanced the distribution of covariates of the selected and unobserved data~\citep{austin:15:IPTWusingPS}. Super Learning %
can reduce bias and increase efficiency at the cost of an increase in computing time \citep{neugebauer:16:PSonCIfromIPW}. Doubly robust estimators estimate the outcome regression as well as the missingness mechanism to get rid of the effect on the outcome due to differences between selected and unobserved data~\citep{diaz:17:DoubleRobustInference,linden:17:ImprovingCI}.
Closest to our approach, \citet{boeken:23:privileged-info} focus on continuous treatments with proxy variables, whereas they do not cover confounding. We discuss their approach in more detail in \Cref{sec:absence}.\!\footnote{Concurrent work by \citet{de-aguas:25:recovery} discusses a similar approach following a sequential adjustment procedure for binary treatment variables. We discuss the relation to their work in Appendix~\ref{sec:appassumptions}.}
\section{Recoverability and Identifiability}
\label{sec:theory}

In the following, 
we first outline the connection between missingness and selection bias, and introduce relevant notation and definitions to define recoverability from selection bias in \Cref{sec:notation-and-missingness}. In \Cref{sec:absence}, we review recoverability from selection bias without confounding, as studied by \citet{boeken:23:privileged-info}. Subsequently, in \Cref{sec:our-identification-result}, we derive a set of assumptions which ensure that the \emph{causal effect is identifiable and recoverable from selection-biased data}.
All proofs are supplemented in Appendix~\ref{sec:appproofs}.

\subsection{Preliminaries and Notation}
\label{sec:notation-and-missingness}

Throughout this paper, we follow the notational conventions introduced by \citet{pearl:09:causality}. We consider a causal directed acyclic graph (DAG) model $(\mathcal{G}, P)$, where $P$ defines a distribution over the set of random variables $V$, which factorizes according to $\mathcal{G}$,
and is consistent under interventions. $V$ includes all variables of interest ($\{\Yv,\Xv,\Zv\}\subset V$) except for the binary \emph{selection variable} $\Sv$, where $\Sv=1$ denotes selection. That is, under selection bias, we only observe $P(V \mid \Sv=1)$. We denote the graph augmented with the selection variable as $\gG_s$
Further, $\Yv \in \mathbb{R}$ denotes the one-dimensional continuous \emph{target} random variable, $\Xv \in \mathbb{R}^p$, with $p \ge 1$, the potentially multidimensional and continuous random vector of %
\emph{treatments}. Our interest lies in estimating the \emph{expected causal effect} $E[\Yv\mid do(\Xv)]$, denoting the expected value of %
$\Yv$ when $\Xv$ is set to a specific %
value by intervention (\emph{hard or surgical intervention}). 
The potentially multidimensional and continuous random vector $\Zv \in \mathbb{R}^d$, with $d \ge 0$, may include confounding variables between $X$ and $Y$ and \emph{proxy variables} for the missing observations. The symbol $\Indep_{\mathcal{G}}$ denotes d-separation in graph $\mathcal{G}$. The graph obtained by deleting all edges pointing towards (away from) a node in $X$ is denoted by $\mathcal{G}_{\overline{X}}$ ($\mathcal{G}_{\underline{X}}$).

\textbf{Missingness and Selection Bias}~~~\emph{Selection bias} is typically induced by preferential selection or systematic missingness~\citep{correa:19:CausalEffectSelection}, also know as \emph{missing not at random} (MNAR), where generally, $E[\Yv\mid \Xv] \neq E[\Yv\mid \Xv,\Sv=1]$. Therefore, proper adjustment is necessary when aiming to estimate $E[\Yv\mid \Xv]$ or %
$E[\Yv\mid do(\Xv)]$ from data affected by such systematic missingness. 
To approach this problem, we need to state some assumptions about the missingness scenario~\citep{little:02:missing-data}. In the following, we distinguish between two missing data settings illustrated in \Cref{fig:datasetsandrunning} (right), that are consistent with prior work~\citep{boeken:23:privileged-info}. For both settings, we have independent and identically distributed (i.i.d.) observations of $(\Xv,\Yv,\Zv)\sim P(\Xv,\Yv,\Zv\mid \Sv=1)$, where $P(\Xv,\Yv,\Zv\mid S=1)$ denotes the joint distribution of $\Xv$, $\Yv$ and $\Zv$ conditioned on $\Sv=1$. All samples drawn under this condition are indexed by index set $\mathcal{S}$. Additionally, we observe i.i.d.~realizations of $(\Xv,\Zv)\sim P(\Xv,\Zv)$ with index set $\mathcal{D}$ not underlying a selection process. In the first setting, the selected sample is a subset of the data not underlying the selection process ($\mathcal{S}\subset\mathcal{D}$). That is, for $\Sv=0$, only the observations of $\Yv$ are missing. For the second setting, we have access to \emph{external data} $\mathcal{D}$ which contain observations of $X$ and $Z$, having no overlap with the observations of $X$ and $Z$ from $\mathcal{S}$ ($\mathcal{S}\cap\mathcal{D}=\emptyset$). Following the notation from \citet{bareinboim:14:recovering-selection}, we call the unbiased data over $X$ and $Z$ in both cases external data. 
If not stated otherwise, our results derived below hold for both settings. 

\textbf{Recoverability}~~~Before introducing a practical estimator, we need to ensure that the causal effect, which we are interested in, is recoverable from the %
available data.
For %
cases involving selection bias, \citet{pearl:12:solution} first proposed the concept of \emph{s-recoverability}, which was later developed and defined as in Definition~\ref{def:srecover} in Appendix~\ref{app:definitions} by \citet{bareinboim:14:recovering-selection}. 
As there are settings that are only s-recoverable under consulting external unbiased data ($\mathcal{D}$)---which is the setting we focus on---\citet{bareinboim:14:recovering-selection} formulated a compatible definition for this case, %
restated below.
\begin{definition}%
\label{def:srecoverexternal}
Given a causal DAG model $(\mathcal{G}_s,P)$
augmented with a node $S$, the distribution $Q = P(\Yv\mid \Xv)$ is said to be \emph{s-recoverable} from selection bias in $\mathcal{G}_s$ %
with external information over $T \subset V$ and selection-biased data
over $M \subset V$ if the assumptions embedded in the causal model render Q expressible in terms
of \mbox{$P(M=m\mid S = 1)$} and $P(T=t)$, both positive. 
\end{definition}
Next, we will review %
selection bias induced by systematic missingness for which $P(\Yv\mid \Xv)$ is s-recoverable when observing \emph{privileged} information. %

\subsection{Recovering from Selection Bias without Confounding}
\label{sec:absence}

To recover from selection-biased data, \citet{boeken:23:privileged-info} presented a special case of missing not at random (MNAR), the concept of \emph{privilegedly missing at random} (PMAR), which is also known as comparability \citep{singh:22:kernelmethodslongterm,miao:24:shadowvariable}, in works that require access to interventional/experimental data to recover from selection bias and confounding. PMAR describes cases in which the target variable $\Yv$ is stochastically independent from the selection variable $\Sv$, when conditioning on the %
treatments $\Xv$ and proxy variables $\Zv$, as formalized in the assumption below.

\begin{figure}[t]
  \centering
  \begin{minipage}[t]{0.15\linewidth}
    \centering
    \scalebox{0.8}{
\begin{tikzpicture}[
    >={Stealth[round]}, %
    node distance=1cm, %
    every node/.style={circle, inner sep=-3pt, draw, minimum size=0.75cm} %
]
    \node[thick] (X) {X};
    \node[thick] (Y) [right=of X] {Y};
    \node[thick] (S) [below=of X, accepting] {S};
    \node[thick] (Z) [below=of Y] {\small $Z^+$};

    \draw[->, thick] (X) -- (S);
    \draw[->, thick] (Z) -- (S);
    \draw[->, thick] (X) -- (Y);
    \draw[->, thick] (Z) -- (Y);
    \node (Q) [below=-0.4cm of S, draw=none] {};

\end{tikzpicture}}
    \caption*{(a)}
  \end{minipage}%
  \hfill%
  \begin{minipage}[t]{0.15\linewidth}
    \centering
    \scalebox{0.8}{
\begin{tikzpicture}[
    >={Stealth[round]}, %
    node distance=1cm, %
    every node/.style={circle, inner sep=-3pt, draw, minimum size=0.75cm} %
]
    \node[thick] (X) {X};
    \node[thick] (Y) [right=of X] {Y};
    \node[thick] (S) [below=of X,accepting] {S};
    \node[thick] (A) [below=of Y] {\small$Z^-$};

    \draw[->,thick] (X) -- (S);
    \draw[->,thick] (A) -- (S);
    \draw[->,thick] (X) -- (Y);
    \draw[->,thick] (A) -- (Y);
    \draw[->,thick] (X) -- (A);
    \node (Q) [below=-0.4cm of S, draw=none] {};

\end{tikzpicture}}
\caption*{(b)}
  \end{minipage}
  \hfill%
  \begin{minipage}[t]{0.15\linewidth}
    \centering
    \scalebox{0.8}{
\begin{tikzpicture}[
    >={Stealth[round]}, %
    node distance=1cm, %
    every node/.style={circle, inner sep=-3pt, draw, minimum size=0.75cm} %
]   
    \node[thick] (X) {X};
    \node[thick] (Y) [right=of X] {Y};
    \node[thick] (S) [below=of X,accepting] {S};
    \node[thick] (A) [below=of Y] {\small$Z^+$};
    \node[thick] (Q) [below=-0.4cm of S, draw=none] {};

    \draw[->,thick] (X) -- (S);
    \draw[->,thick] (A) -- (S);
    \draw[->,thick] (X) -- (Y);
    \draw[->,thick] (A) -- (Y);
    \draw[->,thick] (A) -- (X);

\end{tikzpicture}}
\caption*{(c)}
  \end{minipage}
  \hfill%
  \begin{minipage}[t]{0.2\linewidth}
    \centering
    \scalebox{0.7}{
\begin{tikzpicture}[
    >={Stealth[round]}, %
    every node/.style={circle, inner sep=-3pt, draw, minimum size=0.75cm} %
    ]
\node[thick,draw,shape=circle] (X) at (0, 0) {X};
\node[thick,draw,shape=circle] (Y) at (6/2.3, 0) {Y};
\node[thick,draw,shape=circle,accepting] (S) at (0, -6/2.3) {S};
\node[thick,draw,shape=circle] (Z^-) at (6/2.3, -6/2.3) {\small$Z^-$};
\node[thick,draw,shape=circle] (Z^+) at (3/2.3, -1.2/2.3) {\small$Z^+$};
\node[thick,draw,shape=circle,fill=gray!20] (U) at (4.7/2.3, -3/2.3) {U};

\draw[->,thick] (X) -- (Y);
\draw[->,thick] (X) -- (S);
\draw[->,thick] (X) -- (Z^-);
\draw[->,thick] (Z^-) -- (Y);
\draw[->,thick] (Z^-) -- (S);
\draw[->,thick] (Z^+) -- (X);
\draw[->,thick] (U) -- (Y);
\draw[->,thick] (U) -- (Z^-);
\draw[->,thick] (U) -- (Z^+);

\end{tikzpicture}}
\caption*{(d)}
  \end{minipage}

\caption{Example graphs $\gG_s$ augmented with $S$ that are consistent with Assumption~\ref{ass:new}. For graphs (a) and (b), we only need to adjust for selection bias since $E[Y \mid X]$ coincides with $E[Y \mid do(X)]$, while for graphs (c) and (d), adjustments for selection bias and confounding, as described in \Cref{sec:our-identification-result}, are required.}
\label{fig:vierDAGs}
\end{figure}

\begin{assumption}[PMAR~\citep{boeken:23:privileged-info}]
 \label{ass:PMAR}
Given a privilegedly observed set of variables $\Zv$, $\Yv$ is privilegedly missing at random (PMAR) if \(\Sv \Indep \Yv \mid \{\Xv,\Zv\}.\) 
\end{assumption}
Intuitively, PMAR holds when treatment and proxies block all paths between target and \mbox{selection} variable in $\gG_s$ due to the Markov property, %
as in the augmented DAG in Figure~\ref{fig:vierDAGs} (a). Under Assumption~\ref{ass:PMAR}, by law of total expectation, \[E[\Yv \mid \Xv] = E[E[\Yv \mid \Xv,\Zv]\mid \Xv]= E[E[\Yv \mid \Xv,\Zv, \Sv = 1]\mid \Xv]\;.\]
As discussed by \citet{boeken:23:privileged-info}, concerning discrete variables under certain positivity assumptions, \mbox{s-recoverability} is satisfied when selection biased \mbox{$P(\Xv, \Yv, \Zv\mid \Sv = 1)$} from $\mathcal{S}$, as well as unbiased external $P(\Xv, \Zv)$ from $\mathcal{D}$, are available and PMAR holds. They proposed a repeated regression (RR) %
for estimating $E[\Yv\mid \Xv]$. 
The first regression models $E[\Yv \mid \Xv, \Zv, \Sv = 1]$ using $\mathcal{S}$. Next, %
predictions $\widetilde{\Yv} := \widehat{E}[\Yv \mid \Xv, \Zv, \Sv = 1]$ based on the first regression are computed on population-level in dataset $\mathcal{D}$, which is unaffected by %
selection. %
A second regression models $E[\widetilde{\Yv}\mid \Xv]$ using $\mathcal{D}$, yielding the final estimate \mbox{$\widehat{\mu}_{RR}(x) = \widehat{E}[\widetilde{\Yv}\mid \Xv]$ for $E[\Yv\mid \Xv]$}. Besides repeated regression, \citet{boeken:23:privileged-info} also proposed an estimator based on inverse probability weighting and a doubly robust estimator, which were, however, clearly outperformed by RR.

In absence of confounding between $\Xv$ and $\Yv$, this method also provides a reliable estimate for \mbox{$E[\Yv\mid do(\Xv)]$} as $E[\Yv\mid do(\Xv)]=E[\Yv\mid \Xv]$, %
which is the case in \mbox{\Cref{fig:vierDAGs} (a) and (b)}. However, if %
we flip the edge between $\Xv$ and $\Zv$ in \Cref{fig:vierDAGs} (b), arriving at %
\Cref{fig:vierDAGs} (c), %
we additionally need to adjust for confounding. 

\subsection{Identification under Selection Bias and Confounding}
\label{sec:our-identification-result}

Criteria for causal effect identification and s-recoverability under confounding and selection bias have been generalized by multiple authors \citep{pearl:12:solution,bareinboim:15:algo,correa:18:GeneralizedAdj,correa:19:CausalEffectSelection}. To treat both sources of bias, they propose to decompose $\Zs=\Zps\cup \Zms$ into $\Zps$, the set of the non-descendants of $\Xs$ and $\Zms$, the set of descendants of $\Xs$ that are included in $\Zs$. Based on this distinction, \citet{bareinboim:14:recovering-selection} introduced the \emph{selection backdoor criterion} (provided in Assumption~\ref{ass:Bareinboim}) under which the causal effect is identifiable and s-recoverable. We adjust the assumptions proposed by \citet{bareinboim:14:recovering-selection}, as stated in Assumption~\ref{ass:new} below, to ensure identifiability and s-recoverability of the causal effect for PMAR with potentially unobserved confounding, which we can adjust for by extending the backdoor criterion under Assumption~\ref{ass:new}. An example graph, which is not covered by previous approaches is shown in \mbox{\Cref{fig:vierDAGs} (d)}, where $U$ is not included in $Z$ and may be an unobserved confounder.
We compare our assumptions in more detail with prior works \citep{bareinboim:14:recovering-selection,correa:18:GeneralizedAdj} in Appendix~\ref{sec:appassumptions}.

\begin{assumption}
\label{ass:new}
Decompose the set of variables $\Zs$ into $\Zs=\Zps\cup \Zms$, where $\Zps$ are non-descendants of $\Xs$ and $\Zms$ are descendants of $\Xs$. Assume that
\begin{enumerate}[topsep=2pt,parsep=2pt,partopsep=1pt,leftmargin=*]
    \item $\Xv$ and $\Zv$ block all paths between $\Sv$ and $\Yv$, namely $\Sv \Indep_{\mathcal{G}_s} \Yv \mid \{\Xv,\Zv\}$ (PMAR),
    \item $\Zpv$ blocks all backdoor paths between $\Xv$ and $\Yv$, namely $\Yv \Indep_{{\gG_s}_{\underline{X}}} \Xv \mid \Zpv$, 
\item $\Zs\cup \{\Xv,\Yv\}\subset M$, where variables $M$ are collected under selection bias (dataset $\mathcal{S}$) and $\Zs,X\subset T$, where $T$ is collected on population-level (dataset $\mathcal{D}$).
\end{enumerate}
\end{assumption}

\emph{\textbf{Remark:} When Assumption~\ref{ass:new} is satisfied and  $\Zps=\emptyset$, confounding is excluded and we recover the setting from \citet{boeken:23:privileged-info}. %
$\Zpv$ shields the confounding of $\Xv$ and $\Yv$, and both $\Zv$ and $\Xv$ are needed to adjust for the selection bias. As we will discuss later, under linearity and $\Zms=\emptyset$, we do not need to observe $\Xv$ unbiased.}

To illustrate the assumptions stated above, consider the example graph in \Cref{fig:datasetsandrunning}, which fulfills all structural constraints with \emph{income} as $\Xs$, \emph{job} belonging to $\Zpv$, and $\Zmv = \{\text{\emph{spending, loans, loan amount}}\}$. We may have additional measurements from an unbiased source of $\Xs$ and $\Zs$, for which the label is not available, e.g., because collecting the label is costly as it involves providing a loan to an individual.

When Assumption~\ref{ass:new} is satisfied, the causal effect is identifiable and s-recoverable:

\begin{restatable}{theorem}{neutheoretisch}
\label{theorem:neutheoretisch}
Under Assumption~\ref{ass:new}, the causal effect $E[\Yv\mid do(\Xv)]$ is identifiable, s-recoverable and can be expressed as follows
\begin{align*}
 \int_{z^+}E[E[\Yv\mid \Xv,\Zpv=z^+,\Zmv,\Sv=1]\mid \Xv,\Zpv=z^+]P(\Zpv=z^+)dz^+   \; .
\end{align*}
\end{restatable}
\textit{\textbf{Proof Sketch:} Start with expressing $E[Y\mid do(X)]$ as $\int_{z^+}E[Y\mid do(X),Z^+=z^+]P(Z^+=z^+\mid do(X))dz^+$. Rewrite the first term by using the second rule of do-calculus (Assumption~\ref{ass:new} (2.)), the law of iterated expectations and PMAR (Assumption~\ref{ass:new} (1.)). The second term simplifies since $Z^+$ are non-descendants of $X$.}\\

Based on the identification result above, we develop a practical estimator for continuous targets. %
For notational reasons, we %
propose an estimator for linear Gaussian cases and explain its extension to %
the non-linear setting in \Cref{sec:non-linearity}.
\begin{assumption}
\label{ass:linearity}
Let any observation of $\Yv$ be defined through the following assignment:
\[
y := \beta_0+\beta_1 x+\beta_2z^++\beta_3 z^-+\epsilon\; ,
\]

where $(x,y,z)$ are drawn i.i.d. from $P(\Xv,\Yv,\Zv)$ and $\epsilon$ is drawn i.i.d. from a standard normal distribution $\mathcal{N}(0,1)$. The coefficients $\beta_0$, $\beta_1$, $\beta_2$ and $\beta_3$ are of the dimension of its corresponding vector of variables $\Xv\in\mathbb{R}^p$, $\Zpv\in\mathbb{R}^{d_1}$ or $\Zmv\in\mathbb{R}^{d_2}$ respectively. 
\end{assumption}
Based on Assumption~\ref{ass:linearity}, Theorem~\ref{theorem:neutheoretisch} simplifies to Theorem~\ref{theorem:newassumedlinearity} which implies Corollary~\ref{corollary:ZminusgivendoX} as a special case, where we require additional assumptions about $Z^+$.
\begin{restatable}{theorem}{theoremlinearity}
\label{theorem:newassumedlinearity}
Let Assumption~\ref{ass:new} and Assumption~\ref{ass:linearity} hold, then the causal effect $\mu(x) := E[\Yv\mid do(\Xv := x)]$ is identifiable, s-recoverable and can be expressed as 
\begin{equation}
\mu(x)=\beta_0+\beta_1x+\beta_2 E[\Zpv] + \beta_3\int_{z^+}E[\Zmv\mid X=x,\Zpv=z^+]P(\Zpv=z^+)dz^+.
\label{eq:eydox_linearity}
\end{equation}
\end{restatable}
\begin{restatable}{corollary}{ZminusgivendoX}
\label{corollary:ZminusgivendoX}
Let the conditions given in Theorem~\ref{theorem:newassumedlinearity} hold. If further $\Zpv$ blocks all backdoor paths between $\Xv$ and $\Zmv$, the integral in Equation~\ref{eq:eydox_linearity} reduces to $E[\Zmv\mid do(\Xv)]$. If additionally, $\Xv$ and $\Zmv$ are not confounded, it reduces to $E[\Zmv\mid \Xv]$.
\end{restatable}

\section{A Two-Step Regression Estimator}
\label{sec:our-estimator}

Based on the theoretical results derived in the previous section, we now propose our practical linear estimator and simplifications of it, analyze its bias and variance in \Cref{sec:biasvariance}, and discuss potential non-linear extensions in \Cref{sec:non-linearity}.

To derive an estimator for $E[\Yv\mid do(\Xv)]$ we substitute each component of the causal effect expression in \Cref{eq:eydox_linearity} with its corresponding estimator. We refer to this estimator as the \textbf{Two-Step Regression (TSR)} estimator.
For a specific value of $x$, it is defined as
\begin{equation}
\label{eq:tsr-genral}
\hat{\mu}_{TSR}(x) = \tikzmarkin[color=ForestGreen]{beta0}(-0.4,0.4)(0.4,-0.2)\hat{\beta}_0\tikzmarkend{beta0}+\tikzmarkin[mark at=1,color=ForestGreen]{beta1}(-0.4,0.4)(0.4,-0.2)\hat{\beta}_1\tikzmarkend{beta1}\; x+\tikzmarkin[mark at=1,color=ForestGreen]{beta2}(-0.4,0.4)(0.4,-0.2)\hat{\beta}_2\tikzmarkend{beta2} \;\,\tikzmarkin[mark at=1,color=NavyBlue]{meanZ}(-1,0.4)(1,-0.2)\widehat{E}[\Zpv]\tikzmarkend{meanZ}+\tikzmarkin[mark at=1,color=ForestGreen]{beta3}(-0.4,0.4)(0.4,-0.2)\hat{\beta}_3\tikzmarkend{beta3}\; \int_{z^+}\tikzmarkin[mark at=1,color=NavyBlue]{condmean}(-3.85,0.4)(3.85,-0.2)\widehat{E}[\Zmv\mid \Xv=x,\Zpv=z^+]\tikzmarkend{condmean}\;\;\tikzmarkin[mark at=1,color=NavyBlue]{density}(-2,0.4)(2,-0.2)\hat{P}(Z^+=z^+)\tikzmarkend{density}\;\,dz^+
\;,  
\tikz[beschr]\draw[pfeil,color=ForestGreen](-11.3, -0.25)|-+(0.2,-0.85)node[right]{\text{OLS: $\Yv\sim \Xv,\Zpv,\Zmv$}};
  \tikz[beschr]\draw[pfeil,color=ForestGreen](-10.65,-0.25)|-+(0,-0.65)node[right]{};
  \tikz[beschr]\draw[pfeil,color=ForestGreen](-9.7,-0.25)|-+(0,-0.65)node[right]{};
  \tikz[beschr]\draw[pfeil,color=ForestGreen](-7.85,-0.25)|-+(-0.2,-0.85)node[left]{};
  \tikz[beschr]\draw[pfeil,color=NavyBlue](-8.85,-0.25)|-+(0,-0.25)node[below]{emp. mean};
  \tikz[beschr]\draw[pfeil,color=NavyBlue](-4.95,-0.25)|-+(0,-0.25)node[below]{\text{OLS: $Z^-\sim X, Z^+$}};
  \tikz[beschr]\draw[pfeil,color=NavyBlue](-2,-0.25)|-+(0,-0.25)node[below]{density est.};
  \vspace{1cm}
\end{equation}
where the components, highlighted in \textcolor{ForestGreen}{green}, can be estimated \textcolor{ForestGreen}{based on the selected dataset $\mathcal{S}$}, whereas the estimates, highlighted in \textcolor{NavyBlue}{blue}, \textcolor{NavyBlue}{require access to an external dataset $\mathcal{D}$}, which is not underlying the selection mechanism.
In particular, we obtain the estimates $\hat{\beta}_0$, $\hat{\beta}_1$, $\hat{\beta}_2$, $\hat{\beta}_3$ by OLS for the model \mbox{$E[\Yv\mid \Xv,\Zpv,\Zmv,\Sv=1]=\beta_0+\beta_1\Xv+\beta_2\Zpv+\beta_3\Zmv$} based on the observations in $\mathcal{S}$ in the first step. In the second step, we estimate $E[\Zpv]$ by its empirical mean and approximate the integral \mbox{$\int_{z^+}E[\Zmv\mid \Xv=x,\Zpv=z^+]P(\Zpv=z^+)dz^+$} by OLS estimations of \mbox{$E[\Zmv\mid \Xv=x,\Zpv=z^+]$} weighted by an estimation of the density 
$P(\Zpv=z^+)$ of $\Zpv$, both based on observations from $\mathcal{D}$.

\textbf{Variants of TSR}~~~In the following, we discuss several possible instantiations of our estimator depending on whether or not the data is affected by confounding and whether or not certain sets of variables are empty. For all settings, we assume that the considered variables meet the assumptions required for Theorem~\ref{theorem:newassumedlinearity}.

When $Z^+=\emptyset$ and implicitly $Z=Z^-$, as in \Cref{fig:vierDAGs} (b), there is \emph{no confounding} between $X$ and $Y$. Therefore, $E[\Yv\mid \Xv]=E[\Yv\mid do(\Xv)]$ and TSR reduces to 
\begin{align*}
\hat{\mu}_{TSR}(x)=\hat{\beta}_0+\hat{\beta}_1 x+\hat{\beta}_2 \widehat{E}[\Zmv\mid \Xv=x]\;,   
\end{align*}
which coincides with repeated regression %
(cf. Appendix~\ref{sec:appcoincide}).%

In scenarios for which we cannot exclude \emph{confounding} and thus $E[\Yv\mid do(\Xv)]=E[\Yv\mid \Xv]$ cannot be ensured, we distinguish two cases that are illustrated in \Cref{fig:vierDAGs}.
First, consider the minimal example, presented in Figure~\ref{fig:vierDAGs} (c), with $\Zs=\Zps$ and hence $\Zms=\emptyset$. Here, the TSR estimator reduces to 
\begin{align}
   \hat{\mu}_{TSR}(x)=&\hat{\beta}_0+\hat{\beta}_1 x+\hat{\beta}_2 \widehat{E}[\Zpv]\;,
   \label{eq:TSRex12}
\end{align}
which differs from the RR estimation in that, for RR, $\widehat{E}[Z^+ | X]$ appears instead of $\widehat{E}[Z^+]$ (cf. Appendix~\ref{sec:appcoincide}).
The second estimation step only requires calculating the empirical mean of $\Zpv$. This is intuitively comprehensible because \mbox{$E[\Zpv\mid do(\Xv)]=E[\Zpv]$} when $\Zpv$ is a non-descendant of $\Xv$. Hence, in this setting, only $\Zpv$, but not $\Xv$, needs to be observed in an external unbiased dataset.

For our running example in \Cref{fig:datasetsandrunning} and \Cref{fig:vierDAGs} (d), %
TSR is given by Equation~\ref{eq:tsr-genral}.
Whenever we can assume linearity %
\mbox{$E[\Zmv\mid \Xv=x,\Zpv=z^+]=\gamma_0+\gamma_1 x+\gamma_2 z^+$} with $\gamma_0\in\mathbb{R}^{d_2}$, $\gamma_1\in\mathbb{R}^{d_2\times p}$ and $\gamma_2\in\mathbb{R}^{d_2\times d_1}$, 
TSR takes the form
\begin{align}
\label{eq:estimator-unobserved}
\hat{\mu}_{TSR}(x)=&\textcolor{ForestGreen}{\hat{\beta}_0}+\textcolor{ForestGreen}{\hat{\beta}_1} x+\textcolor{ForestGreen}{\hat{\beta}_2} \textcolor{NavyBlue}{\widehat{E}[\Zpv]}+\textcolor{ForestGreen}{\hat{\beta}_3}   
(\textcolor{NavyBlue}{\hat{\gamma}_0}+\textcolor{NavyBlue}{\hat{\gamma}_1} x+ \textcolor{NavyBlue}{\hat{\gamma}_2} \textcolor{NavyBlue}{\widehat{E}[\Zpv]})\;,
\end{align}
imputing the coefficient estimates of the \textcolor{ForestGreen}{regression of $\Yv$ on $\Xv,\Zpv,\Zmv$ in $\mathcal{S}$} and of the \textcolor{NavyBlue}{regression of $\Zmv$ on $\Zpv$ and $\Xv$ in $\mathcal{D}$}, as well as the \textcolor{NavyBlue}{mean estimate of $\Zpv$ in $\mathcal{D}$}.

In addition to observed confounding, TSR can handle \emph{unobserved confounding} between $\Xv$ and $\Yv$ for cases in which $\Zpv$ blocks all back-door paths between $\Xv$ and $\Yv$ which arise through the confounder.

\textbf{Regularization}~~~Note that since the variables in $\Xv$ and $\Zv$ might be highly correlated in $\mathcal{S}$, it can be profitable to implement the first regression in TSR and RR with a ridge regression penalty, to reduce the variance of the estimator. In addition, even for the second estimation step, ridge regression in $\mathcal{D}$ should be considered because, for instance, in \Cref{fig:vierDAGs} (d), $\Xv$ and $\Zpv$ are correlated. In our empirical evaluation in \Cref{sec:expvariance}, we therefore also instantiate both TSR and RR with a ridge penalty.

\subsection{Analysis of Bias and Variance}
\label{sec:biasvariance}

Next, we examine unbiasedness, derive the variance of the proposed two-step regression (TSR) estimator and compare it to the variance of the repeated regression (RR) estimator from \citet{boeken:23:privileged-info} for graphs aligned with \Cref{fig:vierDAGs} (a), where $E[\Yv\mid \Xv]=E[\Yv\mid do(\Xv)]$, $\Zs=\Zps$ and $\Zms=\emptyset$. For simplicity, we assume that all variables are univariate and linearly related, to get an intuition of the bias and variance of both estimators. We formalize our assumptions below. 
\begin{assumption}
\label{ass:variance}
Let any observation $y$ be generated through the following assignment:
  \[
  y:= \beta_0+\beta_1 x+\beta_2 z^++\epsilon\;,
  \]
  where $(x,z^+)$ are drawn i.i.d. from $P(\Xv,\Zpv)$. In particular, $z^+=\mu_{z^+}+\xi$, with $\mu_{z^+} \in \mathbb{R}$, $\xi$ and $\epsilon$ are drawn i.i.d. from $\mathcal{N}(0,1)$. Value $x$ and the coefficients $\beta_0, \beta_1, \beta_2$ are in $\mathbb{R}$.
\end{assumption}

In this simplified setting, TSR reduces to $E[\Yv\mid do(\Xv=x)]=\beta_0+\beta_1x+\beta_2 \mu_{Z^+}$, where $\mu_{Z^+} = E[\Zpv\mid do(\Xv=x)]$.
Thus, the second step of the TSR estimator reduces to estimating the mean of $\Zpv$. In contrast, RR performs an OLS estimate of $E[\Zpv|\Xv=x]$ in $\mathcal{D}$, which is given by $\widehat{E}[\Zpv\mid X=x]=\alpha_0+\alpha_1 x$ with correct coefficients $\alpha_0=E[\Zpv]$ and $\alpha_1=0$. Assuming that in both regression steps, unbiased estimation is ensured, i.e., the chosen model class includes the ground truth generating mechanism, TSR and RR are ensured to be unbiased under $\mathcal{S}\cap\mathcal{D}=\emptyset$ %
as the data points for both regression steps are independent of each other and therefore the coefficient estimators of the two steps are. In contrast, for $\mathcal{S}\subset\mathcal{D}$, %
the bias in point $x$ is $Cov[\hat{\beta}_2,\overline{\Zpv}]$ for TSR and $Cov[\hat{\beta}_2,\hat{\alpha}_0+\hat{\alpha}_1 x]$ for RR (cf. Appendix~\ref{sec:appproofbiasvar}), which occurs %
as the part of the data that is used to estimate the $\alpha$ terms overlaps with the data used to estimate the $\beta$ coefficients. 
Investigating this %
empirically through simulation studies in \Cref{sec:expvariance} suggest that the bias terms for $\mathcal{S}\subset\mathcal{D}$ might be negligible. Intuitively, the smaller the overlap of $\mathcal{S}$ and $\mathcal{D}$, the smaller the dependence between the estimates of those two samples. We discuss the bias in a more general case in Appendix~\ref{sec:appproofbiasvar}.

After studying the bias of both estimators, we now compare their variance %
restricting %
to %
$\mathcal{S}\cap\mathcal{D}=\emptyset$, exploiting the independence of the observations of $\mathcal{S}$ and $\mathcal{D}$.

\begin{restatable}{theorem}{theoremvariance}
\label{theorem:variance}
Under Assumption~\ref{ass:variance} and $\mathcal{S}\cap\mathcal{D}=\emptyset$, let $\widehat{E}[Z^+\mid X=x]=\hat{\alpha}_0+\hat{\alpha}_1 x$ be the OLS estimate of the second step for $\hat{\mu}_{RR}(x)$ and $\bar{\Xv}=\frac{1}{\mid\mathcal{D}\mid}\sum_{i=1}^{\mid\mathcal{D}\mid} \Xv_i$, then
\begin{enumerate}[label=(\roman*)]
    \item $Var[\hat{\mu}_{TSR}(x)]=Var[\hat{\beta}_0+\hat{\beta}_1 x+\hat{\beta}_2\overline{Z^+}] + 2 Cov[\hat{\beta}_0+\hat{\beta}_1 x+\hat{\beta}_2\overline{Z^+},\hat{\beta}_2\hat{\alpha}_1(x-\bar{X})]$, and
    \item $Var[\hat{\mu}_{RR}(x)]-Var[\hat{\mu}_{TSR}(x)]=Var[\hat{\beta}_2\hat{\alpha}_1(x-\bar{\Xv})]\geq 0$.
\end{enumerate}
\end{restatable}

The result implies that in the second regression step, the irrelevant regressor $\Xv$ inflates the variance in small samples. The magnitude of the difference between the variances depends on the estimated effect of $\Zv$ on $\Yv$, the error in estimating $\alpha_1=0$ and the distance between $x$ and the empirical mean of the distribution of $\Xv$ in $\mathcal{D}$. As $\hat{\alpha}_1$ converges to zero when the sample size of $\mathcal{D}$ goes towards infinity, the difference between the variances of RR and TSR at point $x$ also converges to zero. %

The above results imply that we expect a lower mean squared error (MSE) for TSR than for RR, which we confirm by simulation-based experiments in \Cref{sec:expvariance}. Subsequently, we evaluate both estimators with a ridge regression penalty
and observe that regularization can reduce the mean squared error by introducing some bias.

\subsection{Introducing Non-linearity}
\label{sec:non-linearity}
In the previous sections, we outlined our theory for a linear estimator. As standard, we can extend the TSR estimator to non-linear settings by considering feature maps of the inputs. In particular, we can exchange $\Xv$, $\Zpv$ or $\Zmv$ by vectors $\varphi_X(\Xv)$, $\varphi_{Z^+}(\Zpv)$, $\varphi_{Z^-}(\Zmv)$ respectively, where $\varphi_X$, $\varphi_{Z^+}$, $\varphi_{Z^-}$ denote feature maps from a vector of variables to a vector of functions of the variables in $X$, $Z^+$ and $Z^-$ respectively. For example, in our experiments we perform polynomial regression. The linear case, can hence be seen as a special case with polynomials up to degree 1. 

Notably, including non-linear features comes with the following well-known challenges. First, we need to include correct feature maps in our estimation method, and second, the estimation of boundary regions is more challenging, especially low-density regions of $\mathcal{S}$ on the support of $\mathcal{D}$. Functionals that cannot be written in a linear form of feature maps can be estimated based on \Cref{eq:tsr-genral} using appropriate methods (like splines) and density estimation. Those, may also run into issues if extrapolation from low density regions of $\mathcal{S}$ is required.

\section{Experiments}
\label{sec:experiments}
In this section, we empirically evaluate the proposed Two-Step Regression (TSR) estimator, and compare it to Repeated Regression (RR)~\citep{boeken:23:privileged-info}, which estimates $E[E[\Yv \mid \Xv,\Zv, \Sv = 1]\mid \Xv]$ as detailed in Section~\ref{sec:absence}. We also instantiate both estimators with a ridge penalty in the regression based on $\mathcal{S}$ with penalization parameter $\lambda \in\{10^{-2}, 10^{-1.9}, ..., 10^2\}$, chosen via cross-validation with 10 folds. 
As a \emph{naive} baseline, 
we consider the OLS %
estimator trained only %
on $\mathcal{S}$, %
estimating $E[\Yv|\Xv,\Sv=1]$ instead of $E[\Yv|\Xv]$. 
We generate train and test data, both consisting of a selected dataset $\mathcal{S}$ and a population-level dataset $\mathcal{D}$. We chose the same sample size $n$ for $\mathcal{D}$ in both the test and the training data. The sample size for $\mathcal{S}$ is generated randomized by the selection process. All results are based on 100 simulation runs respectively.

\begin{figure}[t!]
   \begin{minipage}[b]{.28\linewidth}
      \centering
      \includegraphics[scale=0.18]{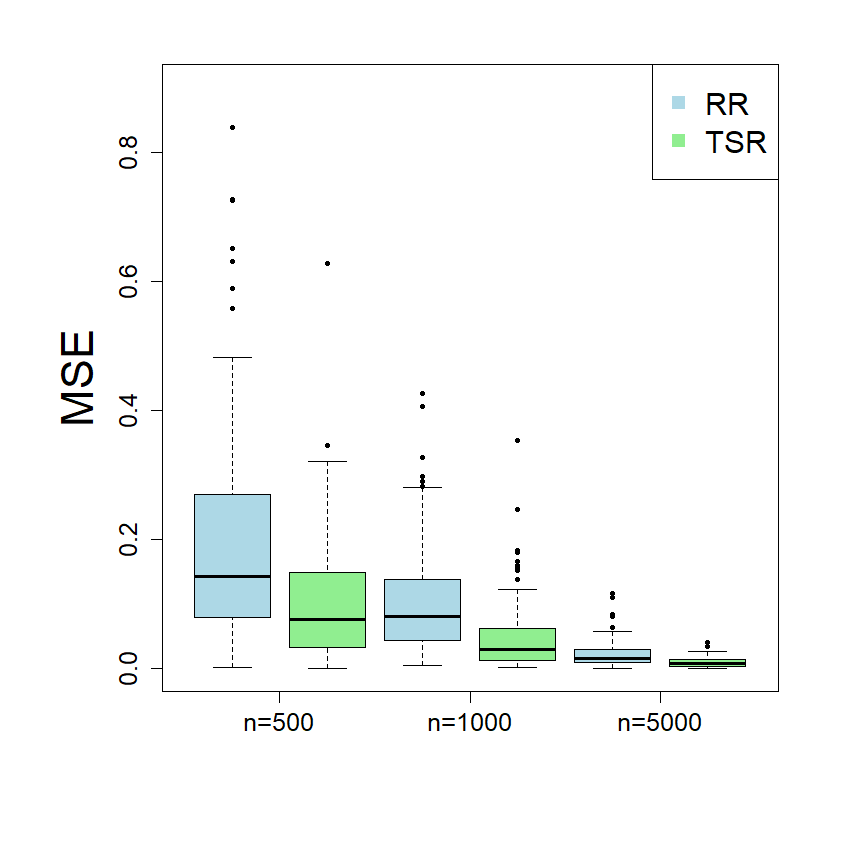}      
      \caption*{\scriptsize$\mathcal{S} \subset \mathcal{D}$}
   \end{minipage}
   \begin{minipage}[b]{.28\linewidth}
    \centering
      \includegraphics[scale=0.18]{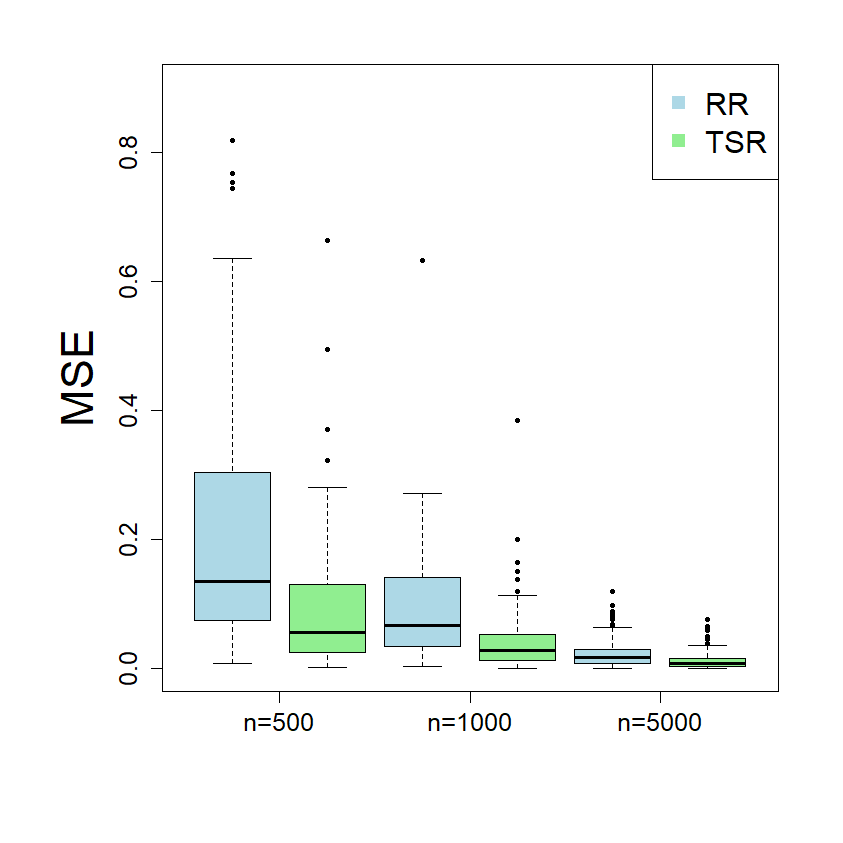}
      \caption*{\scriptsize$\mathcal{S}\cap\mathcal{D}=\emptyset$}
   \end{minipage}
   \begin{minipage}[b]{.4\linewidth} 
      \includegraphics[scale=0.18]{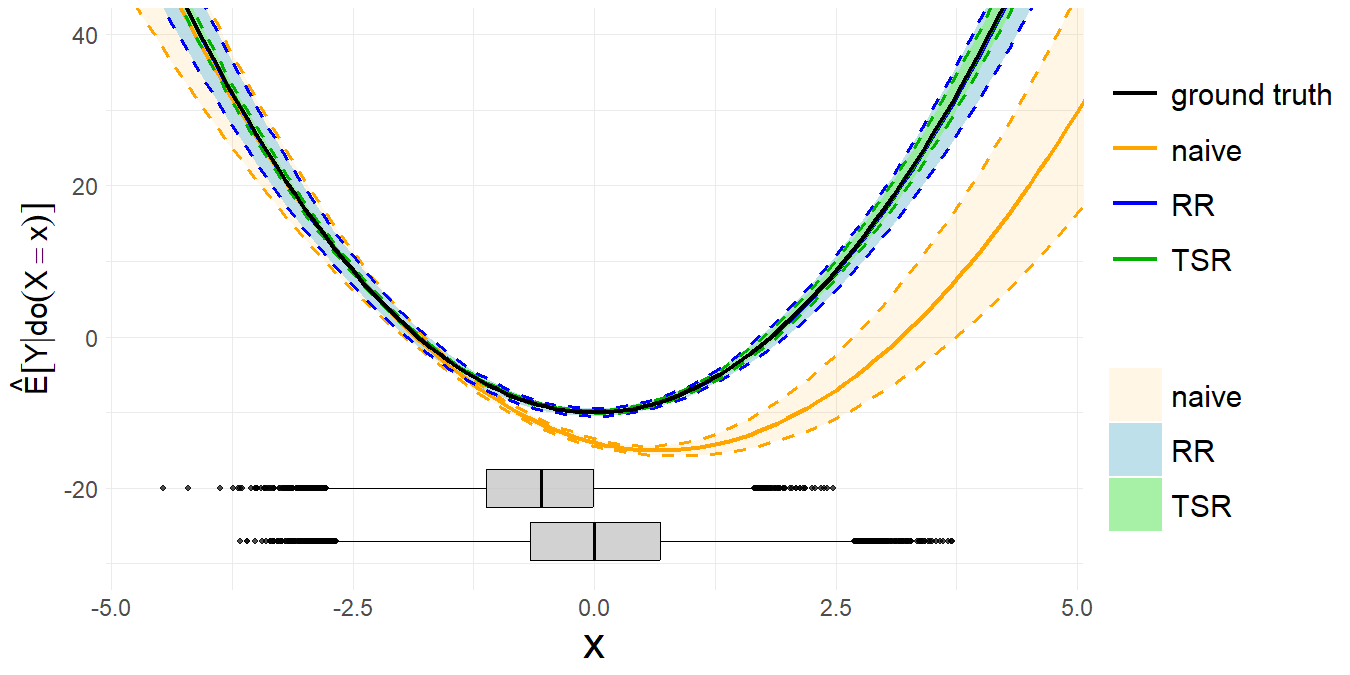}
      \vspace{-0.1em}
      \caption*{Estimates of $E[Y|do(X=x)]$}
   \end{minipage}
   \caption{Left: Boxplots of the MSE over $\mathcal{D}$ for $\mathcal{S}\subset\mathcal{D}$, and $\mathcal{S}\cap\mathcal{D}=\emptyset$ of RR and TSR for $n\in\{500,1000,5000\}$ for the quadratic case. On the right, we show in black the ground truth value of $E[Y|do(X=x)]$, and colored the mean estimates and $95\%$-areas over all simulations for naive, RR, and TSR for $n=500$ with the coefficients fixed at their means (for $\mathcal{S}\cap\mathcal{D}=\emptyset$). We see that RR and TSR recover the ground truth whereas the $95\%$-area of TSR more narrowly contains the ground truth. The boxplots on the x-axis show the distribution of $\Xv$ in $\mathcal{S}$ (upper) and $\mathcal{D}$ (lower).}
   \label{fig:QuadVarD}
\end{figure}

First, in \Cref{sec:expvariance}, we confirm our results for the variance comparison of TSR and RR from \Cref{sec:biasvariance}. Then, in \Cref{sec:expvarious}, we look at several examples with confounding, for which RR is not applicable. Subsequently, we will evaluate the performance of TSR in a more challenging setting based on \mbox{\Cref{fig:vierDAGs} (d)}, in \Cref{sec:expfigure1d}, and investigate the robustness of TSR to certain types of misspecification, introduced by an unobserved latent variable $U$, in \Cref{sec:exprobust}. Finally, we will substantiate our previous results based on semi-synthetic data in \Cref{sec:exprealworld}.

\subsection{Empirical Variance Evaluation}
\label{sec:expvariance}
As discussed in \Cref{sec:biasvariance}, the variance of TSR is at most of the same magnitude as of RR. The result was proven only for $\mathcal{S}\cap\mathcal{D}=\emptyset$. Here, we simulate data according to both settings, $\mathcal{S}\subset\mathcal{D}$ and $\mathcal{S}\cap\mathcal{D}=\emptyset$. %
We generate data according to the DAG shown in \Cref{fig:vierDAGs} (a), as $\Yv_{\text{lin}} := \mu_0+a_1\Xv+a_2{\Xv}^2+b_1\Zpv+b_2(\Zpv)^2+\varepsilon_Y$ with selection indicator $\Sv := \mathbf{1}(\Xv+\Zpv< -2)$ %
with $\Xv,\varepsilon_Y\sim \mathcal{N}(0,1)$ and $\Zpv\sim\mathcal{N}(-2,1)$. Further, we draw  
$a_1\sim \text{Unif}(2.5,3.5)$, $b_1\sim \text{Unif}(4.5,5.5)$, and $\mu_0,a_2,b_2\sim \text{Unif}(-0.5,0.5)$ and frame it the linear case as the coefficients of the quadratic terms have mean zero. For the quadratic case, we exchange the distributions of $a_1$ and $a_2$ as well as $b_1$ and $b_2$.
Computing TSR and RR, we include basis functions up to degree 2. We chose the sample size $n$ to be $500,1000$ and $5000$. 

The results for the quadratic model in \Cref{fig:QuadVarD} show that the mean MSE of TSR is consistently lower than for RR, whereas the difference vanishes when increasing the sample size $n$ (details are provided in \Cref{tab:var_quadratic} in Appendix~\ref{sec:appexpvariance}), validating our theoretical results. The plot on the right shows the central $95\%$-area of the estimates of $E[\Yv|do(\Xv)]$ from 100 simulation runs, fixing $\mu_0,a_1,a_2,b_1,b_2$ at their means (for $\mathcal{S}\cap\mathcal{D}=\emptyset$). We see that the estimates for TSR are more concentrated around its mean than for RR.
As expected, the naive estimator is systematically biased.

We provide all numerical results, and compare the errors on test data sampled from $\mathcal{S}$ to data from $\mathcal{D}$, %
in both scenarios ($\mathcal{S}\cap\mathcal{D} = \emptyset$, and $\mathcal{S}\subset\mathcal{D}$) in Appendix~\ref{sec:appexpvariance}. 
Overall, the results shown in Appendix~\ref{sec:appexpvariance} confirm the conclusions drawn from this section.

\subsection{Simulations with Selection Bias and Confounding}
\label{sec:expvarious}

\begin{figure}[t]\centering
  \begin{minipage}[t]{0.26\linewidth}
    \centering
    \vspace{-4.5cm}
    \scalebox{0.5}{
    \begin{tikzpicture}[
        >={Stealth[round]}, %
        node distance=1cm, %
        every node/.style={circle, inner sep=-3pt, draw, minimum size=0.75cm} %
    ]
        \node (X) {X};
        \node (Y) [right=of X] {Y};
        \node (S) [below=of X,accepting] {S};
        \node (Z) [below=of Y] {\small$Z^+$};

        \draw[->] (X) -- (S);
        \draw[->] (Z) -- (S);
        \draw[->] (X) -- (Y);
        \draw[->] (Z) -- (Y);
        \draw[->] (Z) -- (X);
        \node (Q) [below=-0.4cm of S, draw=none] {};

    \end{tikzpicture}\hspace{0.5cm}
    \begin{tikzpicture}[
        >={Stealth[round]}, %
        node distance=1cm, %
        every node/.style={circle, inner sep=-3pt, draw, minimum size=0.75cm} %
    ]   
        \node (X) {X};
        \node (Y) [right=of X] {Y};
        \node (S) [below=of X,accepting] {S};
        \node (A) [below=of Y] {\small$Z_A^+$};
        \node (W) [above=0.4cm of $(X)!0.5!(Y)$]{\small$Z_B^+$};
        \node (Q) [below=-0.4cm of S, draw=none] {};

        \draw[->] (X) -- (S);
        \draw[->] (A) -- (S);
        \draw[->] (X) -- (Y);
        \draw[->] (A) -- (Y);
        \draw[->] (W) -- (X);
        \draw[->] (W) -- (Y);
    \end{tikzpicture}}\\\vspace{-0.3cm}
    \scalebox{0.6}{(a) Ex. 1 \& 2}\hspace{0.7cm}\scalebox{0.6}{(b) Ex. 3 \& 4}\vspace{0.2cm}

       \scalebox{0.5}{
    \begin{tikzpicture}[
        >={Stealth[round]}, %
        every node/.style={circle, inner sep=-3pt, draw, minimum size=0.75cm} %
        ]

        \node (X) {X};
        \node (Y) [right=of X] {Y};
        \node (S) [below=of X,accepting] {S};
        \node (A) [below=of Y] {\small$Z^-$};
        \node (W) [above=0.4cm of $(X)!0.5!(Y)$]{\small$Z^+$};
        \node (Q) [below=-0.4cm of S, draw=none] {};

        \draw[->] (X) -- (S);
        \draw[->] (A) -- (S);
        \draw[->] (X) -- (Y);
        \draw[->] (A) -- (Y);
        \draw[->] (W) -- (X);
        \draw[->] (W) -- (Y);
        \draw[->] (X) -- (A);
        
    \end{tikzpicture}}\\\vspace{-0.3cm}\scalebox{0.6}{(c) Ex. 5 \& 6}\vspace{0.2cm}
  \end{minipage}
   \begin{minipage}[b]{.22\linewidth}
      \includegraphics[width=1\linewidth]{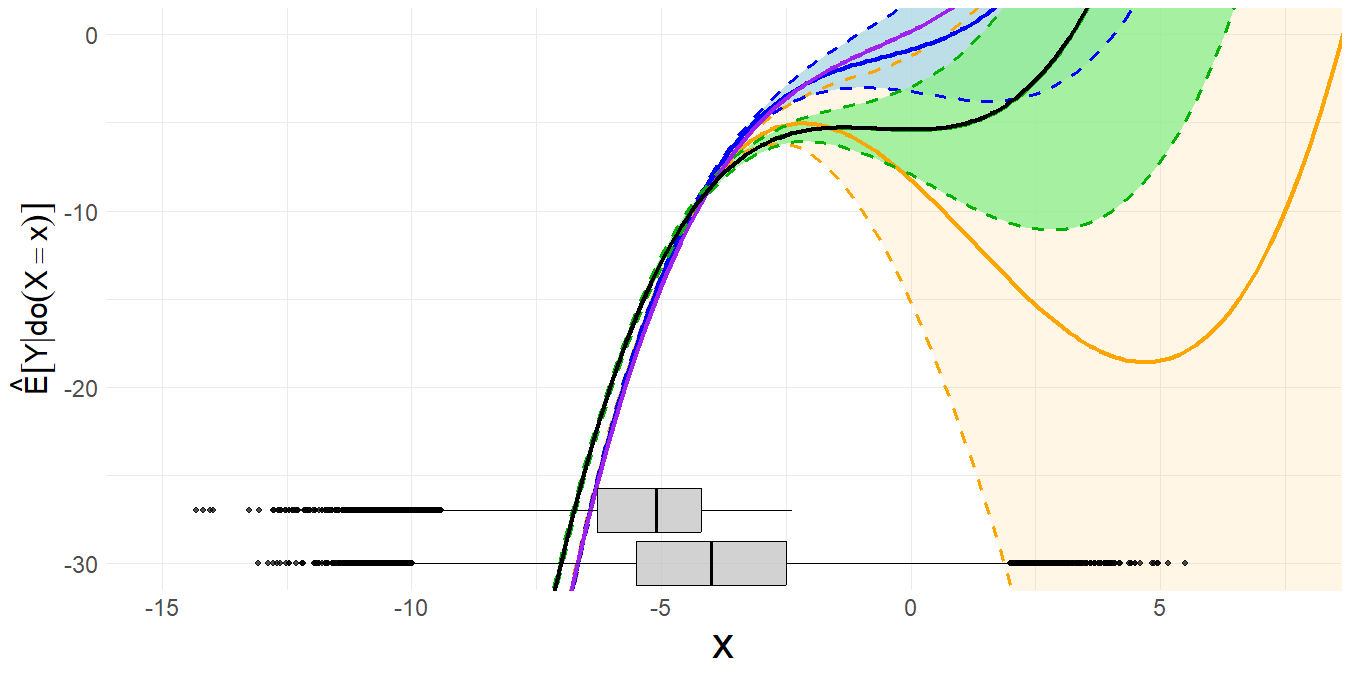}
      \caption*{Ex.~1}
      \includegraphics[width=1\linewidth]{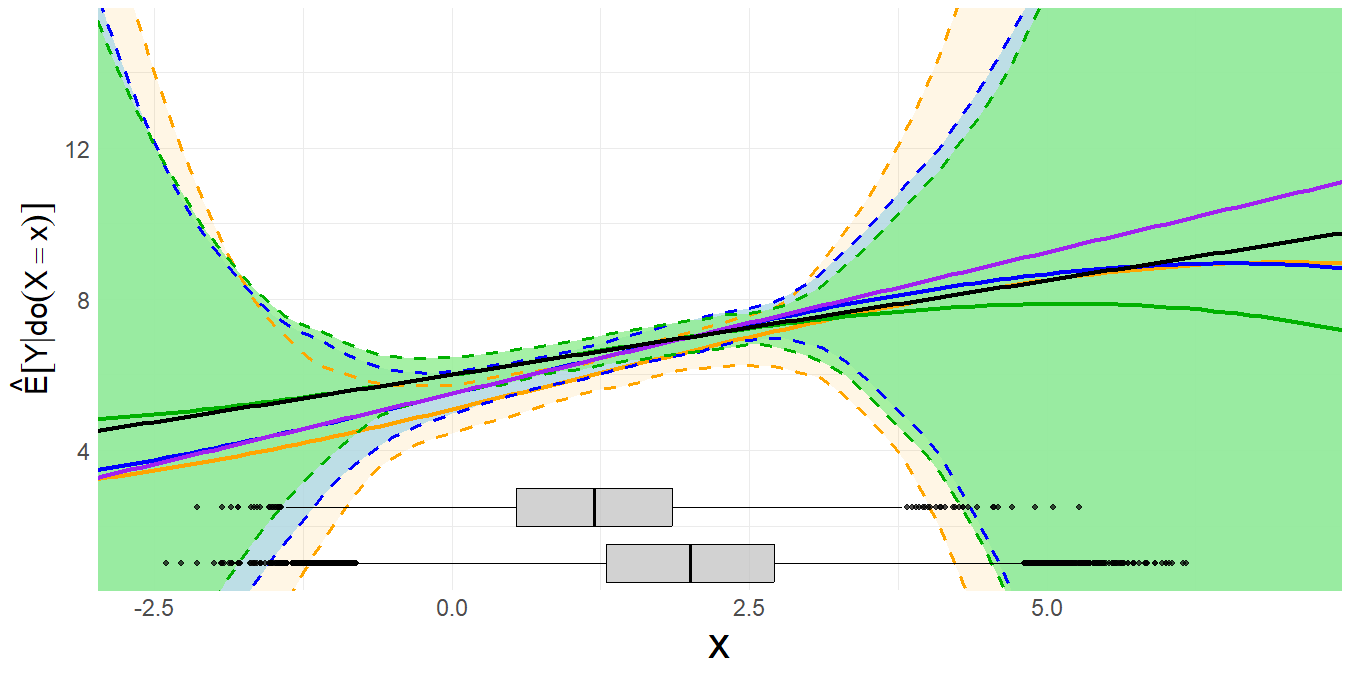}
      \caption*{Ex.~4}
   \end{minipage}%
   \hfill
   \begin{minipage}[b]{.22\linewidth} 
      \includegraphics[width=1\linewidth]{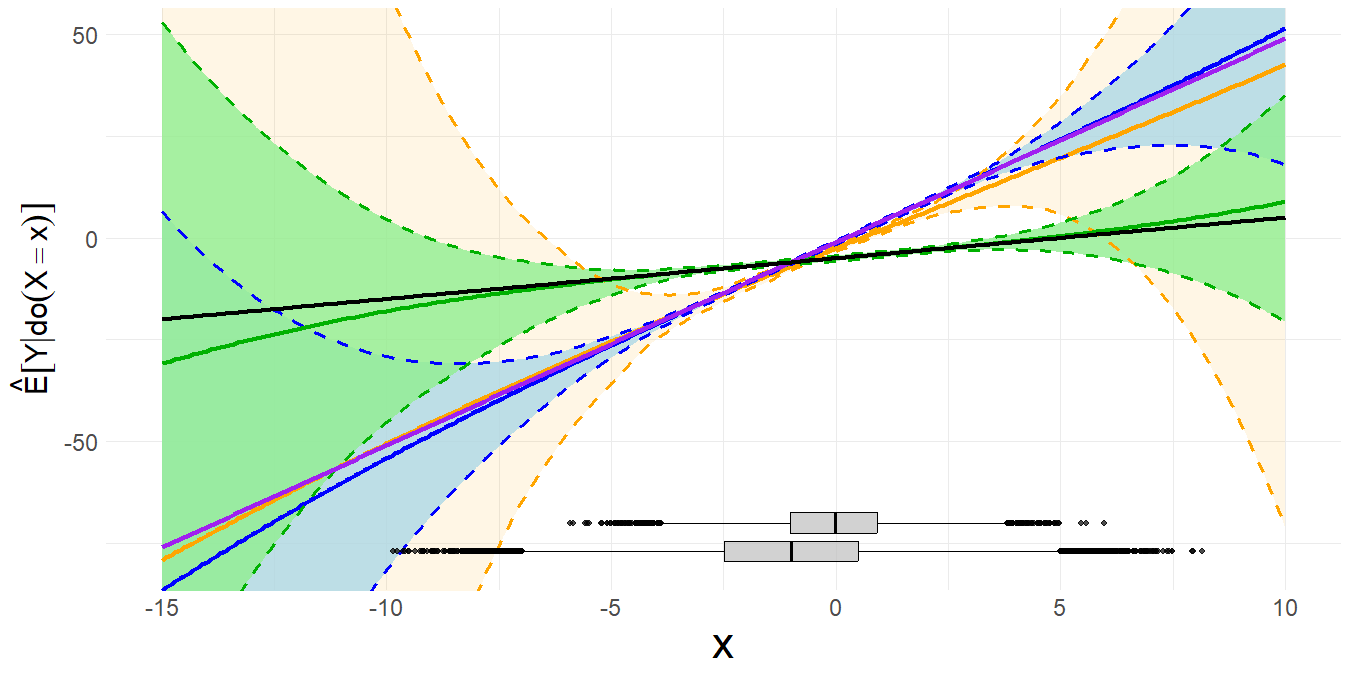}
      \caption*{Ex.~2}
      \includegraphics[width=1\linewidth]{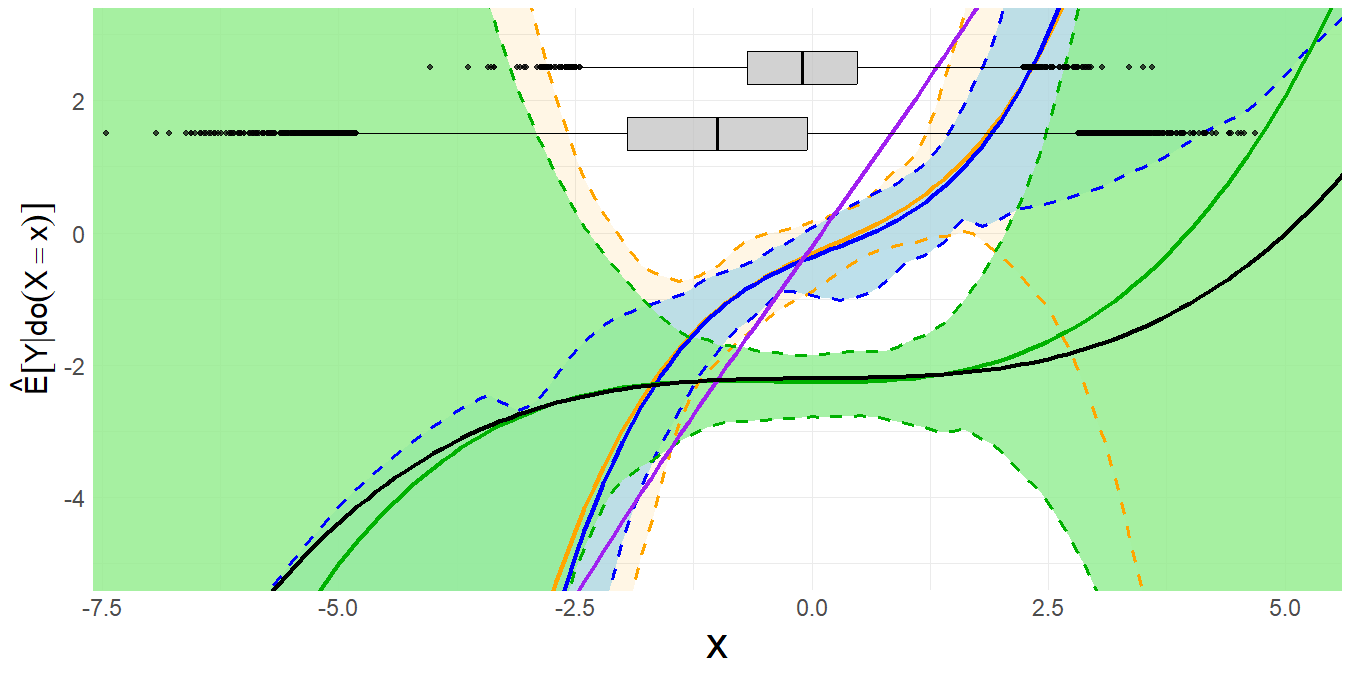}
      \caption*{Ex.~5}
   \end{minipage}%
   \hfill
   \begin{minipage}[b]{.22\linewidth} 
      \includegraphics[width=1\linewidth]{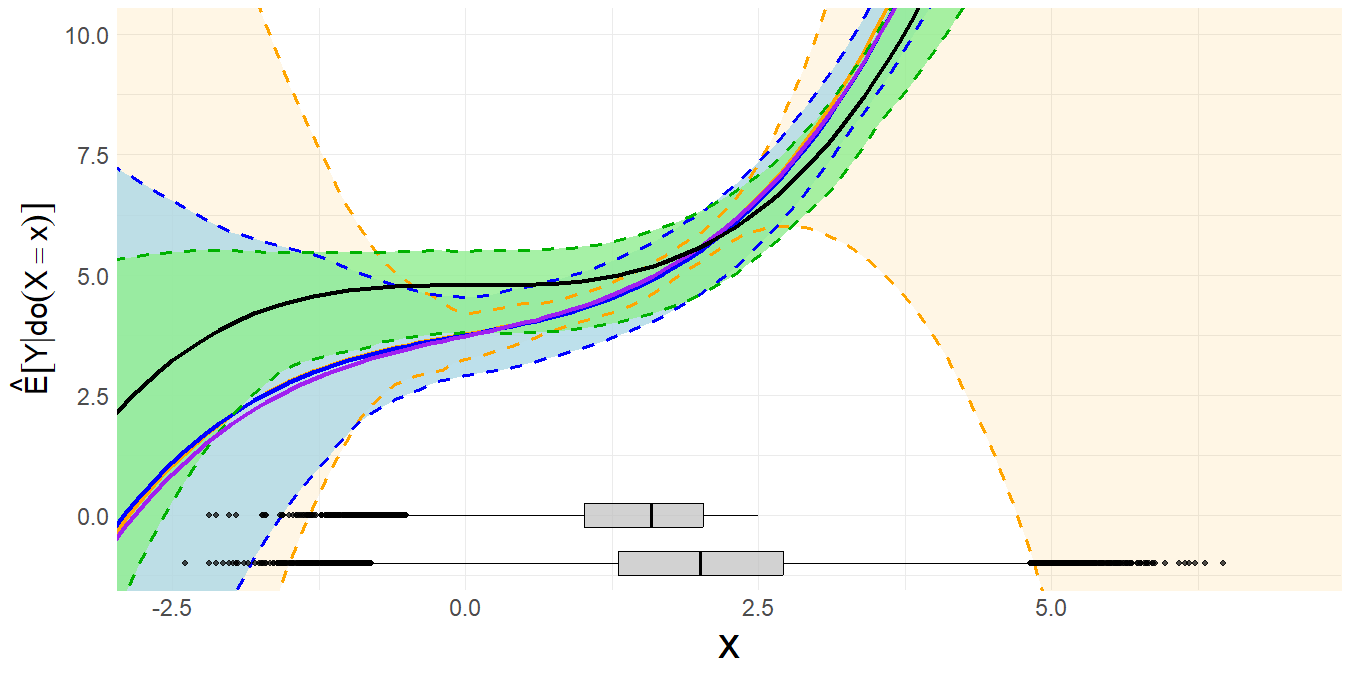}
      \caption*{Ex.~3}
        \includegraphics[width=1\linewidth]{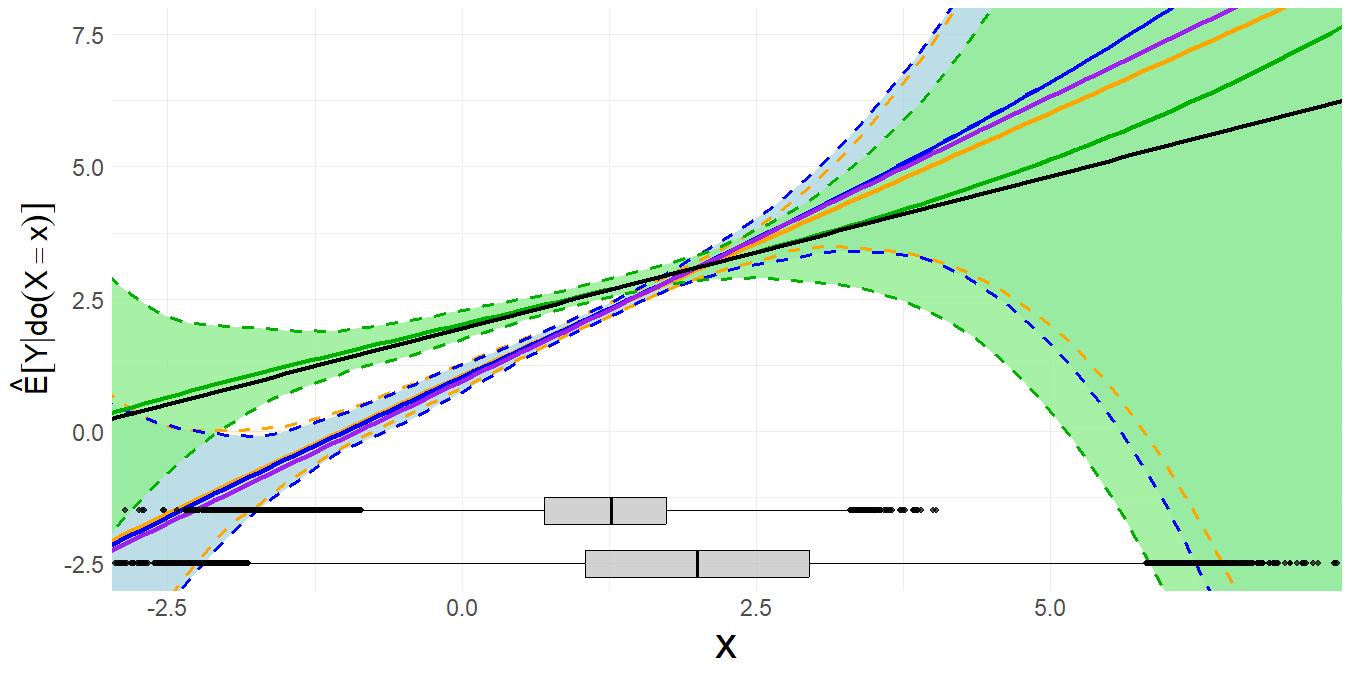}
      \caption*{Ex.~6}
   \end{minipage}
   \begin{minipage}[b]{.06\linewidth}
      \includegraphics[scale=0.11]{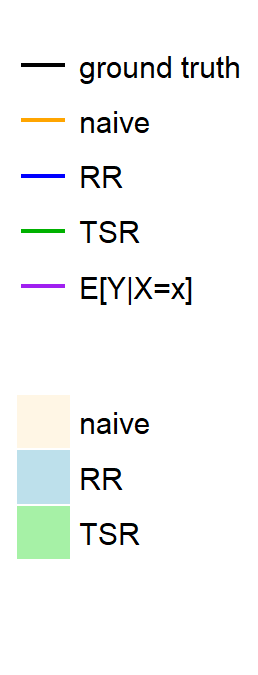}
      \includegraphics[scale=0.11]{Fig4d_4h_10d_11d_12d_13d_14d_15d.png}
      \caption*{}
   \end{minipage}\\
    \caption{Left: DAGs representing the DGPs of Ex.~1-6. On the right, we show in black the ground truth value of $E[Y|do(X=x)]$, and colored the mean estimates and $95\%$-areas over all simulations for naive, RR, and TSR for $n=1000$ and with the coefficients fixed at their means (for $\mathcal{S}\cap\mathcal{D}=\emptyset$) for Examples 1-6, respectively. As $E[Y\mid Y]\neq E[Y\mid do(X)]$ in these examples, we also show the true underlying $E[Y|X=x]$ in purple. We see that TSR recovers the ground truth $E[Y\mid do(X)]$, whereas RR recovers $E[Y\mid X]$. The boxplots on the x-axis show the distribution of $\Xv$ in $\mathcal{S}$ (upper) and $\mathcal{D}$ (lower).}
   \label{fig:expvariousDAGs}
\end{figure}

Next, we consider in total six distinct generating mechanisms, illustrated by the graphs in Figure~\ref{fig:expvariousDAGs}, that include confounding variables. Hence, RR is not applicable for estimating the causal effect since \mbox{$E[\Yv\mid do(\Xv)]\neq E[\Yv\mid \Xv]$}, but should be able to recover $E[Y\mid X]$. The details for the data generating processes are provided in \Cref{sec:appexpvarious}, where for each graph, we provide a linear and cubic generative mechanism. This means, in analogy to Section~\ref{sec:expvariance}, that we include impacts up to degree 3, but for the linear case, the coefficients for the quadratic and cubic impacts have mean zero. Corollary~\ref{corollary:ZminusgivendoX} ensures that no integral has to be computed for TSR in the setting from \mbox{Figure~\ref{fig:expvariousDAGs} (c)}. For Ex. 1 \& 2, TSR is given by \Cref{eq:TSRex12}. For Ex. 3 \& 4, an additional summand for a second element in $\Zpv$ is included, and for Ex. 5 \& 6, we use \Cref{eq:estimator-unobserved}.

We included regressors up to degree 3, to cover the generating mechanisms. In \Cref{fig:expvariousDAGs}, we plot the empirical central $95\%$ confidence intervals for TSR, RR, with OLS regression and the naive baseline for the case of $n=1000$ with $\mathcal{S}\cap \mathcal{D}=\emptyset$ for each setting, fixing the coefficients to their means. As expected, RR recovers $E[Y\mid X]$ well, while TSR is able to additionally correct for the confounders, thus accurately recovering $E[Y\mid do(X)]$ for all cases.
We further report the results on $\mathcal{D}$ and $\mathcal{S}$ for OLS, and ridge regression for both TSR, and RR with increasing sample size in \Cref{sec:appexpvarious}. Additionally, we report the numerical results of the errors on $\mathcal{S}$ and $\mathcal{D}$ for the randomized coefficients. In most cases, ridge regression matches the performance of OLS. In Example~1, however, we observe that ridge regression introduces a bias for both TSR and RR, suggesting that, in regions with low support (at the borders), the $95\%$-areas do not include $E[\Yv|do(\Xv)]$ for TSR and $E[\Yv|\Xv]$ for RR, respectively. As the naive estimator aims to estimate $E[Y\mid X]$, the results are more similar to RR than TSR. In Example~3 and Example~5 the naive estimation even performs well for $E[Y\mid X]$, but not for the causal effect $E[Y\mid do(X)]$ suggesting, that in these cases, the bias due to confounding is stronger than the selection mechanism. This experiment showcases the importance to adjust for both confounding and selection bias.

\subsection{Simulations with Selection Bias and Unobserved Confounding}
\label{sec:expfigure1d}

Next, we consider a case with \emph{unobserved} confounding. The data generating process is based on \Cref{fig:vierDAGs} (d), %
the graph compatible with the motivating example in \Cref{fig:datasetsandrunning}:
 \begin{align*}
 &\Zpv:=2U+\varepsilon_{Z^+} \;\;\;\;\;  \;\;\;\;\;\;\; \,  \Xv:=\Zpv+\varepsilon_X\;\;\;\;\;\;\;\;\;\;\;\;\;\;\;\;\;\;U,\varepsilon_{Z^+},\varepsilon_{Z^-},\varepsilon_X,\varepsilon_Y\sim\mathcal{N}(0,1)\\
&\Zmv:=\Xv+2U+2\varepsilon_{Z^-}  \;\;\; \,  S:=\mathbf{1}(\Xv+\Zmv>5)    \;\;\;\;
\;\;\;\,\Yv:=0.5\Xv^2+2\Zmv+2U+3\varepsilon_Y.
 \end{align*}
As outlined in \Cref{sec:our-estimator}, in this case, TSR is given by \Cref{eq:estimator-unobserved},
where we restrict %
ourselves notational to the linear case. The $\beta$ coefficients are estimated by OLS from $\mathcal{S}$, $\gamma$ coefficients from $\mathcal{D}$. %
Further, $\widehat{E}[\Zpv]$ is the empirical mean of $\Zpv$ in $\mathcal{D}$. Akin to the previous experiments, we use OLS %
regression and optionally add a ridge penalty. %
We explicitly add the penalty also to the second step since $\Xv$ and $\Zpv$ are correlated. 

\begin{figure}[t!]
\centering
   \begin{minipage}[b]{.4\linewidth}
      \includegraphics[scale=0.175]{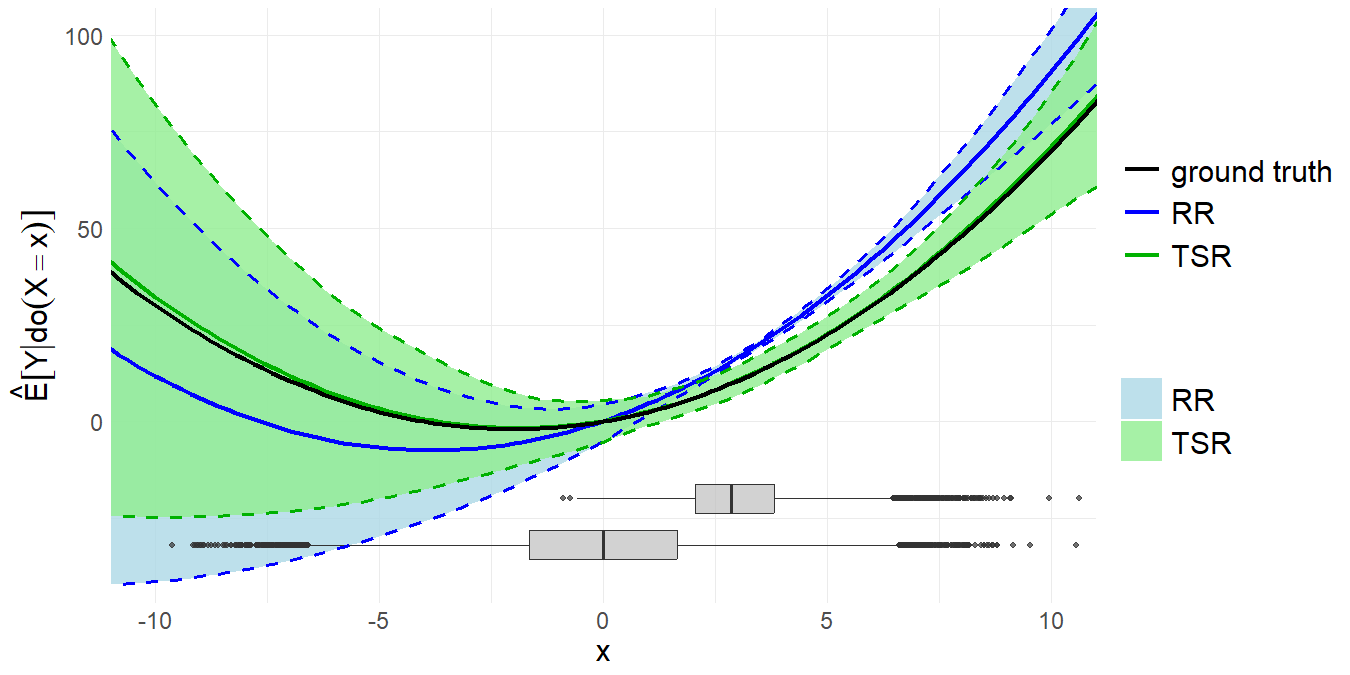}
   \end{minipage}
   \hspace{.01\linewidth}
   \begin{minipage}[b]{.4\linewidth} 
      \includegraphics[scale=0.175]{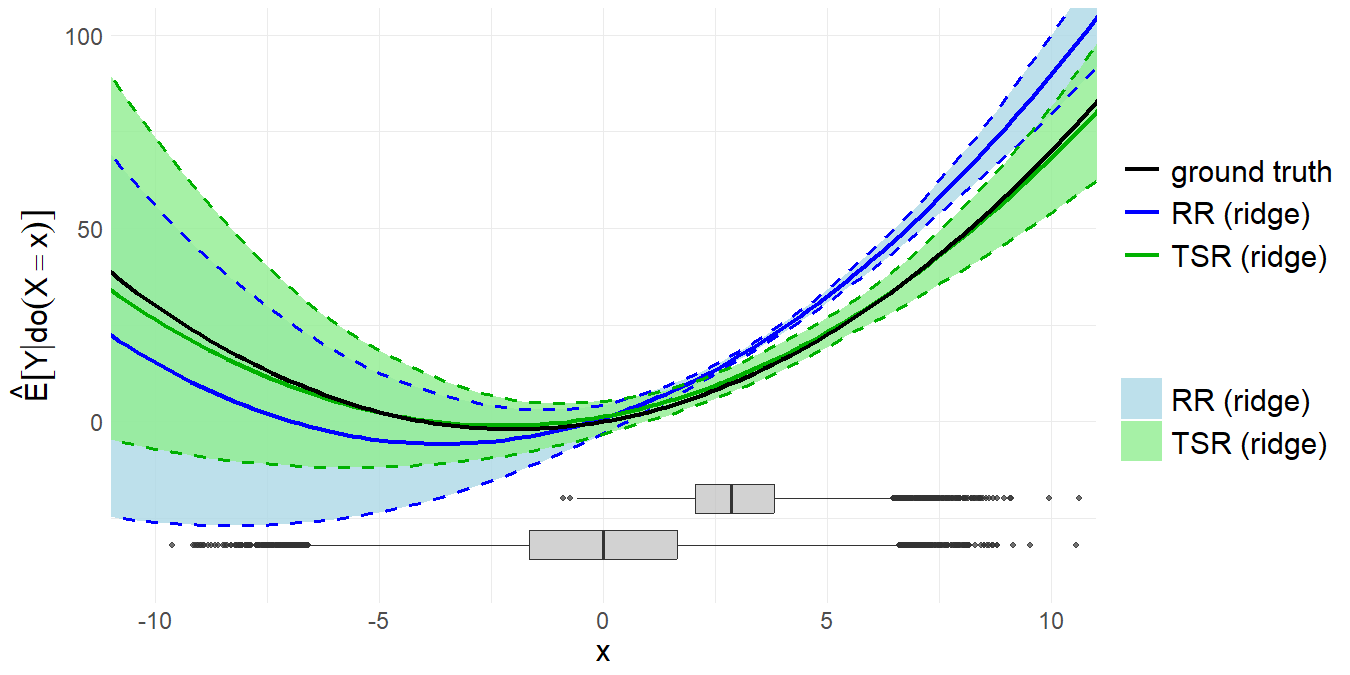}
   \end{minipage}
\caption{We show in black the ground truth value of $E[Y\mid do(X=x)]$ and colored the mean estimates and $95\%$-areas over all simulations for RR and TSR via OLS (left) and ridge regression (right) for $n=500$ (for $\mathcal{S}\cap\mathcal{D}=\emptyset$). We see that TSR recovers the ground truth whereas RR does not. The boxplots on the x-axis show the distribution of $\Xv$ in $\mathcal{S}$ (upper) and $\mathcal{D}$ (lower).}
\label{fig:integraln500}
\end{figure}

We show the results for $n=500$ in Figure~\ref{fig:integraln500}, where we observe that TSR is able to recover the ground truth. We additionally show RR as a baseline, but note that this setting violates its underlying assumptions. Hence, it is expected that it does not recover the ground truth causal effect. In addition, we observe that the confidence intervals for ridge are slightly smaller than for OLS, while a small bias is introduced. We repeat the experiment for $n=2000$, for which we show the results in \Cref{sec:appexpfigure1d}, where we observe that the difference between OLS and ridge is not evident anymore.

\subsection{Robustness to Misspecification}
\label{sec:exprobust}
\begin{figure}[t!]
  \centering
    \begin{minipage}[t]{0.38\linewidth}
    \centering
    \scalebox{0.3}{
\begin{tikzpicture}[
    >={Stealth[round]}, %
    node distance=1cm, %
    every node/.style={circle, inner sep=-3pt, draw, minimum size=0.75cm} %
]   
    \node[thick] (X) at (0,2) {X};
    \node[thick] (Y) at (2,2) {Y};
    \node[thick,accepting] (S) at (0,0) {S};
    \node[thick] (Z^+) at (2,0) {\small$Z^+$};
    \node[thick,blue,fill=lightgray] (U) at (1,1) {U};
    \draw[->,thick] (X) -- (Y);
    \draw[->,thick] (X) -- (S);
    \draw[->,thick] (Z^+) -- (Y);
    \draw[->,thick] (Z^+) -- (S);
    \draw[->,thick,blue] (U) -- (Y);
\end{tikzpicture}}
  \end{minipage}
  \hfill%
  \begin{minipage}[t]{0.05\linewidth}
    \centering
    \scalebox{0.3}{
\begin{tikzpicture}[
    >={Stealth[round]}, %
    node distance=1cm, %
    every node/.style={circle, inner sep=-3pt, draw, minimum size=0.75cm} %
]
    \node[thick] (X) at (0,2) {X};
    \node[thick] (Y) at (2,2) {Y};
    \node[thick,accepting] (S) at (0,0) {S};
    \node[thick] (Z^+) at (2,0) {\small$Z^+$};
    \node[thick,blue,fill=lightgray] (U) at (1,1) {U};
    \draw[->,thick] (X) -- (Y);
    \draw[->,thick] (X) -- (S);
    \draw[->,thick] (Z^+) -- (Y);
    \draw[->,thick] (Z^+) -- (S);
    \draw[->,thick,blue] (U) -- (S);

\end{tikzpicture}}
  \end{minipage}
  \hfill%
  \begin{minipage}[t]{0.38\linewidth}
    \centering
    \scalebox{0.3}{
\begin{tikzpicture}[
    >={Stealth[round]}, %
    node distance=1cm, %
    every node/.style={circle, inner sep=-3pt, draw, minimum size=0.75cm} %
]   
    \node[thick] (X) at (0,2) {X};
    \node[thick] (Y) at (2,2) {Y};
    \node[thick,accepting] (S) at (0,0) {S};
    \node[thick] (Z^+) at (2,0) {\small$Z^+$};
    \node[thick,blue,fill=lightgray] (U) at (1,1) {U};
    \draw[->,thick] (X) -- (Y);
    \draw[->,thick] (X) -- (S);
    \draw[->,thick] (Z^+) -- (Y);
    \draw[->,thick] (Z^+) -- (S);
    \draw[->,thick,blue] (U) -- (Y);
    \draw[->,thick,blue] (U) -- (S);
\end{tikzpicture}}
  \end{minipage}

\centering

   \begin{minipage}[b]{.29\linewidth} 
\includegraphics[scale=0.15]{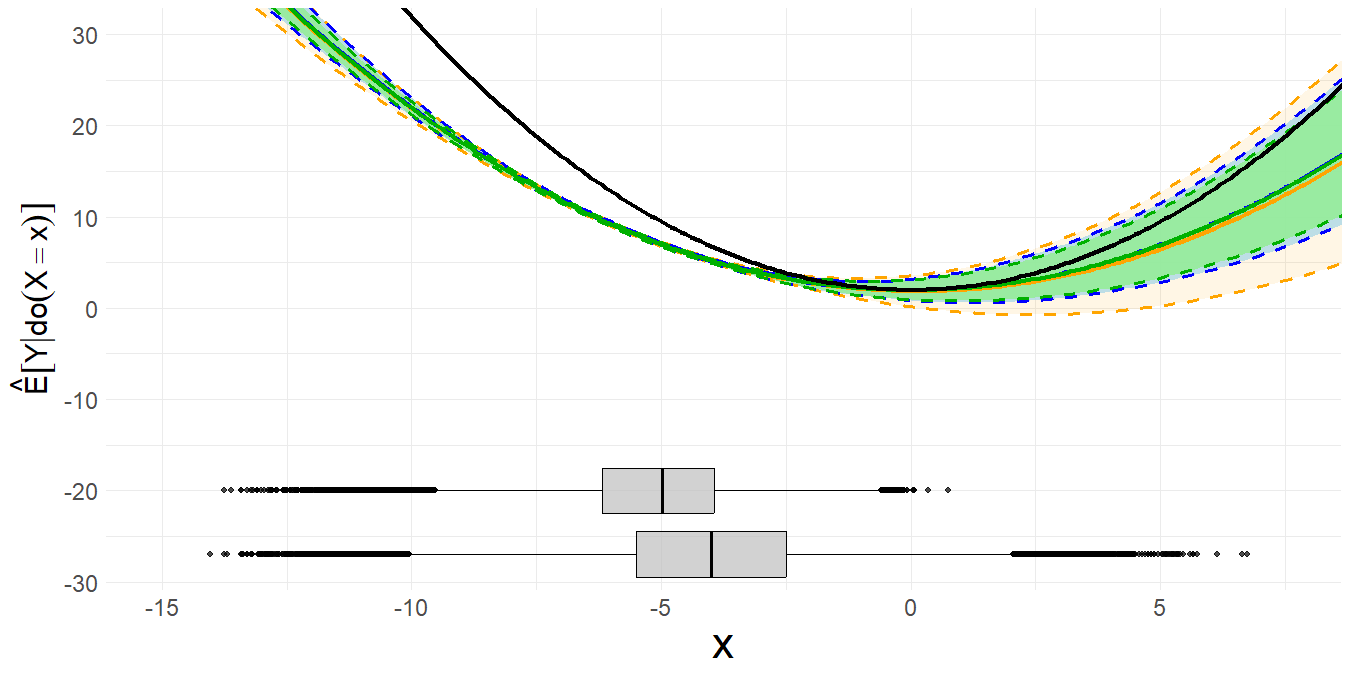}
   \end{minipage}\hspace{0.14cm}
   \begin{minipage}[b]{.29\linewidth} 
\includegraphics[scale=0.15]{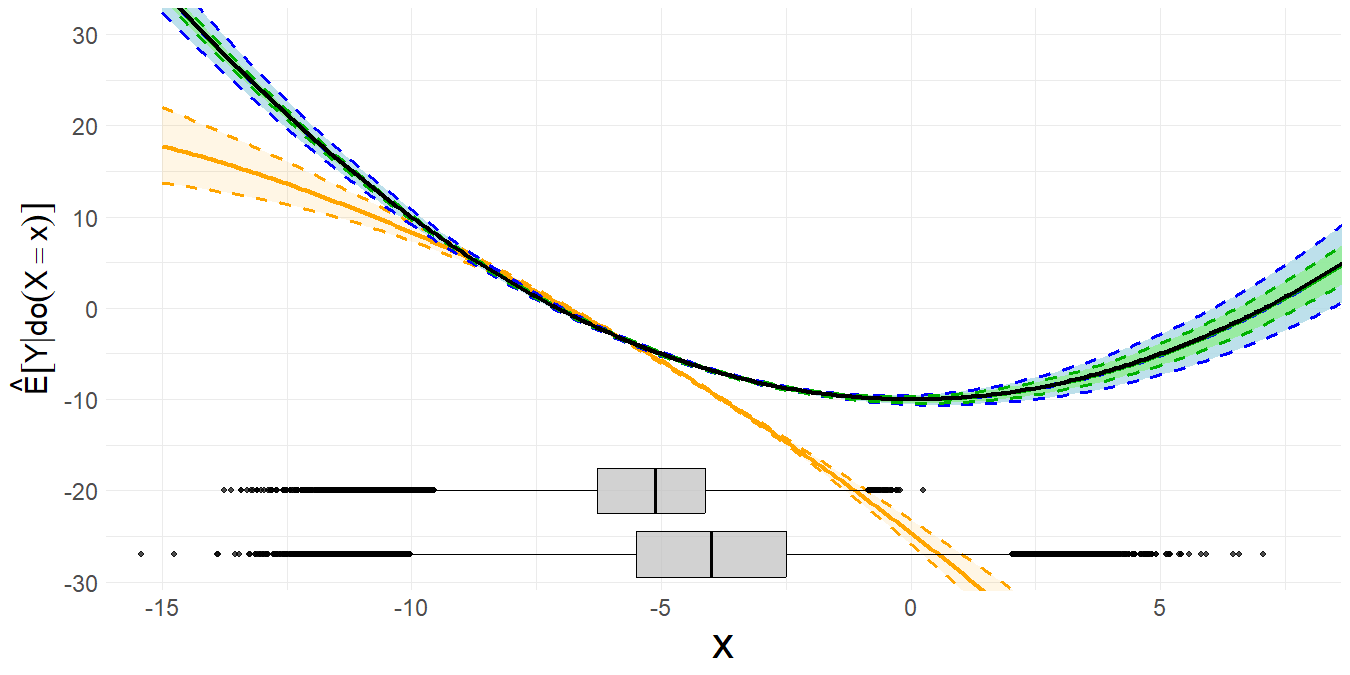}
   \end{minipage}\hspace{0.14cm}
   \begin{minipage}[b]{.29\linewidth} 
\includegraphics[scale=0.15]{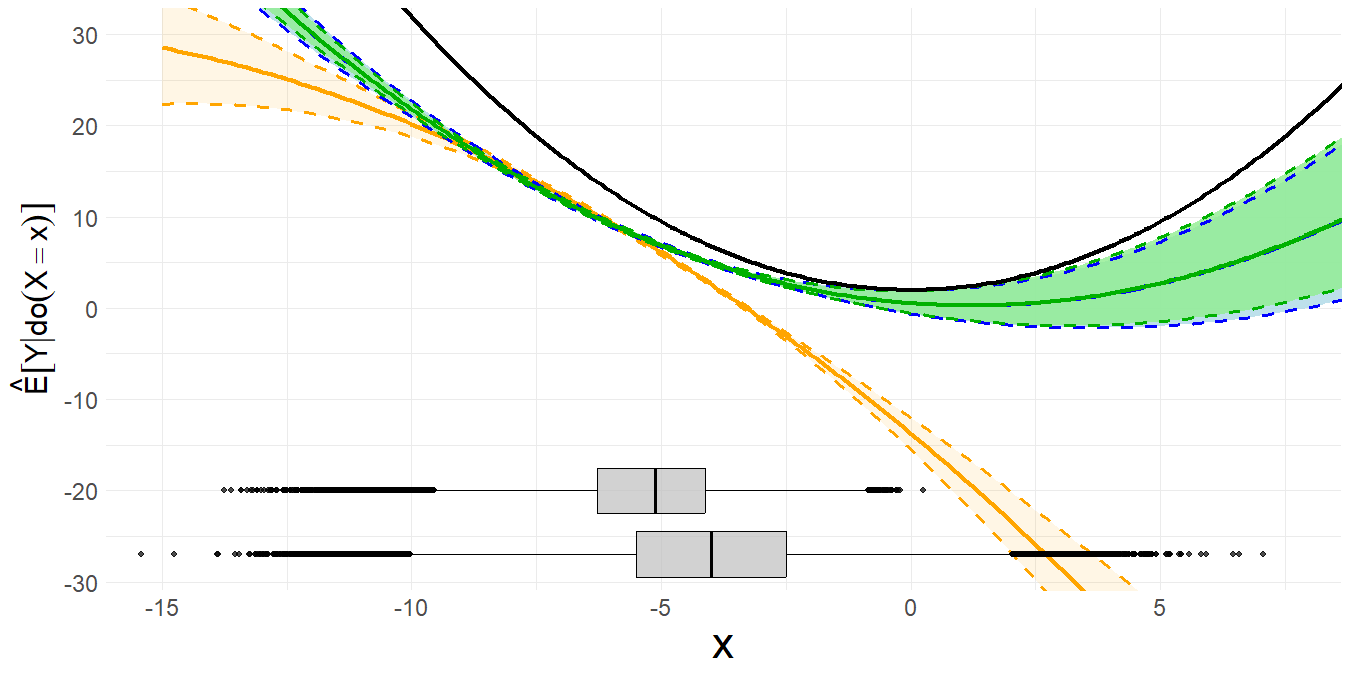}
   \end{minipage}\hspace{0.14cm}
   \begin{minipage}[b]{.05\linewidth}
\includegraphics[scale=0.15]{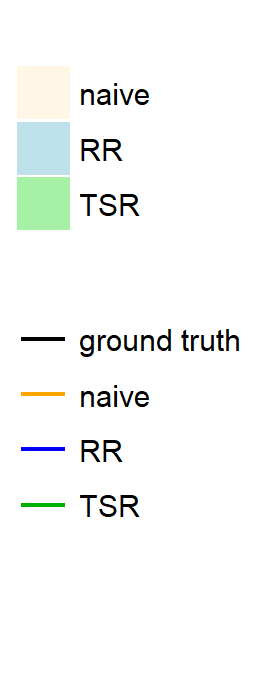}
   \end{minipage}\\

 \caption{We show in black the ground truth value of $E[Y|do(X=x)]$, and colored the mean estimates and $95\%$-areas over all simulations for naive, RR, and TSR in absence of $U$, introducing $U\rightarrow Y$, $S\leftarrow U$ or $S\leftarrow U\rightarrow Y$ (type 1) for $n=5000$ (for $\mathcal{S}\cap\mathcal{D}=\emptyset$). The boxplots on the x-axis show the distribution of $\Xv$ in $\mathcal{S}$ (upper) and $\mathcal{D}$ (lower).}   
\label{fig:dagsandplots}
\end{figure}

Last, we investigate the robustness of TSR to certain types of misspecification which is introduced by an unobserved variable $U$, %
in \Cref{fig:dagsandplots}, where we stick to the scenario in which %
$\mathcal{S}\cap\mathcal{D}=\emptyset$. In case (a) $U$ is a cause of $Y$, in case (b) $U$ is a cause of $S$ and in case (c) $U$ is a confounder between $S$ and $Y$.

 For case (a), our identifiability assumptions are still met (Theorem~\ref{theorem:neutheoretisch} %
 holds), but the missing regressor introduces %
 bias %
 (i.e.~Assumption~\ref{ass:linearity} is violated). Hence, in the first regression, we estimate the $\beta$ coefficients biased. In the second step, we just calculate $\overline{Z^+}$ and $\bar{U}$ cannot enter the estimation as we are not aware of $U$. %
 For case (b), our assumptions are still fulfilled, however, our proxy $Z^+$ is weaker since $S$ has an additional unobserved cause. Last, for case (c), the PMAR assumption is violated. Hence, in addition to the bias %
 through the missing regressor in (b), here the expression for $E[Y\mid do(X)]$ used to construct $\hat{\mu}_{TSR}(x)$ is erroneous. %
Details are given in Appendix~\ref{sec:appexprobust}.

Below we state how the DGP is extended for the three cases including $U\sim \mathcal{N}(3,1)$, respectively. For the edge $S\leftarrow U$, we distinguish between two types. %
\begin{enumerate}[label=(\alph*)]
    \item $U\rightarrow Y$ \hspace{2.9cm}
          $Y:=0.2X^2+5Z+4U+\varepsilon_Y$\vspace{0.1cm}
    \item $S\leftarrow U$ \hspace{2.99cm}
          $S:=\mathbf{1}(X+Z+0.1U<-5)$ \hspace{2.5cm}(type 1)\\
          \hspace*{4.17cm} $S:=\mathbf{1}(X+Z<-5) \mathbf{1}(U< 1.5)$ \hspace{1.82cm}(type 2)\vspace{0.1cm}
    \item $S\leftarrow U\rightarrow Y$\hspace{2.27cm}
          $S:=\mathbf{1}(X+Z+0.1U<-5)$ \hspace{2.5cm}(type 1)\\
          \hspace*{4.28cm}$S:=\mathbf{1}(X+Z<-5) \mathbf{1}(U< 1.5)$ \hspace{1.78cm} (type 2)\\
          \hspace*{4.24cm}$Y:=0.2X^2+5Z+4U+\varepsilon_Y$ 
\end{enumerate}          
We followed the procedure described in \Cref{sec:expvarious} and included covariates up to degree 2 in the regression model %
restricting ourselves to OLS regression.
The results, given in \Cref{fig:dagsandplots} for type 1 and $n=5000$ and \Cref{fig:rob_2} in \Cref{sec:appexprobust} for both types and $n\in\{1000,5000\}$ are consistent with %
our previous considerations. In case (a) a bias arises from the missing regressor. %
For case (b), TSR yields reliable results as our assumptions are met. %
Type 2 requires an increase of the sample size $n$ to compensate the decrease of sample size of the selected sample induced by %
$S\leftarrow U$. For case (c), the estimation gets biased due to the %
unobserved regressor. In addition, the specification of TSR is wrong as the derivation exploits PMAR. For cases (a) and (b), where their respective assumptions are satisfied, TSR again shows a smaller variance than RR. Overall, we still find that the corrected estimators outperform standard OLS regression on the selected sample. Further, it is encouraging to see that we can still achieve good estimates in scenario (b), where we miss some proxies for the selection mechanism.

\subsection{Semi-Synthetic Data}
\label{sec:exprealworld}
Whereas we have so far focused on synthetic data, we will now turn to semi-synthetic data. Since we require knowledge of the full SCM to artificially create confounding, experiments on real-world data are infeasible for many established datasets. %
Therefore, we resort to a semi-synthetic dataset satisfying $X \Indep Z$ that is consistent with \Cref{fig:vierDAGs} (a), where the mean of $Y$ is modelled as a linear combination of $X$ and $Z$, ensuring that the causal effect of $X$ on $Y$ is still the coefficient of $X$ in the model for the mean of $Y$. To introduce confounding we can simply add noise to both $X$ and $Y$.

\begin{figure}[t!]
\centering
\includegraphics[scale=0.14]{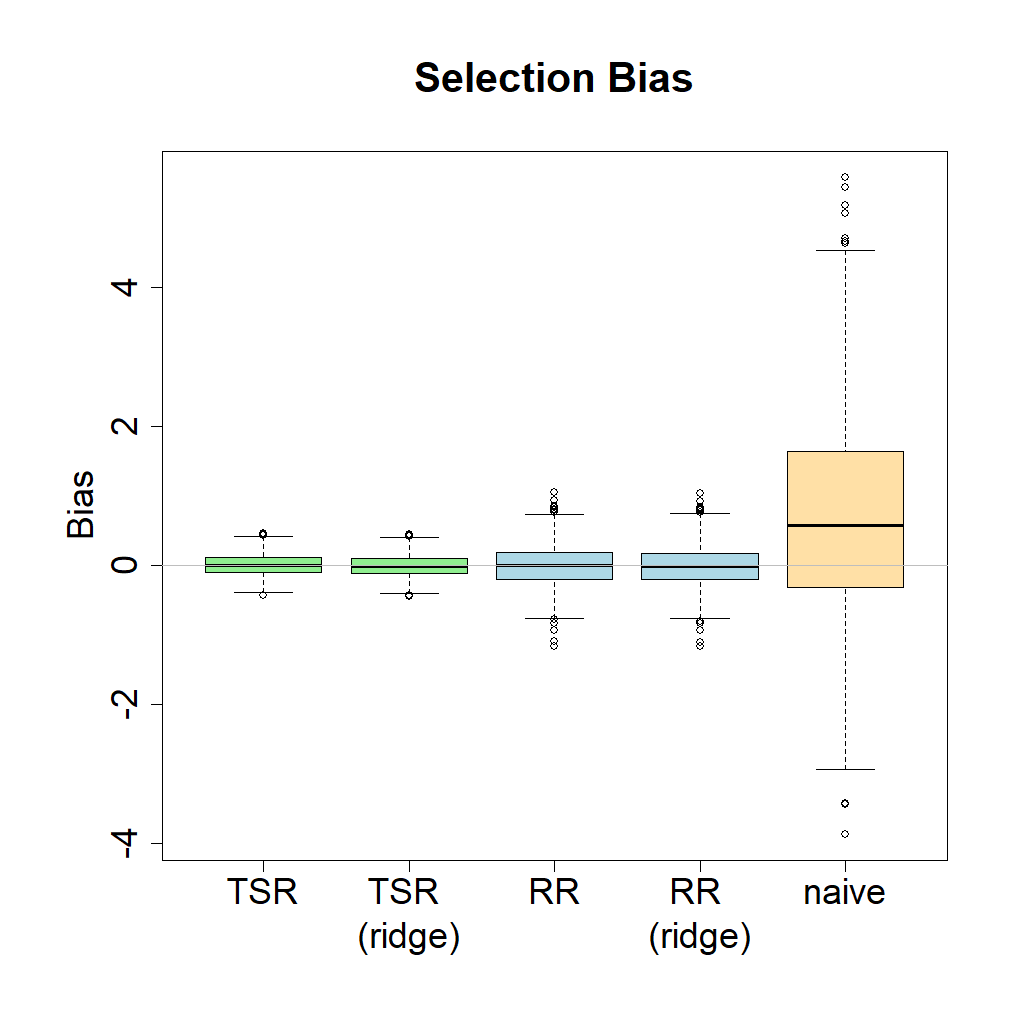}
\includegraphics[scale=0.14]{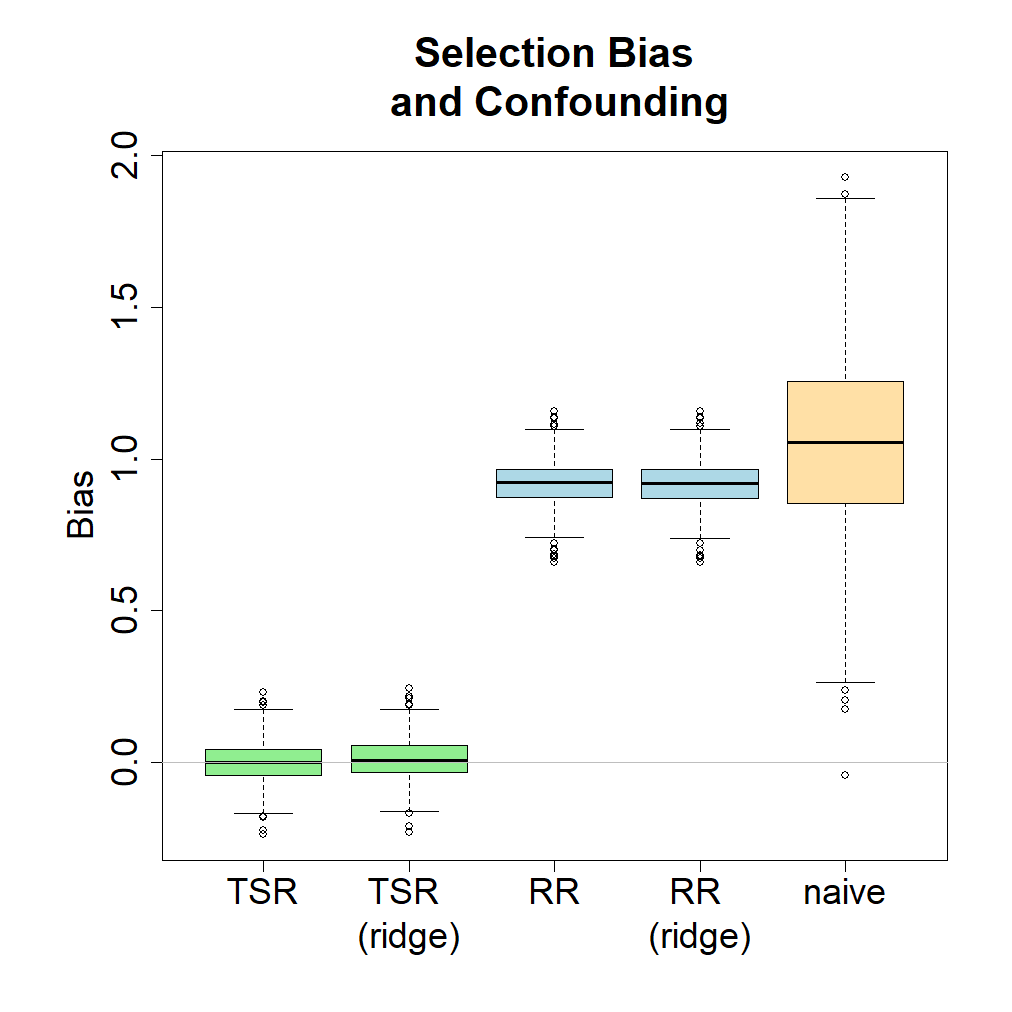}
\caption{We show the bias of TSR, RR and the naive estimator for the setting with selection bias, but without confounding (left) and combined with confounding (right) over $1000$ simulation runs. RR and TSR recover the causal effect in absence of confounding, whereas only TSR recovers the ground truth in presence of confounding.}
\label{fig:real-world}
\end{figure}

Inspired by \citet{hill:11:IHDP}, we use 6 continuous and 19 binary covariates from a randomized experiment, the Infant Health and Development Program (IHDP)\!\footnote{The data is available at \href{https://www.tandfonline.com/doi/suppl/10.1198/jcgs.2010.08162}{https://www.tandfonline.com/doi/suppl/10.1198/jcgs.2010.08162}} with sample size $n=985$, exploring the effect of home visits on cognitive test scores for infants, but simulate a randomized %
treatment $X$ sampled uniformly on $[-1,1]$ and generate the target variable $Y$ by modelling its mean %
as a linear combination of randomized discrete coefficients for $X$ and the 25 variables in $Z$. 
We induce selection bias and %
confounding between target and treatment. Details on the data generation and results are provided \Cref{sec:appexprealworld}. The results in \Cref{fig:real-world} based on 1000 simulation runs show, that RR and TSR recover the causal effect in the setting with selection bias, whereas only TSR recovers the causal effect in the setting with selection bias and confounding, matching our synthetic experiments. %

\section{Conclusion}
\label{sec:conclusion}
We considered estimation of the causal effect $E[Y|do(\Xv)]$ with continuous target $\Yv$ and treatments $\Xv$ under selection bias and confounding when having access to external data for $\Xv$ and $\Zv$ not underlying the selection mechanism. We derived conditions (Assumption~\ref{ass:new}) under which the causal effect is identifiable and s-recoverable (Theorem~\ref{theorem:neutheoretisch}). Assuming linearity with Gaussian errors, we proposed a generalized estimator, the Two-Step Regression (TSR) estimator, in line with our theoretical results. 
We discussed how TSR simplifies in different situations, e.g., when confounding is absent, and how to introduce non-linearity through low degree polynomials.
For a minimal example with uncorrelated $\Xv$ and proxies
$\Zv$, we derived the variance of TSR, proved that it is %
at most of the same magnitude as of repeated regression (RR)~\citep{boeken:23:privileged-info}, and confirmed this result through simulation studies based on synthetic and semi-synthetic data. Further, validating our estimator through extensive simulation studies, it became evident that an estimator capable of handling both selection bias and confounding is necessary because in wide ranges of the support of $\Xv$, even the centralized $95\%$-area of the estimates for $E[\Yv|\Xv]$ did not cover the underlying causal effect. %
Last, we found that, for some examples, we can trade-off some variance for the causal effect estimate by introducing a small bias when adding a ridge penalty to OLS when applying TSR and RR.

\textbf{Limitations and Future Work:} Although our estimator covers a range of different settings, we need to assume access to proxy variables. In \Cref{sec:exprobust}, we provide an experiment to evaluate slight violations of this assumption.
Another important assumption is access to external unbiased data for $\Xv$ and $\Zv$. This assumption may hold when the label is costly but the covariates can be accessed through other databases. As in the loan default example (\Cref{fig:datasetsandrunning}), information about job type, income, etc., may be accessible in other databases which do not contain measurements of the loan default. In other scenarios, however, this assumption may be restrictive (\Cref{sec:appassumptions}). However, for TSR, depending on the specific setting, observing $\Xv$, $\Zmv$ and $\Zpv$ separately unbiased may be sufficient. They only need to be observed jointly if variables appear together within an expectation term---such as $\Xv$ and $\Zmv$ in $E[\Zmv\mid \Xv]$. Hence, the conditions for TSR are more attainable in practical
applications. 
In contrast, computing $\widetilde{\Yv}$ for RR requires access to data containing $\Xv$ and $\Zv$ jointly.

For future work, we plan to work on more flexible estimators that can, e.g., be instantiated through neural networks, and study more assumption violations. For instance the effect of missing variables, or absence of Gaussian errors. 

\backmatter

\bibliography{sn-bibliography}%

\section*{Declarations}
\subsection*{Funding}
Not applicable.
\subsection*{Competing interests}
The authors have no relevant financial or non-financial interests to disclose.
\subsection*{Ethics approval and consent to participate}
Not applicable.
\subsection*{Data availability}
Most evaluations are based on synthetic data, for which the code to generate the data is available at the provided code repository. The experiment shown in Sect. 5.5 leverages publicly available data from the Infant Health and Development Program (IHDP) available at: \url{https://www.tandfonline.com/doi/suppl/10.1198/jcgs.2010.08162}.
\subsection*{Materials availability}
Not applicable.
\subsection*{Code availability}
We provide the code to reproduce our results in the following GitHub repository: \url{https://github.com/causality-lab/TSR}
\subsection*{Author contribution}
All authors contributed to the study conception and design. Material preparation, data collection, analysis and the preparation of the first draft of the manuscript were performed by Marlies Hafer, who was supervised by Alexander Marx. All authors were involved in writing and editing the manuscript, and approved the final manuscript.

\newpage

\begin{appendices}
\appendix
\setcounter{figure}{7}                        %
\renewcommand{\thefigure}{\arabic{figure}}    %
\setcounter{table}{0}                    %
\renewcommand{\thetable}{\arabic{table}} %

\begin{center}
    \rule{\linewidth}{3pt}
    \vspace{-0.25cm}
    
    {\huge Appendix}
    \rule{\linewidth}{1pt}
\end{center}

\renewcommand{\contentsname}{Table of Contents}
\tableofcontents
\addtocontents{toc}{\protect\setcounter{tocdepth}{2}}

\section{Theoretical Details}
\label{sec:apptheory}
In this section we will give the definition of the do-calculus, which is fundamental for our theoretic results and the definition of s-recoverability, we referred to stating the definition of s-recoverability with access to external data. After this, we state and discuss the selection backdoor criterion by \citet{bareinboim:14:recovering-selection} and the generalized adjustment criterion type 3 by \citet{correa:18:GeneralizedAdj}.

\subsection{Definitions and Preliminaries}
\label{app:definitions}

For completeness, we restate the rules of do-calculus, which we need to derive our theoretical results below~\citep[Chapter~3]{pearl:09:causality}, and the definition of s-recoverability without access to external data~\citep{bareinboim:14:recovering-selection}.

\begin{definition}[Rules of do-Calculus]
For arbitrary disjoint sets of nodes $X$, $Y$, $Z$, and $W$ in a causal DAG $\mathcal{G}$, we denote the graph obtained by deleting all edges pointing towards a node in $X$ by $\mathcal{G}_{\overline{X}}$. Similarly, the graph obtained by deleting all edges pointing away from a node in $X$ by $\mathcal{G}_{\underline{X}}$. The graph obtained by deleting edges pointing towards nodes in $X$ and edges pointing away from nodes in $Z$ is denoted by $\mathcal{G}_{\overline{X},\underline{Z}}$. The symbol $\Indep_{\mathcal{G}}$ denotes d-separation in graph $\mathcal{G}$.

\begin{enumerate}
    \item (Insertion / deletion of observations):
    
    $P(Y=y\mid do(X:=x),Z=z,W=w)=P(Y=y\mid do(X:=x),W=w)$
    
    if $(Y\Indep_{\mathcal{G}_{\overline{X}}} Z\mid \{X,W\})$
    \item (Action / observation exchange):
    
     $P(Y=y\mid do(X:=x),do(Z:=z),W=w)=P(Y=y\mid do(X:=x),Z=z,W=w)$
     
     if $Y\Indep_{\mathcal{G}_{\overline{X},\underline{Z}}} Z\mid \{X,W\})$
    \item (Insertion / deletion of actions):

    $P(Y=y\mid do(X:=x),do(Z:=z),W=w)=P(Y=y\mid do(X:=x),W=w)$
    
    if $(Y\Indep_{\mathcal{G}_{\overline{X},\overline{Z(W)}}} Z\mid \{X,W\})$,
    
    where $Z(W)$ is the set of nodes in $Z$ not being ancestors of any node in $W$ in $\mathcal{G}_{\overline{X}}$
\end{enumerate}
\label{do-calculus}
\end{definition}

\begin{definition}[s-recoverability \citep{bareinboim:14:recovering-selection}]
\label{def:srecover}

Given a causal DAG model $(\mathcal{G}_s,P)$
augmented with a node $S$, 
the distribution $Q = P(Y\mid X)$ is said to be s-recoverable from selection biased data in $\mathcal{G}_s$ if the assumptions embedded in the causal model renders $Q$ expressible in terms of the distribution $P(V\mid S=1)$ under selection bias. Formally, for every two probability distributions $P_1$ and $P_2$ compatible with $\mathcal{G}_s$, $P_1(V=v\mid S=1)=P_2(V=v\mid S=1)>0$ implies $P_1(Y = y\mid X=x)=P_2(Y=y\mid X=x)$.
\end{definition}

\subsection{Discussion of Assumptions}
\label{sec:appassumptions}

In the following, we compare our assumptions to prior work, where we first review the selection backdoor criterion proposed by \citet{bareinboim:14:recovering-selection}, as well as the repeated regression estimator by \citet{boeken:23:privileged-info}, and then discuss the generalized adjustment criterion derived by \citet{correa:17:CausalEI}.

\paragraph{Selection backdoor criterion}
\citet{bareinboim:14:recovering-selection} proposed assumptions, given in Assumption~\ref{ass:Bareinboim}, under which the causal effect $P(Y|do(X))$ is, as stated in Theorem~\ref{theorem:bareinboim}, identifiable and s-recoverable.

\begin{assumption}[Selection backdoor criterion~\citep{bareinboim:14:recovering-selection}]
\label{ass:Bareinboim}
  The variables $Z$ can be decomposed as $Z=Z^{+}\cup Z^{-}$, where $Z^+$ are non-descendants of $X$ and $Z^-$ are descendants of $X$.
\begin{enumerate}[topsep=2pt,parsep=2pt,partopsep=1pt,leftmargin=*]
    \item $X$ and $Z$ block all paths between $S$ and $Y$, namely $Y\Indep_{\gG_s} S\mid \{X,Z\}$
    \item $Z^+$ blocks all backdoor paths from $X$ to $Y$, namely $(X\Indep_{{\gG_s}_{\underline{X}}} Y\mid Z^+)$
    \item $X$ and $Z^+$ block all paths between $Z^-$ and $Y$, namely $Z^-\Indep_{\gG_s} Y\mid \{X,Z^+\}$
    \item $Z\cup \{X,Y\}\subset M$, where variables $M$ are collected under selection bias (dataset $\mathcal{S}$) and $Z\subset T$, where $T$ is collected in the population-level (dataset $\mathcal{D}$).
\end{enumerate}
\end{assumption}

In comparison to the assumption above, in Assumption~\ref{ass:new}, we do not need (3.) which characterizes the relationship of $Z^+$ and $Z^-$, whereas we maintain the first and second subpoint. %
On the other hand, we need to observe $X$ also in the sample $\mathcal{D}$. However, this is not necessary for example in the linear case with $Z^-=\emptyset$.

\begin{theorem}[Selection backdoor adjustment~\citep{bareinboim:14:recovering-selection}] 
\label{theorem:bareinboim}
If $Z$ satisfies the  selection backdoor criterion (Assumption~\ref{ass:Bareinboim}) relative to $(X,Y)$ and $(M,T)$, then the causal effect \mbox{$P(Y\mid do(X))$} is identifiable, s-recoverable and can be expressed as
\[P(Y=y\mid do(X))=\int_{z}P(Y=y\mid X,Z=z,S=1)P(Z=z) dz\;.\]
\end{theorem}

Further, recall that for RR we need to assume PMAR as well as that $X$ and $Y$ are not confounded when aiming to estimate $E[Y\mid do(X)]$. We want to explain the relationship between the assumptions for RR, Assumption~\ref{ass:Bareinboim} proposed by \citet{bareinboim:14:recovering-selection}, and our Assumption~\ref{ass:new} for TSR based on the four DAGs in Figure~\ref{fig:vierDAGs}. First, note that Assumption~\ref{ass:new} is met for all of the four cases. The setting in Figure~\ref{fig:vierDAGs} (a) is met by all of the three assumptions. In contrast, in the setting in Figure~\ref{fig:vierDAGs} (b) Assumption~\ref{ass:Bareinboim} (3.) is violated as the edge between $Z^-$ and $Y$ can not be blocked by $X$ and $Z^+$. For the setting in Figure~\ref{fig:vierDAGs} (c) it is exactly the opposite. Here, RR does not recover the causal effect but only $E[Y|X]$, whereas Assumption~\ref{ass:Bareinboim} is fulfilled. The setting in Figure~\ref{fig:vierDAGs} (d) violates Assumption~\ref{ass:Bareinboim} (3.), and induces confounding, which is why RR does not recover the causal effect.

In \Cref{sec:our-identification-result}, we mentioned that Assumption~\ref{ass:new} requires $X$ and $Z$ to be observed unbiased, whereas for Assumption~\ref{ass:Bareinboim} only $Z$ must be observed unbiased. That is, if $X$ is not observable unbiased, and $Z^-\neq\emptyset$, there might be cases which meet Assumption~\ref{ass:Bareinboim}, but not our Assumption~\ref{ass:new}. It is to mention, that settings could occur in which our assumption is met, but the selection backdoor criterion is not met due to a lack of access to unbiased data. Think of cases for which it is difficult to observe $Z^+$ unbiased, whereas observing $X$ unbiased is unproblematic. In those cases, it may be that our assumption is fulfilled but the selection backdoor criterion is not.
As discussed above, for example, Figure~\ref{fig:vierDAGs} (b) does not satisfy the third point of Assumption~\ref{ass:Bareinboim}, whereas Assumption~\ref{ass:new} is met. Swap the direct path $Z^-\rightarrow Y$ to one that goes via confounder $Z^+$, $Z^-\leftarrow Z^+\rightarrow Y$. Assuming, that $Z^+$ can not be observed unbiased, Assumption~\ref{ass:Bareinboim} can not be satisfied, whereas, taking advantage of Corollary~\ref{corollary:ZminusgivendoX}, Assumption~\ref{ass:new} can be satisfied if $(X,Z^-)$ are observable unbiased.

\paragraph{Generalized Adjustment Criterion 3 (GACT3)}
\citet{correa:18:GeneralizedAdj} proposed assumptions, given in Definition~\ref{def:gact3}, under which the causal effect $P(Y|do(X))$ is, as stated in Theorem~\ref{corollary:correa}, identifiable and s-recoverable, which requires some preliminary definitions by \citet{correa:18:GeneralizedAdj} stated below.

\begin{definition}[Proper Causal Path]
Let $X$ and $Y$ be sets of nodes. A causal path from a node in $X$ to a node in $Y$ is called proper if it does not intersect $X$ except at the starting point.
\end{definition}

\begin{definition}[Proper Backdoor Graph]
Given a causal DAG model $(\mathcal{G},P)$ and disjoint subsets $X$ and $Y$ of variables. The proper backdoor graph, denoted as $\mathcal{G}_{XY}^{pbd}$, is obtained from $\mathcal{G}$ by removing the first edge of every proper causal path from $X$ to $Y$.
\end{definition}

\begin{definition}[Adjustment Pair] Given a causal DAG model $(\mathcal{G}_s,P)$
augmented with a \mbox{node $S$}%
, disjoint sets of variables $X$, $Y$, $Z$, and a set $Z^T \subset Z$, $(Z, Z^T)$ is said to be an adjustment pair for recovering the causal effect of $X$ on $Y$
if for every model compatible with $\mathcal{G}_s$, \mbox{$P (Y{=}y\mid do(X{=}x))$} can be expressed as
\[\sum_z
P (Y=y\mid X=x,Z=z,S=1)P (Z=z \setminus Z^T=z^T\mid Z^T=z^T,S=1)P (Z^T=z^T)\;.\]
\label{def:adjpairs}
\end{definition}

\begin{assumption}[Generalized Adjustment Criterion Type 3 (GACT3)]
\label{def:gact3}
Given a causal DAG model $(\gG_s,P)$
augmented with a node $S$%
, disjoint sets of variables $X, Y, Z$ and set $Z^T\subset Z$; $(Z,Z^T)$ is an admissible pair relative to $X, Y$ in $\gG_s$ if:
\begin{enumerate}[topsep=2pt,parsep=2pt,partopsep=1pt,leftmargin=*]
    \item No element in $Z$ is a descendant in ${\gG_s}_{\overline{X}}$ of any $W\notin X$ lying on a proper causal path from \mbox{$X$ to $Y$}.
    \item All non-causal paths in $\gG_s$ from $X$ to $Y$ are blocked by $Z$ and $S$.
    \item $Z^T$ d-separates $Y$ from $S$ in the proper backdoor graph, i.e. $(Y\Indep S\mid Z^T)_{{\gG_s}_{XY}^{pbd}}$
\end{enumerate}
\end{assumption}

\begin{theorem} [Admissible Pairs are Adjustment Pairs] $(Z, Z^T)$ is an adjustment pair for $X$, $Y$ in $\mathcal{G}$ if and only
if it is admissible by Definition~\ref{def:gact3}.
\end{theorem}

\begin{corollary}[Causal Effects Recovery by Adjustment] Given a causal DAG model $(\gG_s,P)$ augmented with a node $S$ representing the selection mechanism. Let $V$ be the set of variables measured under selection bias, and $T\subset V$ the set of variables measured externally in the overall population. Consider disjoint sets of variables $X, Y \subset V$,
then the causal effect \mbox{$P (Y=y\mid do(X:=x))$} is recoverable from
$\{P (V=v\mid S=1), P (T=t)\}$ by the adjustment expression in Definition~\ref{def:adjpairs} while
$Z^T \subset T$, in every model inducing $\gG_s$ if and only if $(Z, Z^T)$
is an admissible pair relative to $X$, $Y$ in $\gG_s$ according to
Definition~\ref{def:gact3}.
\label{corollary:correa}
\end{corollary}

The assumptions in Definition~\ref{def:gact3} are met in \Cref{fig:vierDAGs} (a)--(c), but are not admissible for (d) in cases where $U$ is not contained in $Z$.
Such cases could occur if either we did not include $U$ in $Z$, or it is unobserved. In either case, not all non-causal paths from $X$ to $Y$ can be blocked by $Z$ and $S$. Conditioning on $Z^-$ opens the path $X\rightarrow Z^- \leftarrow U \rightarrow Y$. However, $Z^-$ must be included into $Z$ to meet the PMAR assumption. 

\paragraph{Targeted Sequential Regression}

Focusing on binary treatment variables and aiming to estimate average treatment effects in the context of confounding and attrition or selection bias, \citet{de-aguas:25:recovery} derived the result of Theorem~\ref{theorem:neutheoretisch} under similar conditions by directly leveraging the back-door criterion. In comparison to their approach, we root our analysis in the assumption that data is missing privilegedly at random that was proposed by \citet{boeken:23:privileged-info}, which \citet{de-aguas:25:recovery} independently stated in their work. Further, they do not consider the external data setting. Both identification results hold for categorical and continuous treatments. They then estimate the average treatment effect for binary treatment variables via sequential regression based on propensity scores, while we pursue a direct regression approach for continuous treatments for which we derive the variance and compare it to the repeated regression estimator that has been proposed by~\citet{boeken:23:privileged-info}. 

We illustrate how the assumptions can lead to distinct estimators at the example illustrated in Figure~\ref{fig:vierDAGs}~(a). In essence, the formulation of \citet{de-aguas:25:recovery} will allow for two possible estimators while our proposed approach will only allow for the choice that is more sample efficient. Concretely, they define a set $W$ blocking the backdoor paths taking on the same role as our $Z^+$. Additionally, the sets $W,X$ and $V$ together block the target from the selection variable (PMAR). Taking a look at Figure~\ref{fig:vierDAGs}~(a), the setting with selection bias but without confounding, where $Z^+=\{Z\}$ and $Z^-=\emptyset$, using the minimal s-admissible pair, one can either choose $W=\{Z\}$ and $V=\emptyset$ or $W=\emptyset$ and $V=\{Z\}$. While choosing $W=\{Z\}$ would coincide with $Z^+=\{Z\}$, the latter case, i.e., setting $V=\{Z\}$ would not be valid for our setting since we restrict $Z^{-}$ to descendants of $X$. If we would hypothetically allow for setting $Z^{-}=\{Z\}$, TSR and RR would coincide, leading to an estimator that is less sample efficient than when choosing $Z^{+}=\{Z\}$, as illustrated in Theorem~\ref{theorem:variance}.

\subsection{Identification Proofs}
\label{sec:appproofs}

In this section, we provide the full proofs for our identification results.

\neutheoretisch*

\begin{proof}
We can express the expected causal effect $E[Y\mid do(X)]$ as
\begin{align*}
    E[Y\mid do(X)]&=\int_{z^+}\underbrace{E[Y\mid do(X),Z^+=z^+]}_{\underset{Y \Indep_{{\gG_s}_{\underline{X}}} X\mid Z^+}{=}E[Y\mid X,Z^+=z^+]}\underbrace{P(Z^+=z^+\mid do(X))}_{\underset{\text{non-desc}}=P(Z^+=z^+)}dz^+\\
              &=\int_{z^+}\underbrace{E[Y\mid X,Z^+=z^+]}_{=E[E[Y\mid X,Z^+=z^+,Z^-]\mid X,Z^+=z^+]}P(Z^+=z^+)dz^+\\
              &=\int_{z^+}E[\underbrace{E[Y\mid X,Z^+=z^+,Z^-]}_{=E[Y\mid X,Z^+=z^+,Z^-,S=1]}\mid X,Z^+=z^+]P(Z^+=z^+)dz^+\\
              &=\int_{z^+}E[E[Y\mid X,Z^+=z^+,Z^-,S=1]\mid X,Z^+=z^+]P(Z^+=z^+)dz^+\\           
\end{align*}

Line one and three follow from the law of total expectation. In row two, we can apply the second rule of do-calculus (Definition~\ref{do-calculus}, Appendix~\ref{app:definitions}) since $Y\Indep_{{\gG_s}_{\underline{X}}}  X\mid Z^+$ (Assumption~\ref{ass:new} (2.)), and $Z^+$ is non-descendant of $X$. Assumption~\ref{ass:new} (1.) ensures the final equality.

Following the proof of s-recoverability by \citet[Theorem~\ref{theorem:bareinboim}]{bareinboim:14:recovering-selection}, since the causal effect can be represented in probability terms of the selected sample and of the external data, along with Assumption~\ref{ass:new}, the achieved expression ensures s-recoverability.
\end{proof}

\theoremlinearity*

\begin{proof}
\begin{align*}
    E[Y|do(X:=x)]=&\int_{z^+}E[E[Y\mid X=x,Z^+=z^+,Z^-,S=1]\mid X=x,Z^+=z^+]P(Z^+=z^+)dz^+\\  
              =&\int_{z^+}E[\beta_0+\beta_1 X+\beta_2 Z^++\beta_3 Z^-\mid X=x,Z^+=z^+]P(Z^+=z^+)dz^+\\ 
              =&\int_{z^+}(\beta_0+\beta_1 x+\beta_2 z^++\beta_3 E[Z^-\mid X=x,Z^+=z^+])P(Z^+=z^+)dz^+\\ 
              =&\beta_0+\beta_1 x+\beta_2 \int_{z^+}z^+ P(Z^+=z^+)dz^+\\
              &+\beta_3 \int_{z^+} E[Z^-\mid X=x,Z^+=z^+]P(Z^+=z^+)dz^+\\
              =&\beta_0+\beta_1 x+\beta_2 E[Z^+]+\beta_3 \int_{z^+} E[Z^-\mid X=x,Z^+=z^+]P(Z^+=z^+)dz^+\\
\end{align*}\\
\end{proof}

\ZminusgivendoX*

\begin{proof}
\begin{align*}
   E[Z^-\mid do(X)]&=\int_{z^+}E[Z^-\mid do(X),Z^+=z^+]P(Z^+=z^+\mid do(X))dz^+\\
   &=\int_{z^+}E[Z^-\mid X,Z^+=z^+]P(Z^+=z^+)dz^+
\end{align*}
In the derivation above, we exploit that $Z^-\Indep X\mid Z^+$ in $\mathcal{G}_{\underline{X}}$ along with the second rule of do-calculus, the Law of total expectation and $Z^+$ being non-descendant of $X$. Further, \mbox{$E[Z^-\mid X]=E[Z^-\mid do(X)]$}, if $X$ and $Z^-$ are not confounded.
\end{proof}

\subsection{Comparison of Bias and Variance of RR and TSR}
\label{sec:appcoincide}
RR and TSR coincide when $\hat{\mu}_{TSR}(x)=\hat{\beta}_0+\hat{\beta}_1x+\hat{\beta}_2\widehat{E}[Z^-\mid X=x]$, which is the case in absence of confounding and $Z^+=\emptyset$ (implying $Z=Z^-$) as in \Cref{fig:vierDAGs} (b).

The first step of regression for both estimators, yields
\(\hat{\beta}=(\hat{\beta}_x,\hat{\beta}_z)\),
where $\hat{\beta}_x:=(\hat{\beta}_0,\hat{\beta}_1)$ and $\hat{\beta}_z:=\hat{\beta}_2$.
Then TSR is given by
\[\hat{\mu}_{TSR}(x)=\hat{\beta}_x\overrightarrow{x}+\hat{\beta}_z((B_X^TB_X)^{-1}B_X^TB_Z)^T\overrightarrow{x}\;,\]
where $\overrightarrow{x}:=(1,x^T)^T$, $B_X:=(1,X_i^T)_{i\in\mathcal{D}}\in\mathbb{R}^{|\mathcal{D}|\times (p+1)}$ and $B_Z:=(Z_i^T)_{i\in\mathcal{D}}\in\mathbb{R}^{|\mathcal{D}|\times d}$.
RR is derived as follows:
\begin{align*}
\hat{\mu}_{RR}(x)
&=(\overrightarrow{x})^T(B_X^TB_X)^{-1}B_X^T(B_X(\hat{\beta}_x)^T+B_Z(\hat{\beta}_z)^T)\\
&=(\overrightarrow{x})^T(B_X^TB_X)^{-1}B_X^TB_X(\hat{\beta}_x)^T+(\overrightarrow{x})^T(B_X^TB_X)^{-1}B_X^TB_Z(\hat{\beta}_z)^T\\
&=(\overrightarrow{x})^T(\hat{\beta}_x)^T+(\overrightarrow{x})^T(B_X^TB_X)^{-1}B_X^TB_Z(\hat{\beta}_z)^T\\
&=\hat{\beta}_x\overrightarrow{x}+\hat{\beta}_z((B_X^TB_X)^{-1}B_X^TB_Z)^T\overrightarrow{x}\;.
\end{align*}
This verifies the equality $\hat{\mu}_{RR}(x)=\hat{\mu}_{TSR}(x)$ for the linear case. With the same calculation but other definitions for $B_X,B_Z,\hat{\beta}_x,\hat{\beta}_z$ and $\overrightarrow{x}$, the setting can be generalized---for example to OLS with impacts up to a specific degree.

In settings, for which we cannot exclude confounding and $Z^-=\emptyset$ (implying $Z=Z^+$), as in \Cref{fig:vierDAGs} (c), TSR, given by
$\hat{\mu}_{TSR}(x)=\hat{\beta}_0+\hat{\beta}_1x+\hat{\beta}_2\widehat{E}[Z^+]$ in this setting, does not exactly equal the RR estimation. We consider the case, where $X$ and $Z$ are of dimension one.

As we will show below, using OLS,
\begin{align*}
    \hat{\mu}_{TSR}(x) &=\hat{\beta}_0+\hat{\beta}_1x+\hat{\beta}_2\overline{Z^+} \text{ , and}\\
    \hat{\mu}_{RR}(x)
    &=\hat{\mu}_{TSR}(x)+\hat{\beta}_2\frac{\widehat{Cov}[Z^+,X]}{\widehat{Var}[X]}(x-\bar{X})\;.
\end{align*}
The derivation is at follows. First, define $\widetilde{Y}$ for the second regression step for RR as
\begin{align*}
\widetilde{Y}:=\hat{\beta}_0+\hat{\beta}_1X+\hat{\beta}_2Z^+\;.
\end{align*}
Then, running again a linear regression, we search for estimators $\hat{\alpha}_0$ as intercept and $\hat{\alpha}_1$ as coefficient of $X$ for the second step of RR. They are given as follows:
\begin{align*}
\hat{\alpha}_1 &=\frac{\widehat{Cov}[\widetilde{Y},X]}{\widehat{Var}[X]}=\hat{\beta}_1+\hat{\beta}_2\frac{\widehat{Cov}[Z^+,X]}{\widehat{Var}[X]} \; \\
    \hat{\alpha}_0
    &=\bar{\widetilde{Y}}-\hat{\alpha}_1\bar{X} 
    =\hat{\beta}_0+\hat{\beta}_1\bar{X}+\hat{\beta}_2\overline{Z^+}-\hat{\alpha}_1\bar{X}\\
    &=\hat{\beta}_0+\hat{\beta}_1\bar{X}+\hat{\beta}_2\overline{Z^+}-\biggl(\hat{\beta}_1+\hat{\beta}_2\frac{\widehat{Cov}[Z^+,X]}{\widehat{Var}[X]}\biggr)\bar{X}\\
    &=\hat{\beta}_0+\hat{\beta}_2\overline{Z^+}-\hat{\beta}_2\frac{\widehat{Cov}[Z^+,X]}{\widehat{Var}[X]}\bar{X}\;.
\end{align*}
This results in 
\begin{align*}
    \hat{\mu}_{RR}(x)
    &=\hat{\alpha}_0+\hat{\alpha}_1x\\
    &=\hat{\beta}_0+\hat{\beta}_2\overline{Z^+}-\hat{\beta}_2\frac{\widehat{Cov}[Z^+,X]}{\widehat{Var}[X]}\bar{X}
      +\biggl(\hat{\beta}_1+\hat{\beta}_2\frac{\widehat{Cov}[Z^+,X]}{\widehat{Var}[X]}\biggr)x\\
    &=\underbrace{\hat{\beta}_0+\hat{\beta}_1x+\hat{\beta}_2\overline{Z^+}}_{=\hat{\mu}_{TSR}(x)}+\hat{\beta}_2\frac{\widehat{Cov}[Z^+,X]}{\widehat{Var}[X]}(x-\bar{X})\;.
\end{align*}
Even if $Cov[Z^+,X]=0$, $\widehat{Cov}[Z^+,X]\neq 0$ for finite samples. With $\mathcal{S}\cap\mathcal{D}=\emptyset$,
\begin{align*}
    E\biggl[\hat{\beta}_2\frac{\widehat{Cov}[Z^+,X]}{\widehat{Var}[X]}(x-\bar{X})\biggr]
    &=E[\hat{\beta}_2]E\biggl[\frac{\widehat{Cov}[Z^+,X]}{\widehat{Var}[X]}(x-\bar{X})\biggr]\;,
\end{align*}
which diminishes for $n\rightarrow \infty$, provided that the estimate of the variance in the denominator does not asymptotically approach zero.

\subsubsection{Bias and Variance}
\label{sec:appproofbiasvar}

For the calculation of the bias and variance of RR, we restrict ourselves to the case, where $X$ and $Z$ are of dimension one. We make use of the explicit form \mbox{$\hat{\mu}_{RR}(x)=\hat{\beta}_0+\hat{\beta}_1x+\hat{\beta}_2(\hat{\alpha}_0+\hat{\alpha}_1 x)$} and justify this expression in the following, deriving the repeated regression estimator by hand.
For the second regression, we estimate a simple linear regression and therefore  set \mbox{$\widehat{E}[\widetilde{Y}\mid X=x]=\hat{\delta}_0+\hat{\delta}_1 x$}. The derivation of $\hat{\delta}_0$ and $\hat{\delta}_1$ will be given in the following:

First, we use that for the simple linear regression using OLS, $\hat{\delta}_1=\frac{\widehat{Cov}[\widetilde{Y},X]}{\widehat{Var}[X]}$, such that
\begin{align*}
        \widehat{Var}[X]\hat{\delta}_1=\widehat{Cov}[\widetilde{Y},X]
        &=\widehat{Cov}[\hat{\beta}_0+\hat{\beta}_1 X+\hat{\beta}_2Z,X]\\
    &=\underbrace{\widehat{Cov}[\hat{\beta}_0,X]}_{=0}+\underbrace{\widehat{Cov}[\hat{\beta}_1X,X]}_{\hat{\beta}_1\widehat{Var}[X]}+\underbrace{\widehat{Cov}[\hat{\beta}_2Z,X]}_{=\hat{\beta}_2\widehat{Cov}[Z,X]} \; .
\end{align*}
Consequently, we arrive at
\[\hat{\delta}_1=\hat{\beta}_1+\hat{\beta}_2\frac{\widehat{Cov}[Z,X]}{\widehat{Var}[X]}\;.\]
From this, we can calculate $\hat{\delta}_0$ as follows:
   \[ \hat{\delta}_0=\bar{\widetilde{Y}}-\hat{\delta}_1\bar{X}=\hat{\beta}_0+\hat{\beta}_1\bar{X}+\hat{\beta}_2\bar{Z}-\hat{\beta}_1\bar{X}-\hat{\beta}_2\frac{\widehat{Cov}[Z,X]}{\widehat{Var}[X]}\bar{X}=\hat{\beta}_0+\hat{\beta}_2\bar{Z}-\hat{\beta}_2\frac{\widehat{Cov}[Z,X]}{\widehat{Var}[X]}\bar{X}\;,\]
where $\bar{\widetilde{Y}}=\hat{\beta}_0+\hat{\beta}_1\bar{X}+\hat{\beta}_2\bar{Z}$. Plugging in $\hat{\delta}_0$ and $\hat{\delta}_1$, the repeated regression estimator can be expressed as
   \begin{align*}
        \hat{\mu}_{RR}(x)&=\hat{\delta}_0+\hat{\delta}_1x\\
    &=\hat{\beta}_0+\hat{\beta}_2\bar{Z}-\hat{\beta}_2\frac{\widehat{Cov}[Z,X]}{\widehat{Var}[X]}\bar{X}+\hat{\beta}_1x+\hat{\beta}_2\frac{\widehat{Cov}[Z,X]}{\widehat{Var}[X]}x\\
    &=\hat{\beta}_0+\hat{\beta}_1x+\hat{\beta}_2\biggl(\underbrace{\bar{Z}-\frac{\widehat{Cov}[X,Z]}{\widehat{Var}[X]}\bar{X}}_{=\hat{\alpha}_0}+\underbrace{\frac{\widehat{Cov}[X,Z]}{\widehat{Var}[X]}}_{=\hat{\alpha}_1}x\biggr)\\
    &=\hat{\beta}_0+\hat{\beta}_1X+\hat{\beta}_2(\hat{\alpha}_0+\hat{\alpha}_1 x)\;,
\end{align*}
where $\hat{\alpha}_0$ and $\hat{\alpha}_1$ denote the OLS coefficient estimates of $\alpha_0$ and $\alpha_1$ corresponding to the simple linear regession model $E[Z|X=x]=\alpha_0+\alpha_1x$.

\textbf{Unbiasedness}~~~Now, we can calculate the empirical mean and discuss its bias for RR and TSR:
\begin{align*}
    E[\hat{\mu}_{RR}(x)]
    &= E[\hat{\beta_0}+\hat{\beta}_1 x+\hat{\beta}_2 (\hat{\alpha}_0+\hat{\alpha}_1 x)]\\
    \underset{\text{linearity}}{}&= E[\hat{\beta}_0]+E[\hat{\beta}_1]x+E[\hat{\beta}_2(\hat{\alpha}_0+\hat{\alpha}_1 x)]\\
    &=E[\hat{\beta}_0]+E[\hat{\beta}_1] x+E[\hat{\beta}_2]E[\hat{\alpha}_0+\hat{\alpha}_1 x]+Cov[\hat{\beta}_2,\hat{\alpha}_0+\hat{\alpha}_1 x]\\
    \underset{\text{step 1 model correctly specified}}{}&=\beta_0+\beta_1 x+{\beta}_2E[\hat{\alpha}_0+\hat{\alpha}_1 x]+Cov[\hat{\beta}_2,\hat{\alpha}_0+\hat{\alpha}_1 x]\\
    \underset{\text{step 2 model correctly specified}}{}&=\beta_0+\beta_1 x+\beta_2(\alpha_0+\alpha_1 x)
        +Cov[\hat{\beta}_2,\hat{\alpha}_0+\hat{\alpha}_1 x]\\
    \underset{\alpha_0=E[\overline{Z^+}]=E[Z^+],\;\alpha_1=0}{}&=\beta_0+\beta_1 x+\beta_2E[Z^+]
        +\underbrace{Cov[\hat{\beta}_2,\hat{\alpha}_0+\hat{\alpha}_1 x]}_{=\begin{cases}
        \neq 0 & \mathcal{S}\subset \mathcal{D}\\
        =0     & \mathcal{S}\cap \mathcal{D}=\emptyset
        \end{cases}}
\end{align*}

\begin{align*}
E[\hat{\mu}_{TSR}(x)]
    &=E[\hat{\beta_0}+\hat{\beta}_1 x+\hat{\beta}_2 \overline{Z^+}]\\
    \underset{linearity}{}&= E[\hat{\beta}_0]+E[\hat{\beta}_1]x+E[\hat{\beta}_2\overline{Z^+}]\\
    &=E[\hat{\beta}_0]+E[\hat{\beta}_1] x+E[\hat{\beta}_2]E[\overline{Z^+}]+Cov[\hat{\beta}_2,\overline{Z^+}]\\
    \underset{\text{step 1 model correctly specified}}{}&=\beta_0+\beta_1 x+\beta_2E[\overline{Z^+}]+Cov[\hat{\beta}_2,\overline{Z^+}]\\
        &=\beta_0+\beta_1 x+\beta_2E[Z^+]+\underbrace{Cov[\hat{\beta}_2,\overline{Z^+}]}_{=\begin{cases}
        \neq 0 & \mathcal{S}\subset \mathcal{D}\\
        =0     & \mathcal{S}\cap \mathcal{D}=\emptyset
        \end{cases}}
\end{align*}
We discuss the bias for the general case for TSR, where $X$ and $Z$ are not restricted to dimension one. Suppose we are in the setting with \[\hat{\mu}_{TSR}(x)=\hat{\beta}_0+\hat{\beta}_1x+\hat{\beta}_2\overline{Z^+}+\hat{\beta}_3(\hat{\gamma}_0+\hat{\gamma}_1x+\hat{\gamma}_2\overline{Z^+})\;.\] Due to the unbiasedness of OLS and $\mathcal{S}\cap\mathcal{D}=\emptyset$, the bias of TSR reduces to the bias of just $\hat{\beta}_3\hat{\gamma}_2\overline{Z^+}$ w.r.t. $\beta_3\gamma_2E[Z^+]$, which is given by
\begin{align*}
    \text{Bias}[\hat{\mu}_{TSR}]=&E[\hat{\beta}_3\hat{\gamma}_2\overline{Z^+}]-\beta_3\gamma_2E[Z^+]\\
   =&\underbrace{E[\hat{\beta}_3]}_{=\beta_3}E[\hat{\gamma}_2\overline{Z^+}]-\beta_3\gamma_2E[Z^+]\\
   =&\beta_3(E[\hat{\gamma}_2\overline{Z^+}]-\gamma_2E[Z^+])\\
   =&\beta_3(E[\hat{\gamma}_2\overline{Z^+}]-E[\hat{\gamma}_2]E[\overline{Z^+}])\;,
\end{align*} where for the second line we used $\mathcal{S}\cap\mathcal{D}=\emptyset$ and unbiasedness of OLS, and for the fourth we used unbiasedness of OLS and linearity of the expectation together with the fact that the repetitions of $Z^+$ are identically distributed. We can ensure unbiasedness by splitting the unbiased dataset into two disjoint subsets---one to estimate $\hat{\gamma}_0$, $\hat{\gamma}_1$ and $\hat{\gamma}_2$ via the second regression, and one to compute the empirical mean of $Z^+$---although this comes at the cost of reduced efficiency.

\textbf{Variance}~~~Next, we prove the result comparing the variances of RR and TSR.
\theoremvariance*
\begin{proof}
First, we derive the variance for $\hat{\mu}_{RR}(x)$.
\begin{align*}
    Var[\hat{\mu}_{RR}(x)]
    =& Var[\hat{\beta_0}+\hat{\beta}_1 x+\hat{\beta}_2 (\underbrace{\underbrace{\hat{\alpha}_0}_{=\overline{Z^+}-\hat{\alpha}_1\bar{X}}+\hat{\alpha}_1 x}_{=\widehat{E}[Z^+\mid X]})]\\
    =& Var[\hat{\beta}_0+\hat{\beta}_1 x+\hat{\beta}_2\overline{Z^+}] + Var[\hat{\beta}_2\hat{\alpha}_1(x-\bar{X})]\\
    &\;+2 Cov[\hat{\beta}_0+\hat{\beta}_1 x+\hat{\beta}_2\overline{Z^+},\hat{\beta}_2\hat{\alpha}_1(x-\bar{X})]\\
\end{align*}
Based on the above result, we can write the difference in variance of both estimators as \mbox{$\Delta = Var[\hat{\mu}_{RR}(x)]-Var[\hat{\mu}_{TSR}(x)]$}, where we can express $\Delta$ as
\begin{align*}
\Delta
    =&Var[\hat{\mu}_{RR}(x)]-Var[\hat{\beta}_0+\hat{\beta}_1 x+\hat{\beta}_2\overline{Z^+}]\\
    =& Var[\hat{\beta}_2\hat{\alpha}_1(x-\bar{X})]+ 2 Cov[\hat{\beta}_0+\hat{\beta}_1 x+\hat{\beta}_2\overline{Z^+},\hat{\beta}_2\hat{\alpha}_1(x-\bar{X})]\\
    =& Var[\hat{\beta}_2\hat{\alpha}_1(x-\bar{X})] + 2 ( Cov[\hat{\beta}_0,\hat{\beta}_2\hat{\alpha}_1x]
    -Cov[\hat{\beta}_0,\hat{\beta}_2\hat{\alpha}_1\bar{X}]
    +Cov[\hat{\beta}_1 x,\hat{\beta}_2\hat{\alpha}_1x]\\
    &-Cov[\hat{\beta}_1 x,\hat{\beta}_2\hat{\alpha}_1\bar{X}]
    +Cov[\hat{\beta}_2\overline{Z^+},\hat{\beta}_2\hat{\alpha}_1x]
    -Cov[\hat{\beta}_2\overline{Z^+},\hat{\beta}_2\hat{\alpha}_1\bar{X}])\\
    =&Var[\hat{\beta}_2\hat{\alpha}_1(x-\bar{X})]+2 \cdot ( \underbrace{E[\hat{\beta}_0\hat{\beta}_2\hat{\alpha}_1x]}_{\underset{\mathcal{S}\cap\mathcal{D}=\emptyset}{=}xE[\hat{\beta}_0\hat{\beta}_2]\underbrace{E[\hat{\alpha}_1]}_{\underset{X\Indep Z^+}{=}0}}-\underbrace{E[\hat{\beta}_0]E[\hat{\beta}_2\hat{\alpha}_1x]}_{\underset{\mathcal{S}\cap\mathcal{D}=\emptyset}{=}E[\hat{\beta}_0]xE[\hat{\beta}_2]\underbrace{E[\hat{\alpha}_1]}_{=0}}\\
    &\hspace{2.85cm}-\underbrace{E[\hat{\beta}_0\hat{\beta}_2\hat{\alpha}_1\bar{X}]}_{\underset{\mathcal{S}\cap\mathcal{D}=\emptyset}{=}E[\hat{\beta}_0\hat{\beta}_2]\underbrace{E[\hat{\alpha}_1\bar{X}]}_{=\underbrace{E[\hat{\alpha}_1]}_{=0}E[\bar{X}]}}+E[\hat{\beta}_0]\underbrace{E[\hat{\beta}_2\hat{\alpha}_1\bar{X}]}_{\underset{\mathcal{S}\cap\mathcal{D}=\emptyset}{=}E[\hat{\beta}_2]\underbrace{E[\hat{\alpha}_1]}_{=0}E[\bar{X}]}\\
    &\hspace{2.85cm}+\underbrace{E[\hat{\beta}_1x\hat{\beta}_2\hat{\alpha}_1x]}_{\underset{\mathcal{S}\cap\mathcal{D}=\emptyset}{=}x^2E[\hat{\beta}_1\hat{\beta}_2]\underbrace{E[\hat{\alpha}_1]}_{=0}}-\underbrace{E[\hat{\beta}_1x]E[\hat{\beta}_2\hat{\alpha}_1x]}_{\underset{\mathcal{S}\cap\mathcal{D}=\emptyset}{=}E[\hat{\beta}_1]x^2E[\hat{\beta}_2]\underbrace{E[\hat{\alpha}_1]}_{=0}}\\
    &\hspace{2.85cm}-\underbrace{E[\hat{\beta}_1 x\hat{\beta}_2\hat{\alpha}_1\bar{X}]}_{\underset{\mathcal{S}\cap\mathcal{D}=\emptyset}{=}xE[\hat{\beta}_1\hat{\beta}_2]\underbrace{E[\hat{\alpha}_1\bar{X}]}_{=\underbrace{E[\hat{\alpha}_1]}_{=0}E[\bar{X}]}}+E[\hat{\beta}_1 x]\underbrace{E[\hat{\beta}_2\hat{\alpha}_1\bar{X}]}_{\underset{\mathcal{S}\cap\mathcal{D}=\emptyset}{=}E[\hat{\beta}_2]\underbrace{E[\hat{\alpha}_1\bar{X}]}_{=\underbrace{E[\hat{\alpha}_1]}_{=0}E[\bar{X}]}}\\
    &\hspace{2.85cm}+\underbrace{E[\hat{\beta}_2\overline{Z^+}\hat{\beta}_2\hat{\alpha}_1x]}_{\underset{\mathcal{S}\cap\mathcal{D}=\emptyset}{=}xE[\hat{\beta}_2\hat{\beta}_2]\underbrace{E[\overline{Z^+}\hat{\alpha}_1]}_{=E[\overline{Z^+}]\underbrace{E[\hat{\alpha}_1]}_{=0}}}-E[\hat{\beta}_2\overline{Z^+}]\underbrace{E[\hat{\beta}_2\hat{\alpha}_1x]}_{\underset{\mathcal{S}\cap\mathcal{D}=\emptyset}{=}xE[\hat{\beta}_2]\underbrace{E[\hat{\alpha}_1]}_{=0}}\\
    &\hspace{2.85cm}-\underbrace{E[\hat{\beta}_2\overline{Z^+}\hat{\beta}_2\hat{\alpha}_1\bar{X}]}_{\underset{\mathcal{S}\cap\mathcal{D}=\emptyset}{=}E[\hat{\beta}_2\hat{\beta}_2]\underbrace{E[\overline{Z^+}\hat{\alpha}_1\bar{X}]}_{=0}}+E[\hat{\beta}_2\overline{Z^+}]\underbrace{E[\hat{\beta}_2\hat{\alpha}_1\bar{X}]}_{\underset{\mathcal{S}\cap\mathcal{D}=\emptyset}{=}E[\hat{\beta}_2]\underbrace{E[\hat{\alpha}_1\bar{X}]}_{=\underbrace{E[\hat{\alpha}_1]}_{=0}E[\bar{X}]}})\\
    &=Var[\hat{\beta}_2\hat{\alpha}_1(x-\bar{X})]\geq 0\;,
\end{align*}
where we used $Cov[\bar{X},\hat{\alpha}_1]=0$ and $Cov[\overline{Z^+},\hat{\alpha}_1]=0$ exploiting $E[\bar{X}\hat{\alpha}_1]=E[\bar{X}]E[\hat{\alpha}_1]$ and $E[\overline{Z^+}\hat{\alpha}_1]=E[\overline{Z^+}]E[\hat{\alpha}_1]$, as well as $E[\overline{Z^+}\hat{\alpha}_1\bar{X}]=0$ which will be proven in the following.

We start with deriving $Cov[\bar{X},\hat{\alpha}_1]=0$. Denoting the unbiased empirical variance of $X$ by \mbox{$\hat{\sigma}_X^2=\frac{1}{\mid\mathcal{D}\mid -1}\sum_{j\in\mathcal{D}}(X_j-\bar{X})^2$}, we have
\begin{align*}
    Cov[\bar{X},\hat{\alpha}_1]
    =&Cov\biggl[\bar{X},\frac{1}{(\mid\mathcal{D}\mid -1)\hat{\sigma}_X^2}\sum_{j\in\mathcal{D}}(X_j-\bar{X})(Z^+_j-\overline{Z^+})\biggr]\\
    =&E\biggl[\bar{X}\frac{1}{(\mid\mathcal{D}\mid -1)\hat{\sigma}_X^2}\sum_{j\in\mathcal{D}}(X_j-\bar{X})(Z^+_j-\overline{Z^+})\biggr]\\
    &\;-E[\bar{X}]E\biggl[\frac{1}{(\mid\mathcal{D}\mid -1)\hat{\sigma}_X^2}\sum_{j\in\mathcal{D}}(X_j-\bar{X})(Z^+_j-\overline{Z^+})\biggr]\\
    =&E\biggl[\bar{X}\frac{1}{(\mid\mathcal{D}\mid -1)\hat{\sigma}_X^2}\sum_{j\in\mathcal{D}}(X_j-\bar{X})\underbrace{E[Z^+_j-\overline{Z^+}\mid X]}_{=E[Z^+_j-\overline{Z^+}]=0}\biggr]\\
    &\;-E[\bar{X}]E\biggl[\frac{1}{(\mid\mathcal{D}\mid -1)\hat{\sigma}_X^2}\sum_{j\in\mathcal{D}}(X_j-\bar{X})\underbrace{E[Z^+_j-\overline{Z^+}\mid X]}_{=E[Z^+_j-\overline{Z^+}]=0}\biggr]=0\
\end{align*}
due to the unbiasedness of the empirical mean.

Next, recall that by assumption $Z^+_i=\mu_{z^+}+\xi_i$, where $\xi_i\overset{i.i.d.}{\sim}\mathcal{N}(0,1)$ implicating \mbox{$\overline{Z^+}=\mu_{Z^+}+\frac{1}{|\mathcal{D}|}\sum_{i\in\mathcal{D}}\xi_i$}. Hence, we can rewrite $Cov[\overline{Z^+},\hat{\alpha}_1]$ as follows:

\begin{align*}
    Cov[\overline{Z^+},\hat{\alpha}_1]
    =&Cov\biggl[\overline{Z^+},\frac{1}{(\mid\mathcal{D}\mid -1)\hat{\sigma}_X^2}\sum_{j\in\mathcal{D}}(X_j-\bar{X})(Z^+_j-\overline{Z^+})\biggr]\\
    =&Cov\biggl[\bar{\xi},\frac{1}{(\mid\mathcal{D}\mid -1)\hat{\sigma}_X^2}\sum_{j\in\mathcal{D}}(X_j-\bar{X})(\xi_j-\bar{\xi})\biggr]\\
    =&E\biggl[\bar{\xi}\frac{1}{(\mid\mathcal{D}\mid -1)\hat{\sigma}_X^2}\sum_{j\in\mathcal{D}}(X_j-\bar{X})(\xi_j-\bar{\xi})\biggr]\\
    &\;-\underbrace{\underbrace{E[\bar{\xi}]}_{=0}E\biggl[\frac{1}{(\mid\mathcal{D}\mid -1)\hat{\sigma}_X^2}\sum_{j\in\mathcal{D}}(X_j-\bar{X})(\xi_j-\bar{\xi})\biggr]}_{=0}\\
    =&\sum_{j\in\mathcal{D}}E\biggl[\biggl(\underbrace{\frac{X_j}{(\mid\mathcal{D}\mid -1)\hat{\sigma}_X^2}}_{=:\tilde{X}_j}-\frac{1}{|\mathcal{D}|}\sum_{i\in\mathcal{D}}\underbrace{\frac{X_i}{(\mid\mathcal{D}\mid -1)\hat{\sigma}_X^2}}_{=:\tilde{X}_i}\biggr)(\xi_j-\bar{\xi})\bar{\xi}\biggr]\\=&\sum_{j\in\mathcal{D}}\underbrace{E\biggl[\biggl(\tilde{X}_j-\frac{1}{|\mathcal{D}|}\sum_{i\in\mathcal{D}}\tilde{X}_i\biggr)(\xi_j-\bar{\xi})\bar{\xi}\biggr]}_{\underset{X\Indep\xi}{=}\;\underbrace{E[\tilde{X}_j-\bar{\tilde{X}}]}_{=0}E[(\xi_j-\bar{\xi})\bar{\xi}]}=0
\end{align*}
The last line follows due to centrality of $\xi$, $X\Indep\xi$ and unbiasedness of the empirical mean. Last, it remains to show that $E[\bar{X}\hat{\alpha}_1\overline{Z^+}]$ is zero.
\begin{align*}
    E[\bar{X}\hat{\alpha}_1\overline{Z^+}]
    =&E\biggl[\bar{X}\frac{1}{(\mid\mathcal{D}\mid-1)\hat{\sigma}_X^2}\sum_{j\in\mathcal{D}}(X_j-\bar{X})(Z^+_j-\overline{Z^+})\overline{Z^+}\biggr]\\
    =&\sum_{j\in\mathcal{D}}E\biggl[\frac{\bar{X}}{(\mid\mathcal{D}\mid-1)\hat{\sigma}_X^2}(X_j-\bar{X})(Z^+_j-\overline{Z^+})\overline{Z^+}\biggr]\\
    =&\sum_{j\in\mathcal{D}}\underbrace{E\biggl[\frac{\bar{X}}{(\mid\mathcal{D}\mid-1)\hat{\sigma}_X^2}(X_j-\bar{X})\biggr]}_{=E\biggl[\underbrace{\frac{X_j\bar{X}}{(\mid\mathcal{D}\mid-1)\hat{\sigma}_X^2}}_{=:\check{X}_j}-\frac{1}{|\mathcal{D}|}\sum_{i\in\mathcal{D}}\underbrace{\frac{X_i\bar{X}}{(\mid\mathcal{D}\mid-1)\hat{\sigma}_X^2}}_{=:\check{X}_i}\biggr]}E[(Z^+_j-\overline{Z^+})\overline{Z^+}]\\
    =&\sum_{j\in\mathcal{D}}\underbrace{E[\check{X}_j-\bar{\check{X}}]}_{=0}E[(Z^+_j-\overline{Z^+})\overline{Z^+}]=0\;,
\end{align*}
Here, we used $X\Indep Z^+$ and the unbiasedness of the empirical mean, which concludes the proof.
\end{proof}

\section{Experimental Details}
\label{sec:appexperiments}

In the following section, we show additional experiments to support the results discussed in \Cref{sec:experiments}.

Simulations were run in \textit{R (version 4.4.3)}.
\begin{itemize}[label={}]
    \item R Core Team (2025). R: A Language and Environment for Statistical Computing. \mbox{R Foundation for Statistical Computing}, Vienna,
  Austria. \mbox{\url{https://www.R-project.org/}}.
  \begin{itemize}
      \item \textit{ggplot2} was used for data visualization.
        \begin{itemize}[label={}]
          \item Wickham, H. (2016), ggplot2: Elegant Graphics for Data Analysis. Springer-Verlag New York.
      \end{itemize}
      \item \textit{glmnet} was used to perform ridge regression. 
       \begin{itemize}[label={}]
          \item Friedman J., Tibshirani R., Hastie T. (2010). Regularization Paths for Generalized Linear Models via Coordinate Descent. Journal of Statistical Software, 33(1), 1-22.
       \end{itemize}
  \end{itemize}
  \end{itemize}

\subsection{Supplementary Information to \Cref{sec:expvariance}}
\label{sec:appexpvariance}

Below, in Figures~\ref{fig:QuadVarBoxArea}~and~\ref{fig:LinVarBoxArea} we show the results for the quadratic and the linear model for RR and TSR as well as its versions with ridge penalty for the first of the two regression steps, respectively. For the boxplots, we differentiated between two cases. One is based on the biased dataset $\mathcal{S}$ and the other on the unbiased dataset $\mathcal{D}$. On the right hand side, we show the 95\%-areas and mean of the estimations. For completeness, we accompany these results by providing the numerical values for them in \Cref{tab:var_quadratic} and \Cref{tab:var_linear}.

The mean and standard deviation of the MSE remain smaller in $\mathcal{S}$ than $\mathcal{D}$, as one would expect, since the first regression was performed based only on the data underlying selection. For TSR and RR, the mean and standard deviation of MSE decrease for increasing $n$. For the naive estimator, this effect is not that pronounced. Mean and standard deviation of MSE are smaller for TSR than for RR, whereby the difference also vanishes when $n$ increases. Furthermore, we observe that mean and standard deviation of the MSE of the RR and TSR estimator do not differ distinctly between OLS and ridge regression, respectively. Last, there were no clear differences between $\mathcal{S}\cap\mathcal{D}=\emptyset$ and $\mathcal{S}\subset\mathcal{D}$ recognizable. 

\begin{figure}[h!]
\caption*{$\mathcal{S}\subset\mathcal{D}$}

   \begin{minipage}[b]{.3\linewidth}
      \caption*{$\mathcal{D}$}
      \vspace{-0.5em}
        \includegraphics[scale=0.18]{Fig3a_8a.png}
   \end{minipage}
   \begin{minipage}[b]{.3\linewidth} 
   \caption*{$\mathcal{S}$}
   \vspace{-0.5em}
      \includegraphics[scale=0.18]{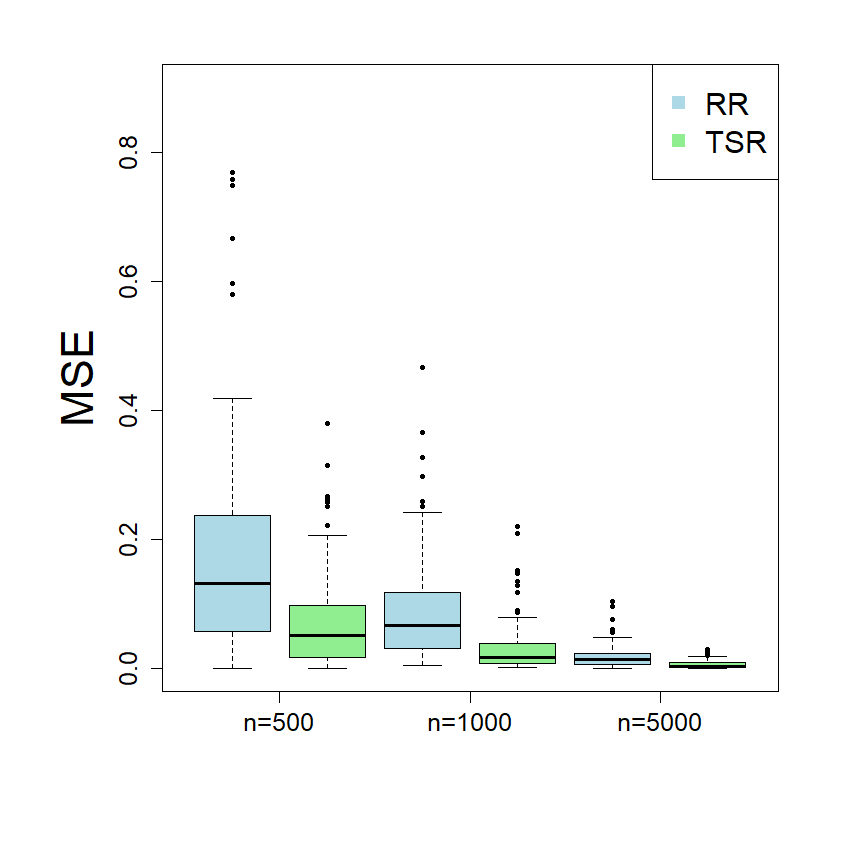}
   \end{minipage}
   \begin{minipage}[b]{.4\linewidth} 
       \includegraphics[width=1\linewidth]{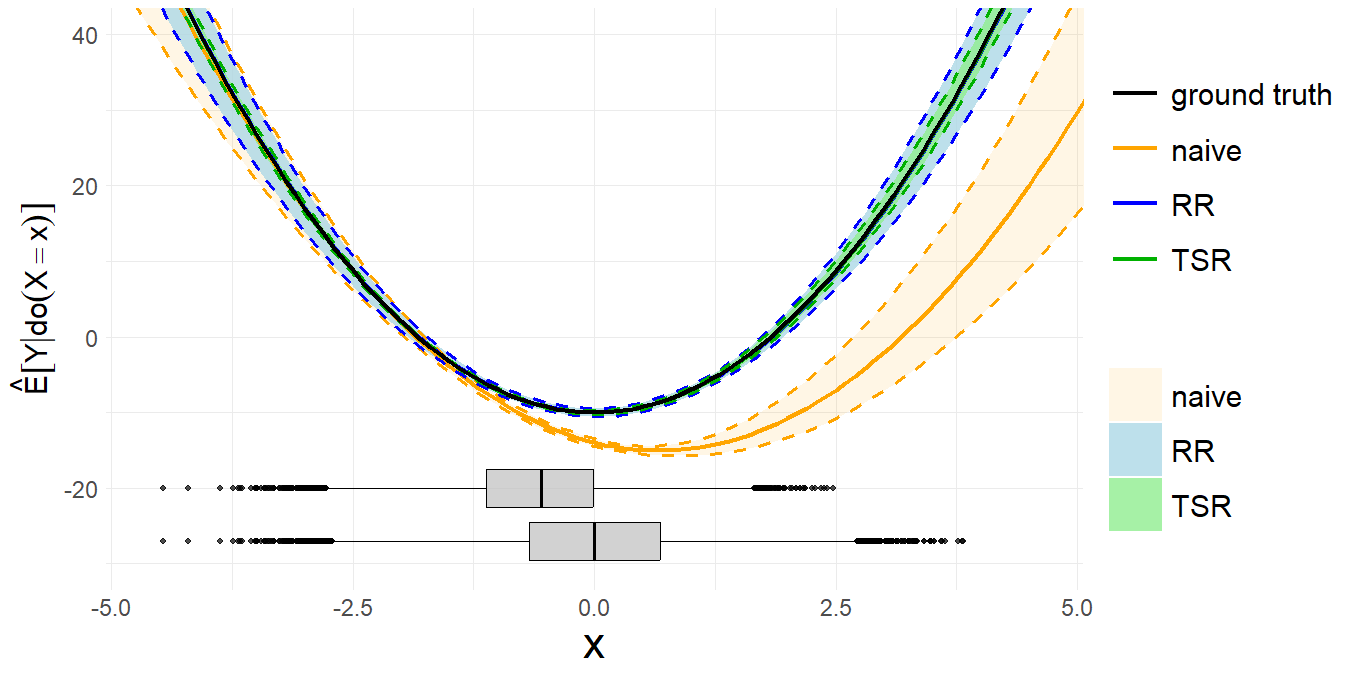}
       \caption*{Estimates of $E[Y|do(X=x)]$}
   \end{minipage}
   \begin{minipage}[b]{.3\linewidth}
   \vspace{-0.5em}
      \includegraphics[scale=0.18]{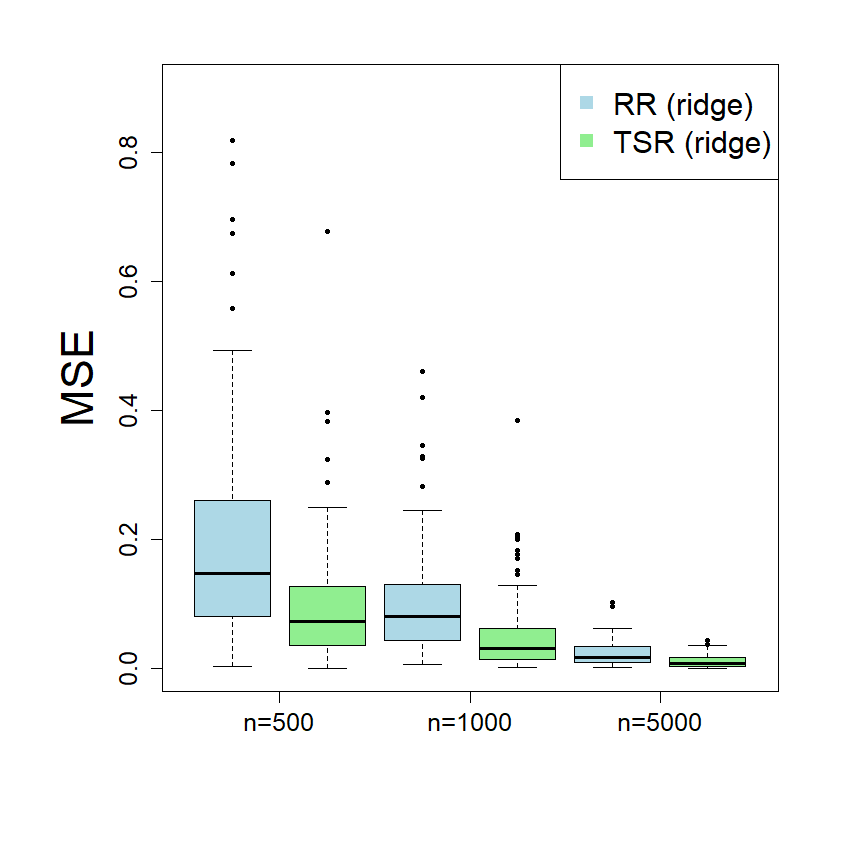}
   \end{minipage}
   \begin{minipage}[b]{.3\linewidth} 
   \vspace{-0.5em}
      \includegraphics[scale=0.18]{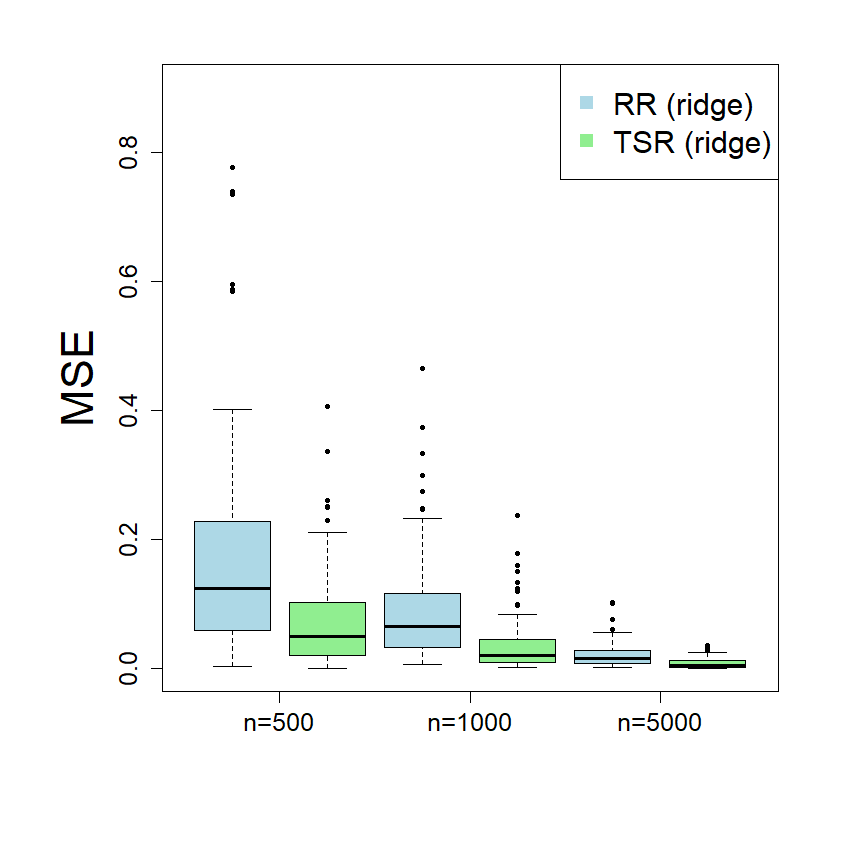}
   \end{minipage}
   \begin{minipage}[b]{.4\linewidth} 
       \includegraphics[width=1\linewidth]{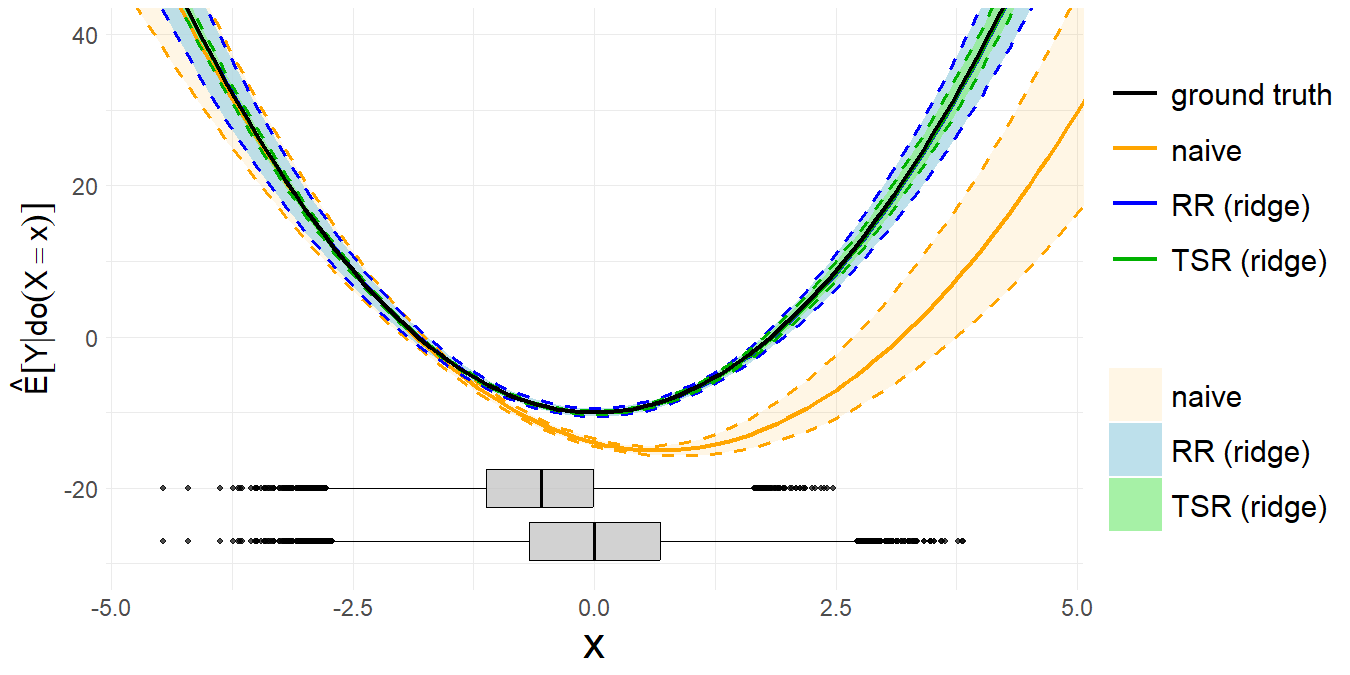}
       \caption*{Estimates of $E[Y|do(X=x)]$}
   \end{minipage}
   \caption*{$\mathcal{S}\cap\mathcal{D}=\emptyset$}
   \begin{minipage}[b]{.3\linewidth}
   \caption*{$\mathcal{D}$}
   \vspace{-0.5em}
      \includegraphics[scale=0.18]{Fig3b_8g.png}
   \end{minipage}
   \begin{minipage}[b]{.3\linewidth} 
   \caption*{$\mathcal{S}$}
   \vspace{-0.5em}
      \includegraphics[scale=0.18]{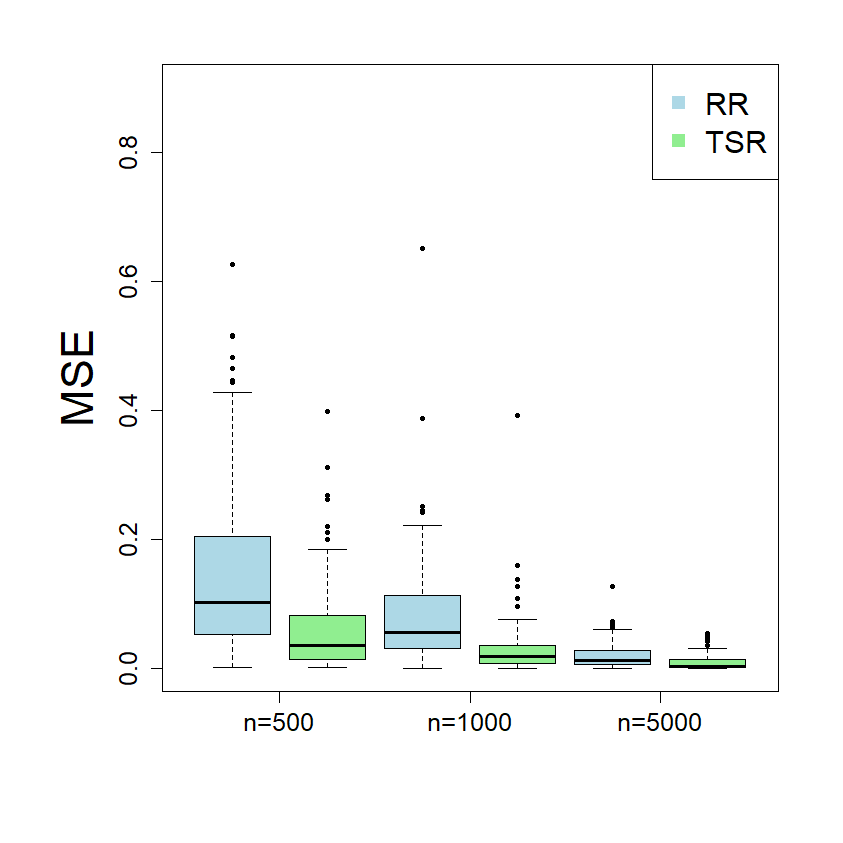}
   \end{minipage}
   \begin{minipage}[b]{.4\linewidth} 
      \includegraphics[width=1\linewidth]{Fig3c_8i.png} 
      \caption*{Estimates of $E[Y|do(X=x)]$}
   \end{minipage}
   \begin{minipage}[b]{.3\linewidth}
   \vspace{-0.5em}
       \includegraphics[scale=0.18]{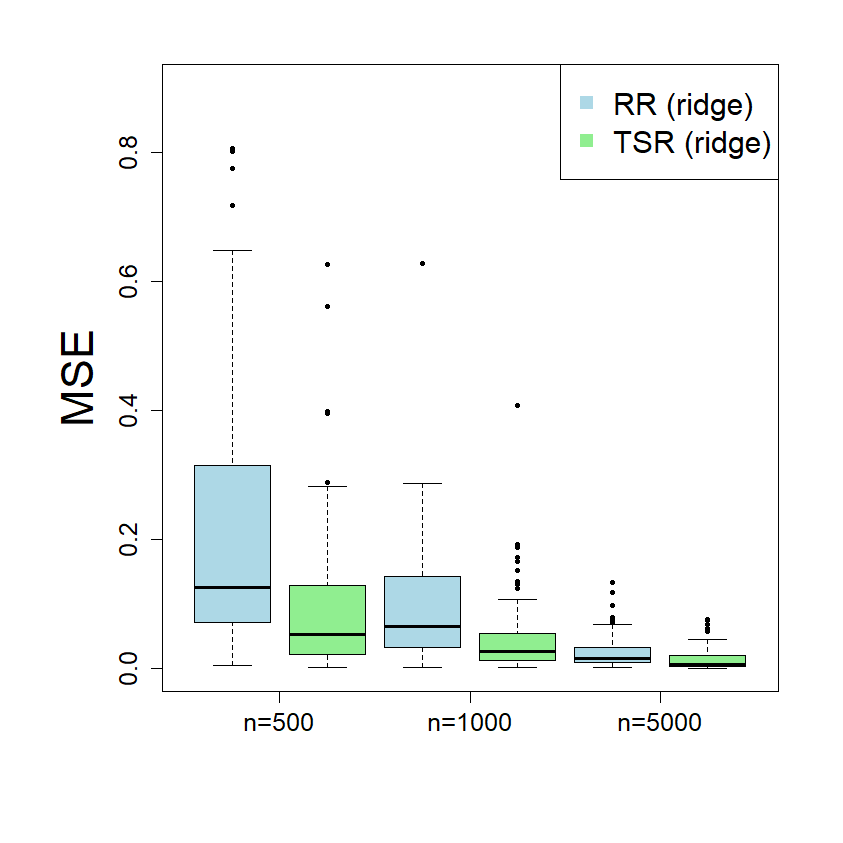}
   \end{minipage}
   \begin{minipage}[b]{.3\linewidth} 
   \vspace{-0.5em}
      \includegraphics[scale=0.18]{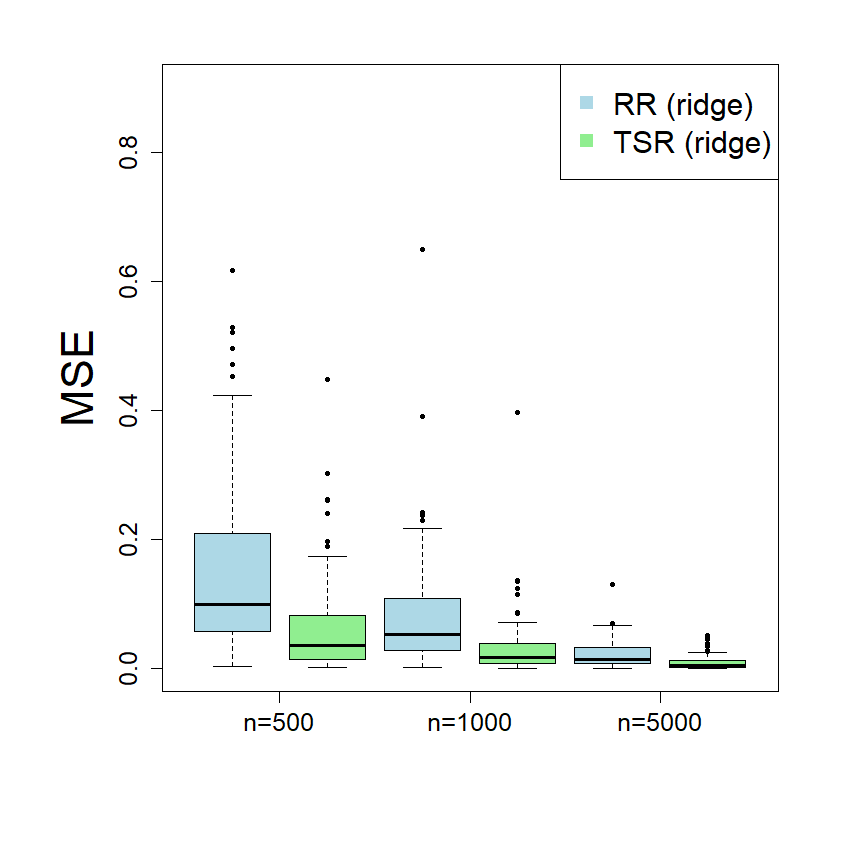}
   \end{minipage}
   \begin{minipage}[b]{.4\linewidth} 
      \includegraphics[width=1\linewidth]{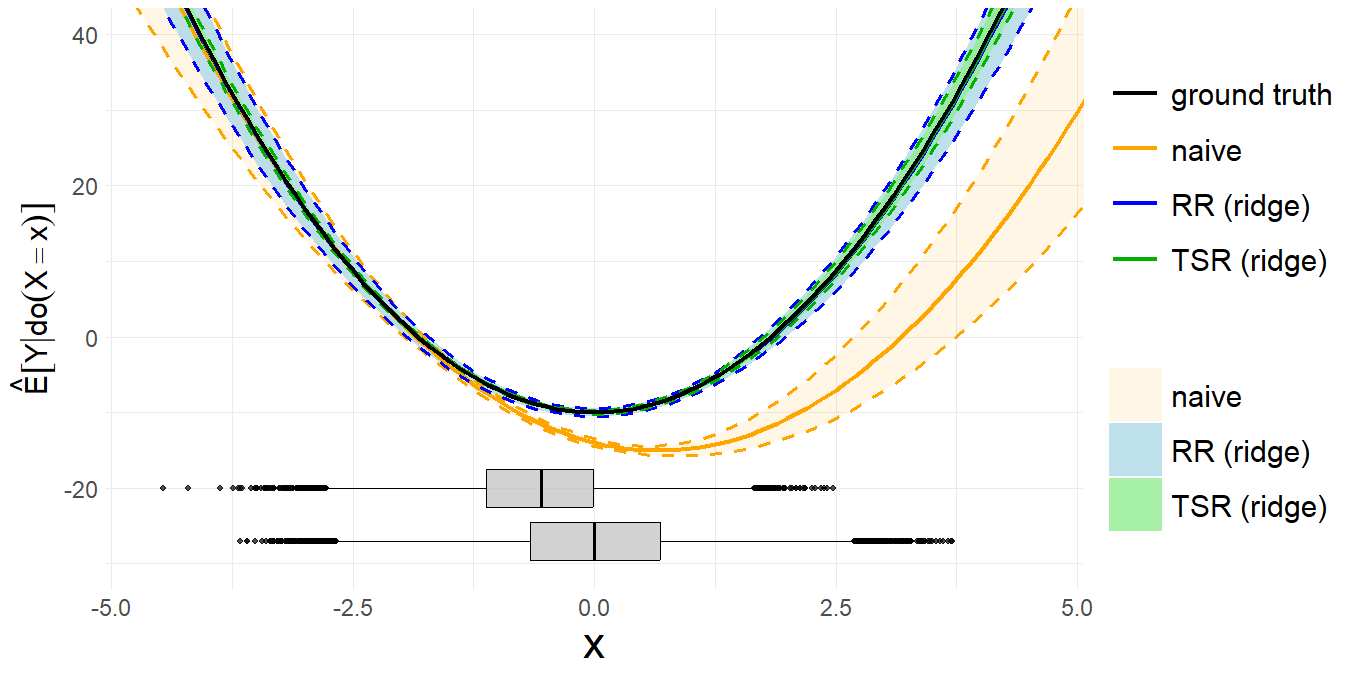}
      \caption*{Estimates of $E[Y|do(X=x)]$}
   \end{minipage}
   \caption{Left: Boxplots of the MSE  over $\mathcal{D}$ and $\mathcal{S}$ (top: $\mathcal{S}\subset\mathcal{D}$, bottom: $\mathcal{S}\cap\mathcal{D}=\emptyset$) of RR and TSR for $n\in\{500,1000,5000\}$ for the \textbf{quadratic} case. On the right, we show in black the ground truth value of $E[Y|do(X=x)]$, and colored the mean estimates and $95\%$-areas over all simulations for naive, RR, and TSR for $n=500$ and with the coefficients fixed at their means (for $\mathcal{S}\cap\mathcal{D}=\emptyset$). We see that RR and TSR recover the ground truth whereas the $95\%$-area of TSR more narrowly contains the ground truth. The boxplots on the x-axis show the distribution of $\Xv$ in $\mathcal{S}$ (upper) and $\mathcal{D}$ (lower).}
  \label{fig:QuadVarBoxArea}
\end{figure}
\begin{figure}[h!]
\caption*{$\mathcal{S}\subset\mathcal{D}$}

   \begin{minipage}[b]{.3\linewidth}
   \caption*{$\mathcal{D}$}
   \vspace{-0.5em}
      \includegraphics[scale=0.18]{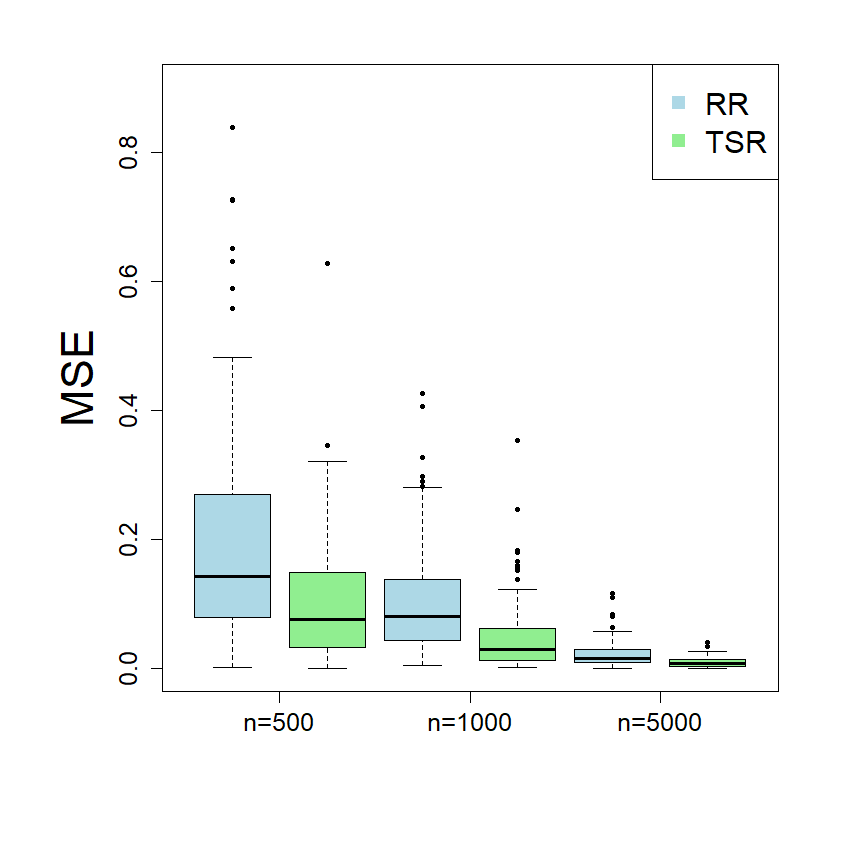}
   \end{minipage}
   \begin{minipage}[b]{.3\linewidth} 
   \caption*{$\mathcal{S}$}
   \vspace{-0.5em}
   \includegraphics[scale=0.18]{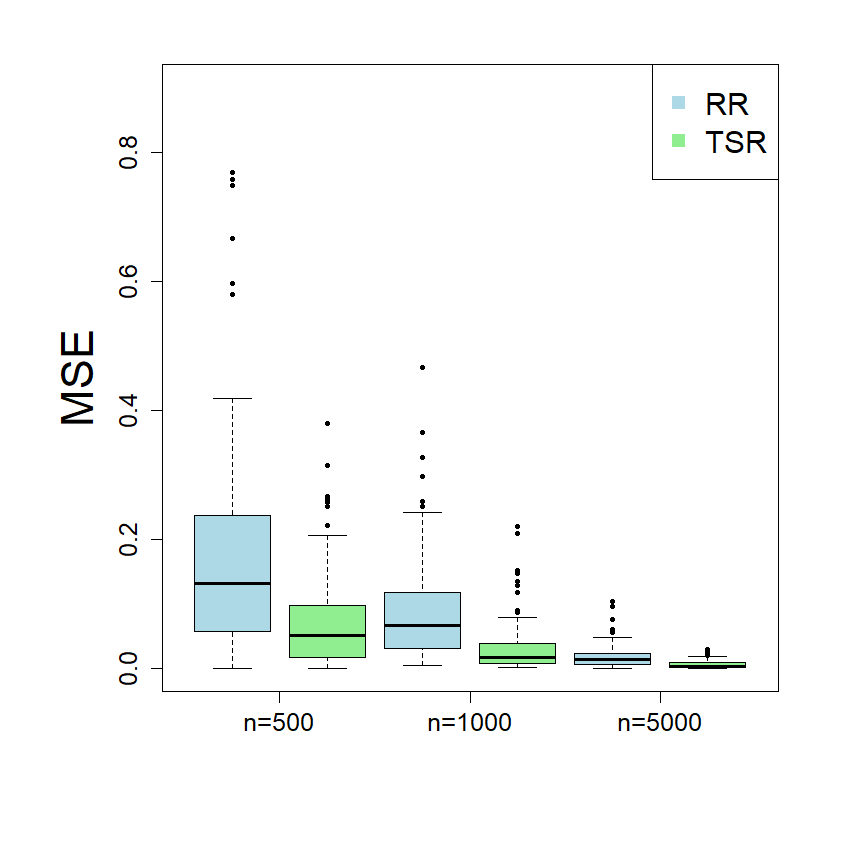}
   \end{minipage}
   \begin{minipage}[b]{.4\linewidth} 
      \includegraphics[width=1\linewidth]{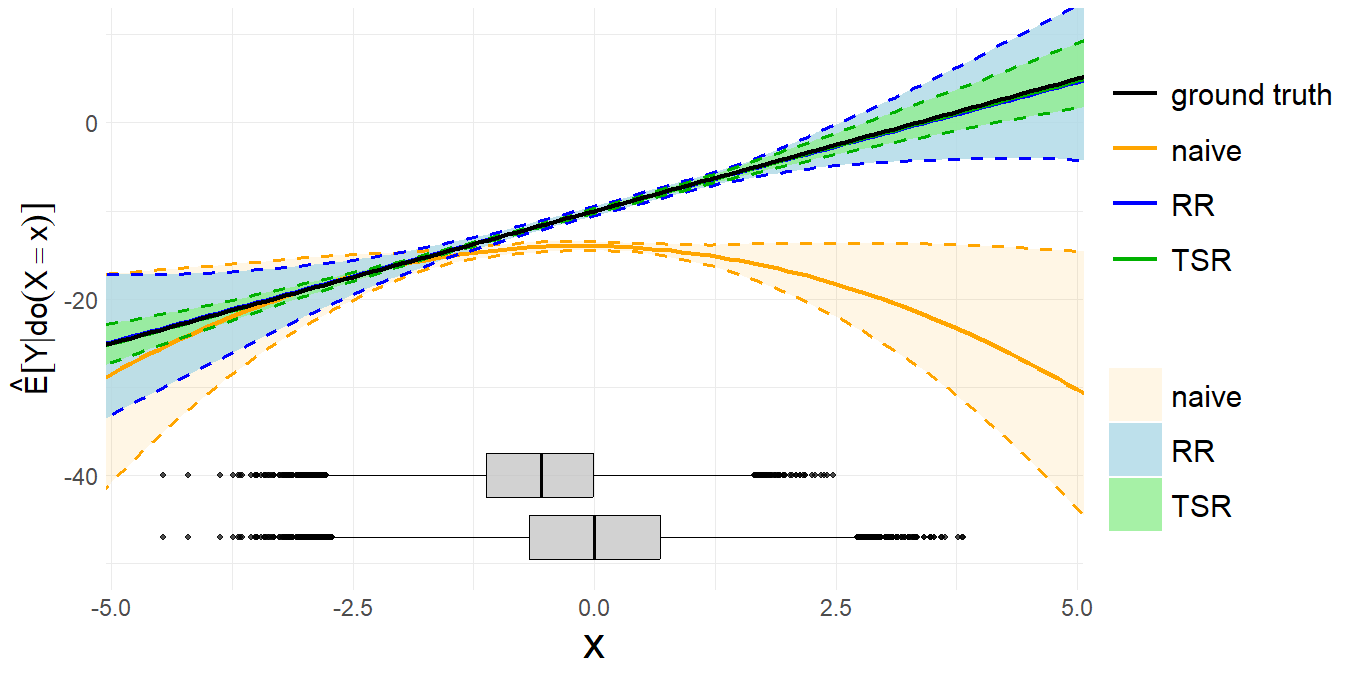}
      \caption*{Estimates of $E[Y|do(X=x)]$}
   \end{minipage}
   \begin{minipage}[b]{.3\linewidth}
   \vspace{-0.5em}
       \includegraphics[scale=0.18]{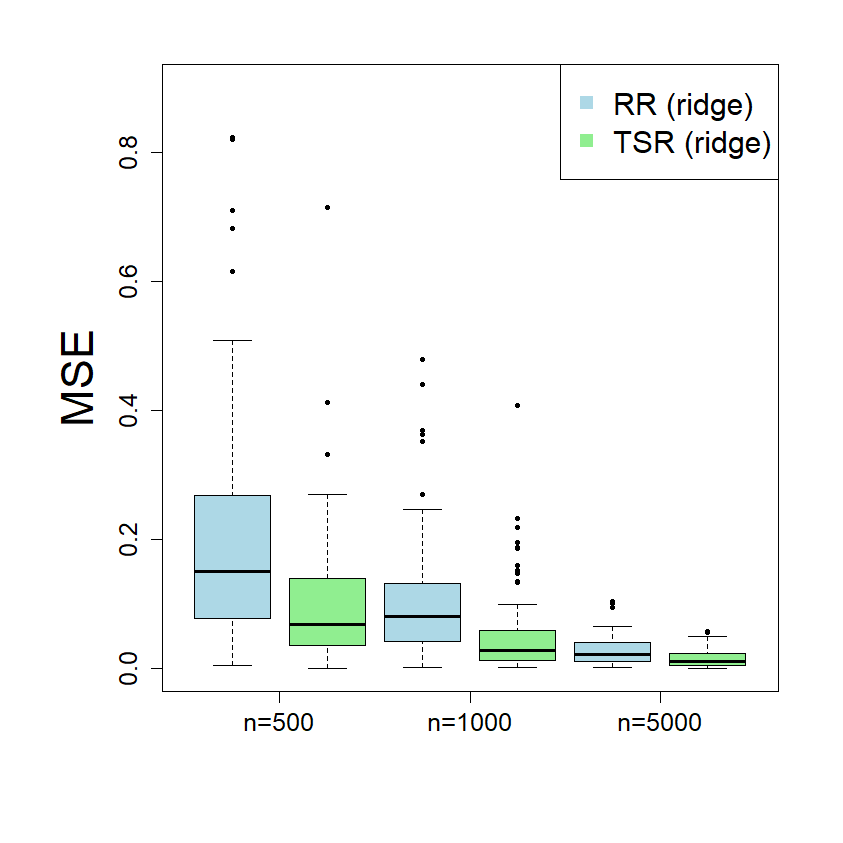}
   \end{minipage}
   \begin{minipage}[b]{.3\linewidth} 
   \vspace{-0.5em}
       \includegraphics[scale=0.18]{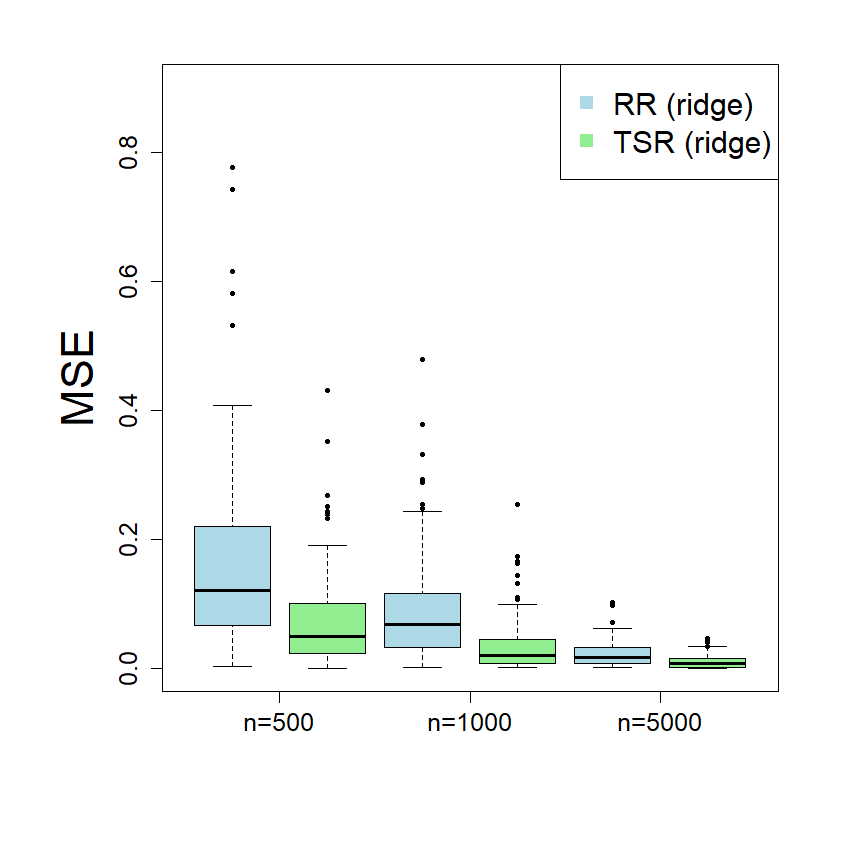}
   \end{minipage}
   \begin{minipage}[b]{.4\linewidth} 
       \includegraphics[width=1\linewidth]{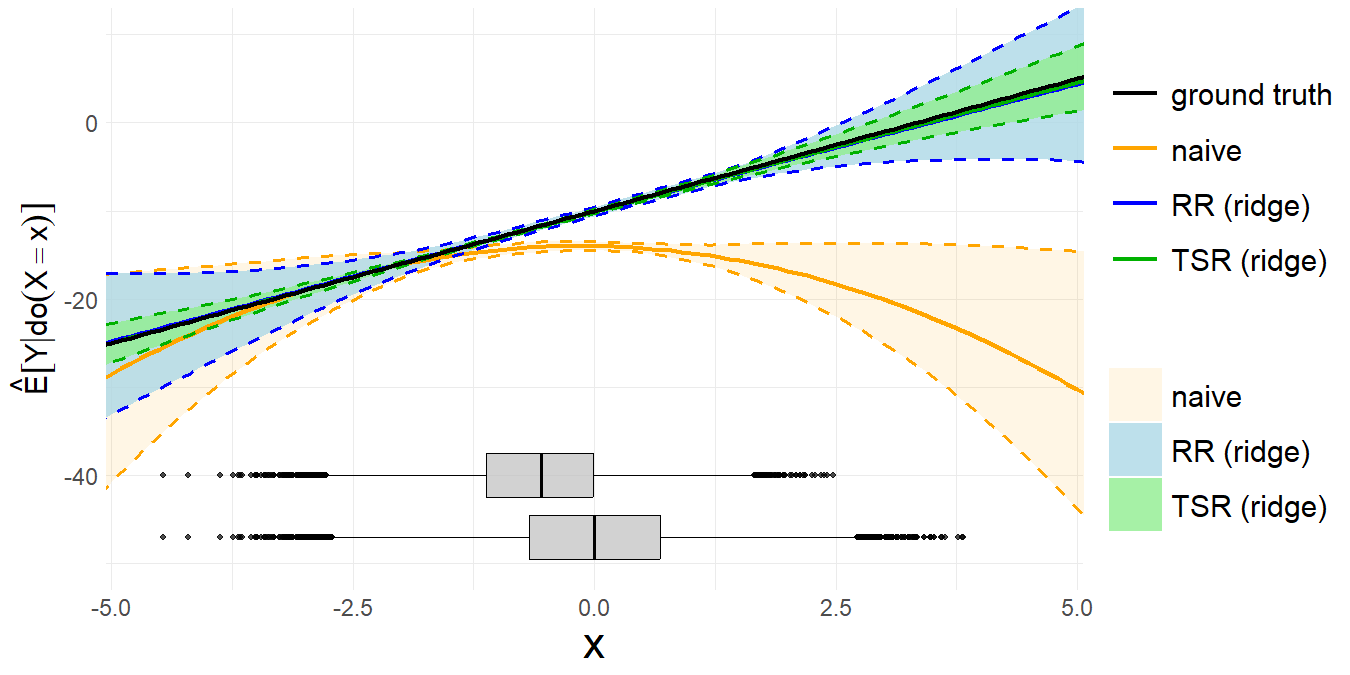}
       \caption*{Estimates of $E[Y|do(X=x)]$}
   \end{minipage}
   \caption*{$\mathcal{S}\cap\mathcal{D}=\emptyset$}
   \begin{minipage}[b]{.3\linewidth}
   \caption*{$\mathcal{D}$}
   \vspace{-0.5em}
      \includegraphics[scale=0.18]{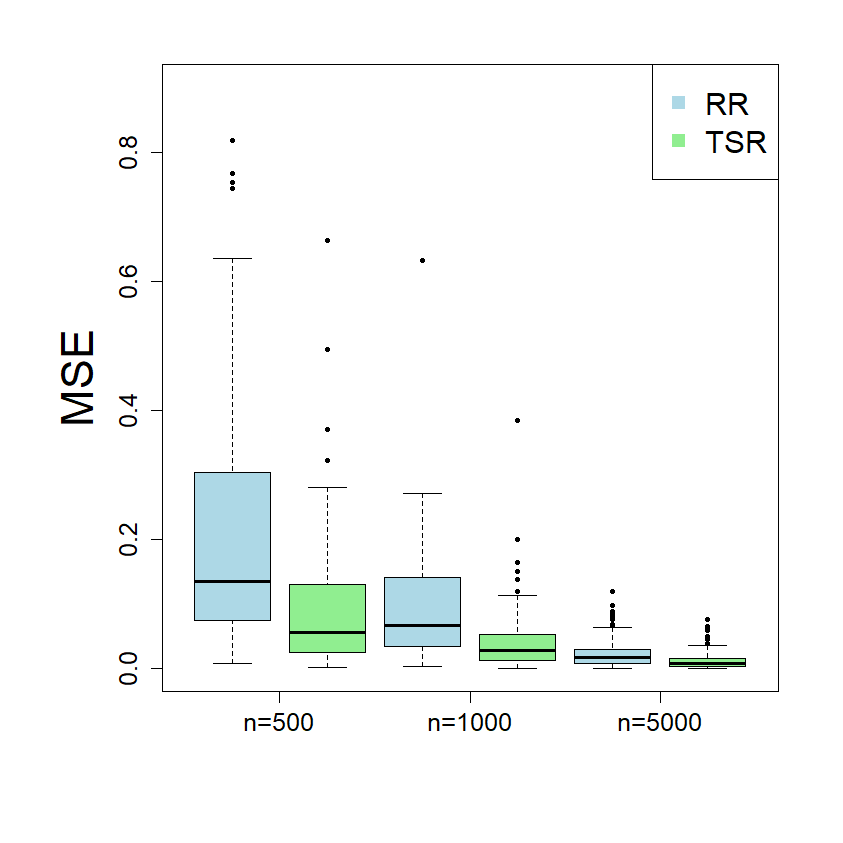}
   \end{minipage}
   \begin{minipage}[b]{.3\linewidth} 
   \caption*{$\mathcal{S}$}
   \vspace{-0.5em}
   \includegraphics[scale=0.18]{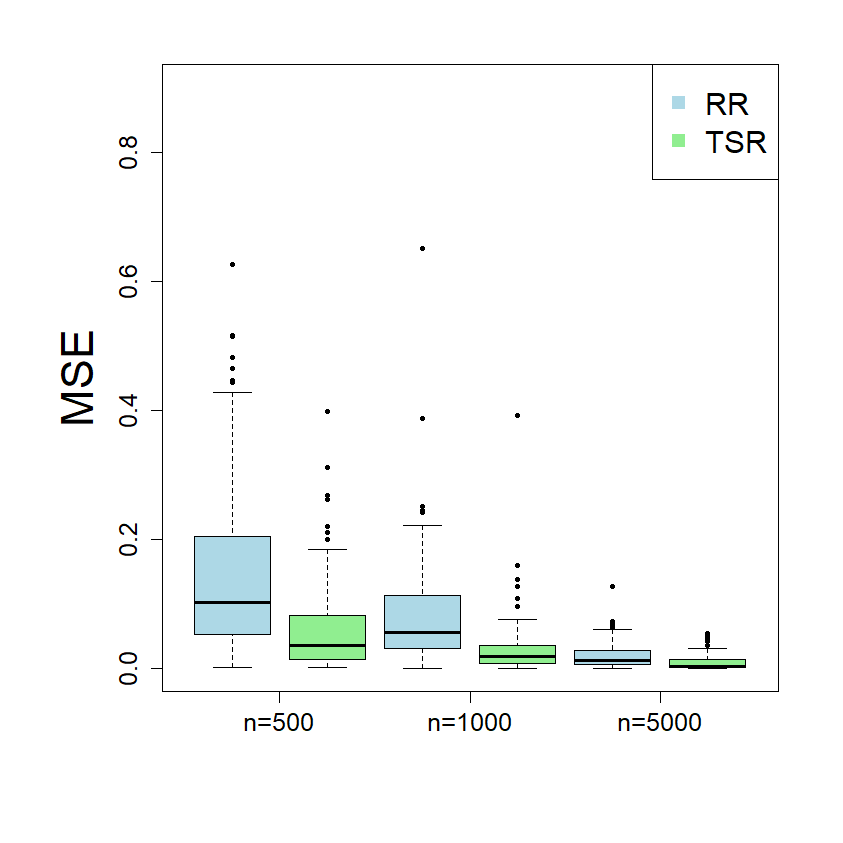}
   \end{minipage}
   \begin{minipage}[b]{.4\linewidth} 
      \includegraphics[width=1\linewidth]{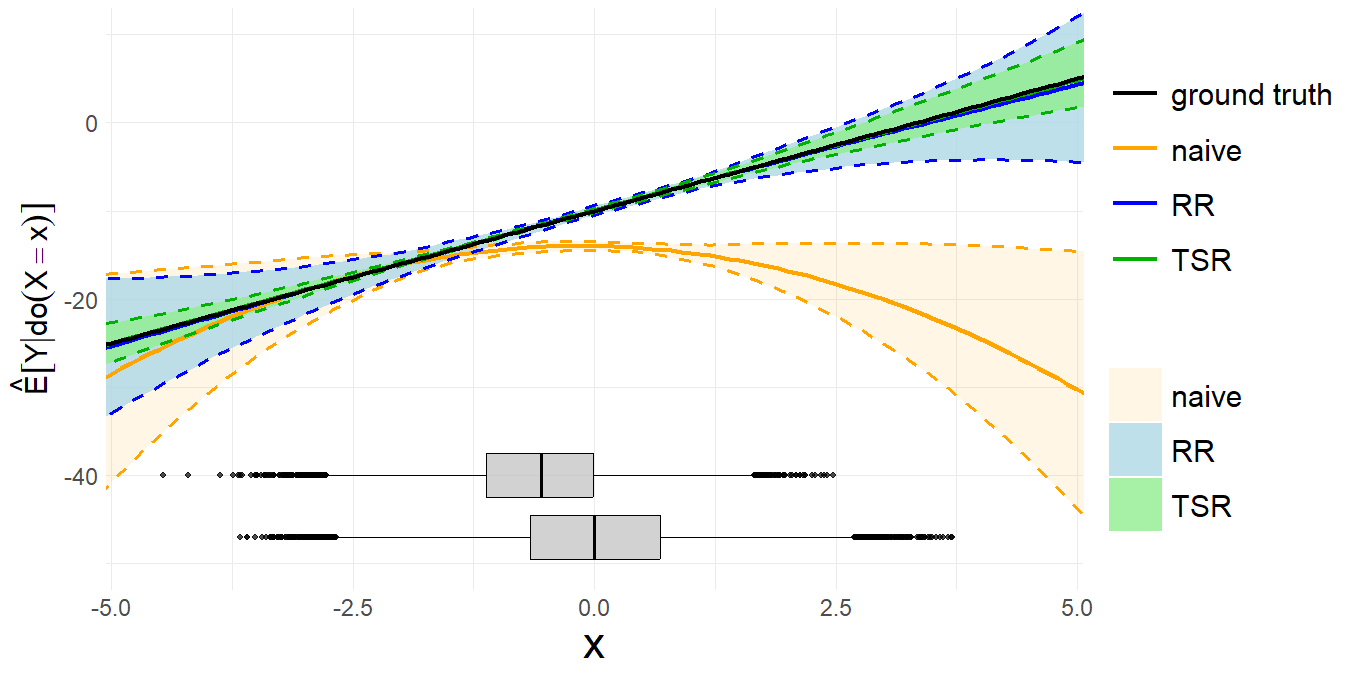}
      \caption*{Estimates of $E[Y|do(X=x)]$}
   \end{minipage}
   \begin{minipage}[b]{.3\linewidth}
   \vspace{-0.5em}
      \includegraphics[scale=0.18]{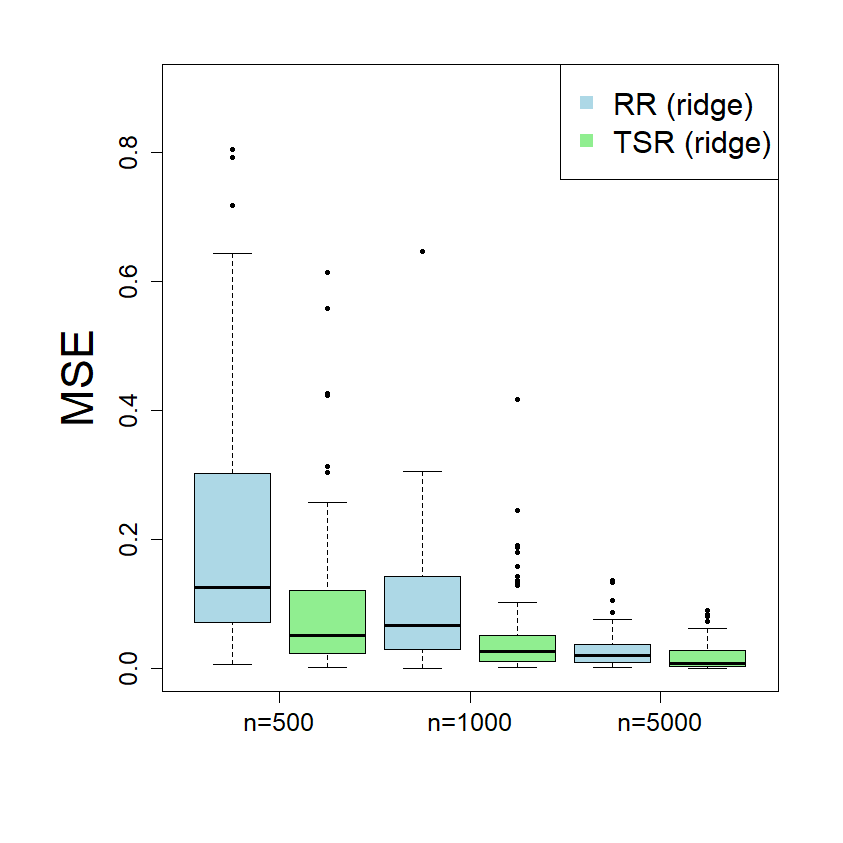}
   \end{minipage}
   \begin{minipage}[b]{.3\linewidth} 
   \vspace{-0.5em}
      \includegraphics[scale=0.18]{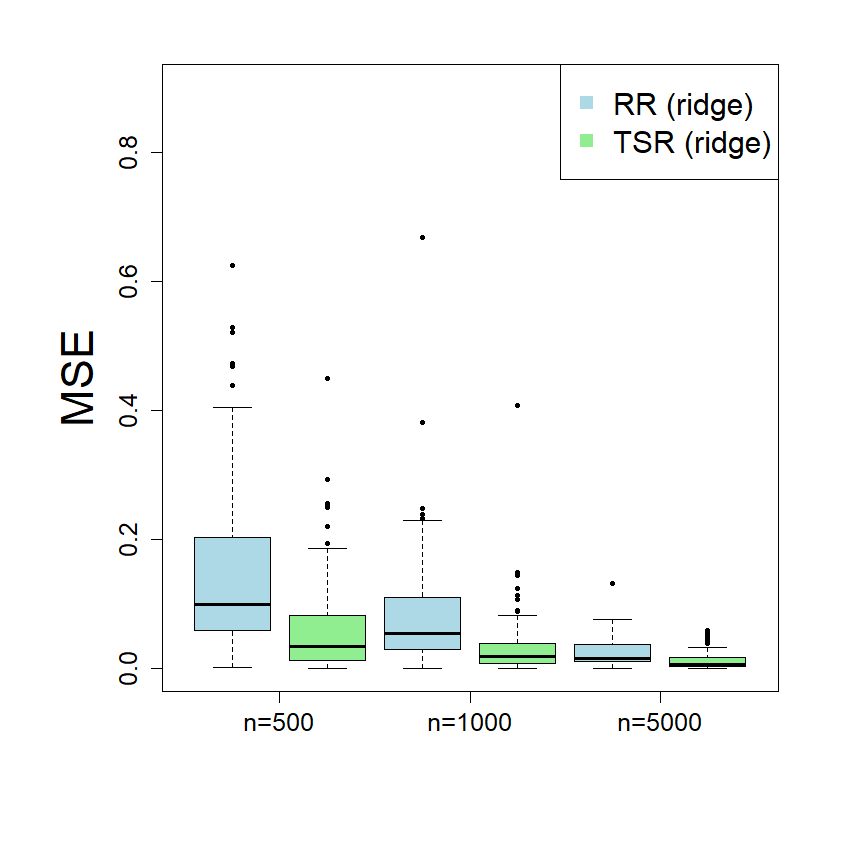}
   \end{minipage}
   \begin{minipage}[b]{.4\linewidth} 
       \includegraphics[width=1\linewidth]{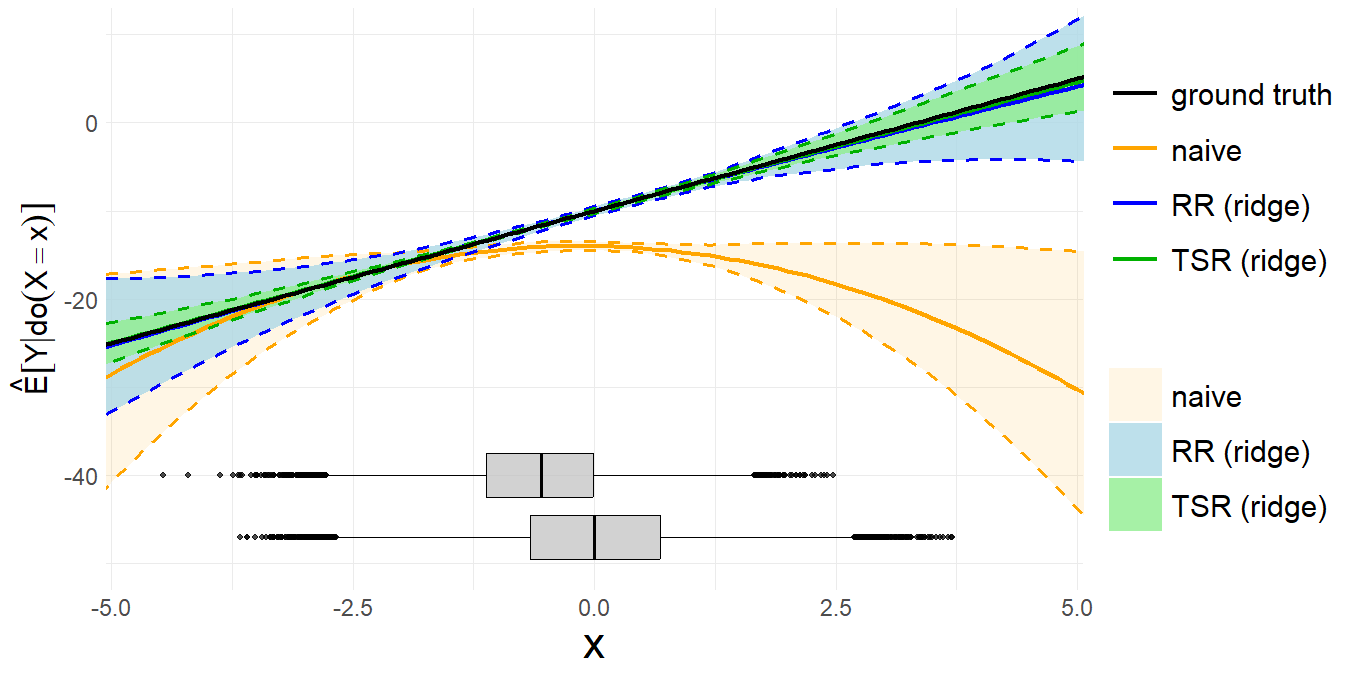}
       \caption*{Estimates of $E[Y|do(X=x)]$}
   \end{minipage}
   \caption{Left: Boxplots of the MSE  over $\mathcal{D}$ and $\mathcal{S}$ (top: $\mathcal{S}\subset\mathcal{D}$, bottom: $\mathcal{S}\cap\mathcal{D}=\emptyset$) of RR and TSR for $n\in\{500,1000,5000\}$ for the \textbf{linear} case. On the right, we show in black the ground truth value of $E[Y|do(X=x)]$, and colored the mean estimates and $95\%$-areas over all simulations for naive, RR, and TSR for $n=500$ and with the coefficients fixed at their means (for $\mathcal{S}\cap\mathcal{D}=\emptyset$). We see that RR and TSR recover the ground truth whereas the $95\%$-area of TSR more narrowly contains the ground truth. The boxplots on the x-axis show the distribution of $\Xv$ in $\mathcal{S}$ (upper) and $\mathcal{D}$ (lower).}
\label{fig:LinVarBoxArea}
\end{figure}

\begin{table}[h!]
\caption{\textbf{Quadratic model}: mean (sd) of MSE over $\mathcal{S}$ and $\mathcal{D}$.}\vspace{0.4cm}
    \centering
    \tiny
    \begin{tabular}{lcccccc}
    &\multicolumn{6}{c}{$\mathcal{S}$}\\
    \cmidrule(rl){2-7}
    &\multicolumn{3}{c}{$\mathcal{S}\subset\mathcal{D}$}&\multicolumn{3}{c}{$\mathcal{S}\cap\mathcal{D}=\emptyset$}\\
&$n=500$&$n=1000$&$n=5000$&$n=500$&$n=1000$&$n=5000$\\
    \cmidrule(rl){2-4} \cmidrule(rl){5-7}
         naive& 13.13 (6.62) & 12.85 (6.10) & 13.27 (6.02) 
         & 13.39 (6.80) & 12.77 (5.85) & 13.22 (5.97) \\
         RR& 0.18 (0.21) & 0.09 (0.08) & 0.02 (0.02) 
         & 0.16 (0.18)& 0.08 (0.09) & 0.02 (0.02) \\
         RR (ridge)& 0.18 (0.20) & 0.09 (0.09) & 0.02 (0.02) 
         & 0.16 (0.18) & 0.08 (0.09) & 0.02 (0.02) \\
         TSR& 0.08 (0.08) & 0.03 (0.04) & 0.01 (0.01) 
         & 0.06 (0.07) & 0.03 (0.05) & 0.01 (0.01) \\
         TSR (ridge)& 0.07 (0.08) & 0.03 (0.04) & 0.01 (0.01) 
         & 0.06 (0.07) & 0.03 (0.05)& 0.01 (0.01)\\
    \end{tabular}
\vspace{0.4cm}
    \centering
    \begin{tabular}{lcccccc}
     &\multicolumn{6}{c}{$\mathcal{D}$}\\
    \cmidrule(rl){2-7}
    &\multicolumn{3}{c}{$\mathcal{S}\subset\mathcal{D}$}&\multicolumn{3}{c}{$\mathcal{S}\cap\mathcal{D}=\emptyset$}\\
    &$n=500$&$n=1000$&$n=5000$&$n=500$&$n=1000$&$n=5000$\\
    \cmidrule(rl){2-4} \cmidrule(rl){5-7}
         naive& 35.28 (19.46) & 33.05 (16.32) & 34.92 (17.29) 
         & 35.19 (19.89) & 33.42 (16.65) & 35.07 (17.52) \\
         RR& 0.20 (0.17) & 0.10 (0.08) & 0.02 (0.02) 
         & 0.23 (0.23) & 0.10 (0.09) & 0.02 (0.02) \\
         RR (ridge)& 0.19 (0.17) & 0.10 (0.09) & 0.02 (0.02) 
         & 0.22 (0.23) & 0.10 (0.09) & 0.03 (0.03) \\
         TSR& 0.10 (0.10) & 0.05 (0.06) & 0.01 (0.01) 
         & 0.09 (0.11)& 0.04 (0.05) & 0.01 (0.02) \\
         TSR (ridge)& 0.10 (0.10) & 0.05 (0.06)& 0.01 (0.01) 
         & 0.09 (0.11) & 0.04 (0.06) & 0.01 (0.02) \\
    \end{tabular}
    \label{tab:var_quadratic}
\end{table}
\begin{table}[h!]
 \caption{\textbf{Linear model}: mean (sd) of MSE over $\mathcal{S}$ and $\mathcal{D}$.}\vspace{0.4cm}
    \centering
    \tiny
    \begin{tabular}{lcccccc}
     &\multicolumn{6}{c}{$\mathcal{S}$}\\
    \cmidrule(rl){2-7}
    &\multicolumn{3}{c}{$\mathcal{S}\subset\mathcal{D}$}&\multicolumn{3}{c}{$\mathcal{S}\cap\mathcal{D}=\emptyset$}\\
    &$n=500$&$n=1000$&$n=5000$&$n=500$&$n=1000$&$n=5000$\\
        \cmidrule(rl){2-4} \cmidrule(rl){5-7}
         naive& 13.13 (6.62) & 12.85 (6.10) & 13.27 (6.02) 
         & 13.39 (6.80) & 12.77 (5.85) & 13.22 (5.97) \\
         RR& 0.18 (0.21) & 0.09 (0.08) & 0.02 (0.02) 
         & 0.16 (0.18) & 0.08 (0.09) & 0.02 (0.02) \\
         RR (ridge)& 0.18 (0.20) & 0.09 (0.09) & 0.02 (0.02) 
         & 0.16 (0.18) & 0.08 (0.09) & 0.02 (0.02) \\
         TSR& 0.08 (0.08) & 0.03 (0.04) & 0.01 (0.01)
         & 0.06 (0.07) & 0.03 (0.05) & 0.01 (0.01) \\
         TSR (ridge)& 0.08 (0.08) & 0.04 (0.04) & 0.01 (0.01)
         & 0.06 (0.08) & 0.03 (0.05) & 0.01 (0.01)\\
    \end{tabular}
\vspace{0.4cm}
    \centering
    \begin{tabular}{lcccccc}
     &\multicolumn{6}{c}{$\mathcal{D}$}\\
    \cmidrule(rl){2-7}
    &\multicolumn{3}{c}{$\mathcal{S}\subset\mathcal{D}$}&\multicolumn{3}{c}{$\mathcal{S}\cap\mathcal{D}=\emptyset$}\\
    &$n=500$&$n=1000$&$n=5000$&$n=500$&$n=1000$&$n=5000$\\
        \cmidrule(rl){2-4} \cmidrule(rl){5-7}
         naive& 35.28 (19.46) & 33.05 (16.32) & 34.92 (17.29) 
         & 35.19 (19.89) & 33.42 (16.65) & 35.07 (17.52) \\
         RR& 0.20 (0.17) & 0.10 (0.08) & 0.02 (0.02) 
         & 0.23 (0.23) & 0.10 (0.09) & 0.02 (0.02) \\
         RR (ridge)& 0.20 (0.17) & 0.10 (0.09) & 0.03 (0.02) 
         & 0.22 (0.23) & 0.10 (0.09) & 0.03 (0.03) \\
         TSR& 0.10 (0.06) & 0.05 (0.04) & 0.01 (0.01) 
         & 0.09 (0.11) & 0.04 (0.05) & 0.01 (0.02) \\
         TSR (ridge)& 0.10 (0.10) & 0.05 (0.06) & 0.02 (0.01) 
         & 0.10 (0.11) & 0.05 (0.06) & 0.02 (0.02)\\
    \end{tabular}
    \vspace{0.2cm}
    \label{tab:var_linear}
\end{table}

\subsection{Supplementary Information to \Cref{sec:expvarious}}
\label{sec:appexpvarious}

\begin{figure}[h!]
   \begin{minipage}[b]{.26\linewidth}
      \includegraphics[width=1\linewidth]{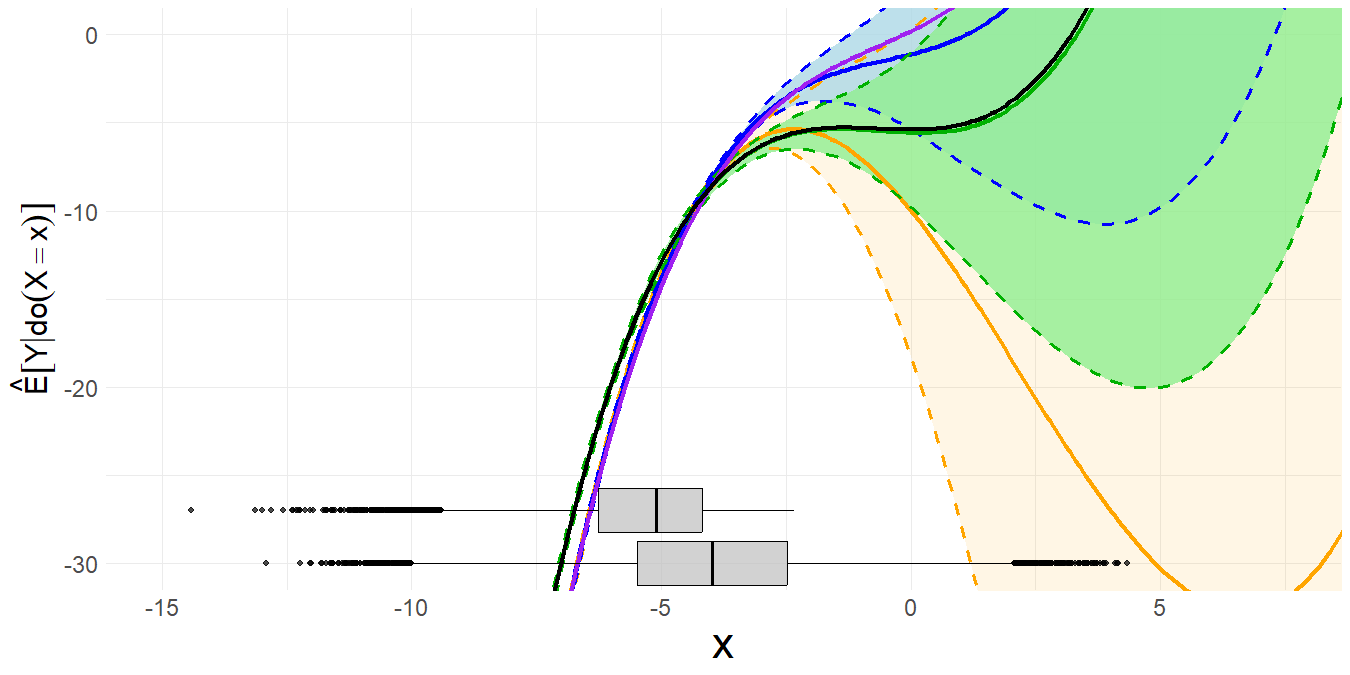}
      \caption*{$n=500$}
   \end{minipage}
   \hspace{.01\linewidth}
   \begin{minipage}[b]{.26\linewidth} 
      \includegraphics[width=1\linewidth]{Fig4a_10b.png}
      \caption*{$n=1000$}
   \end{minipage}
   \hspace{.01\linewidth}
   \begin{minipage}[b]{.26\linewidth} 
      \includegraphics[width=1\linewidth]{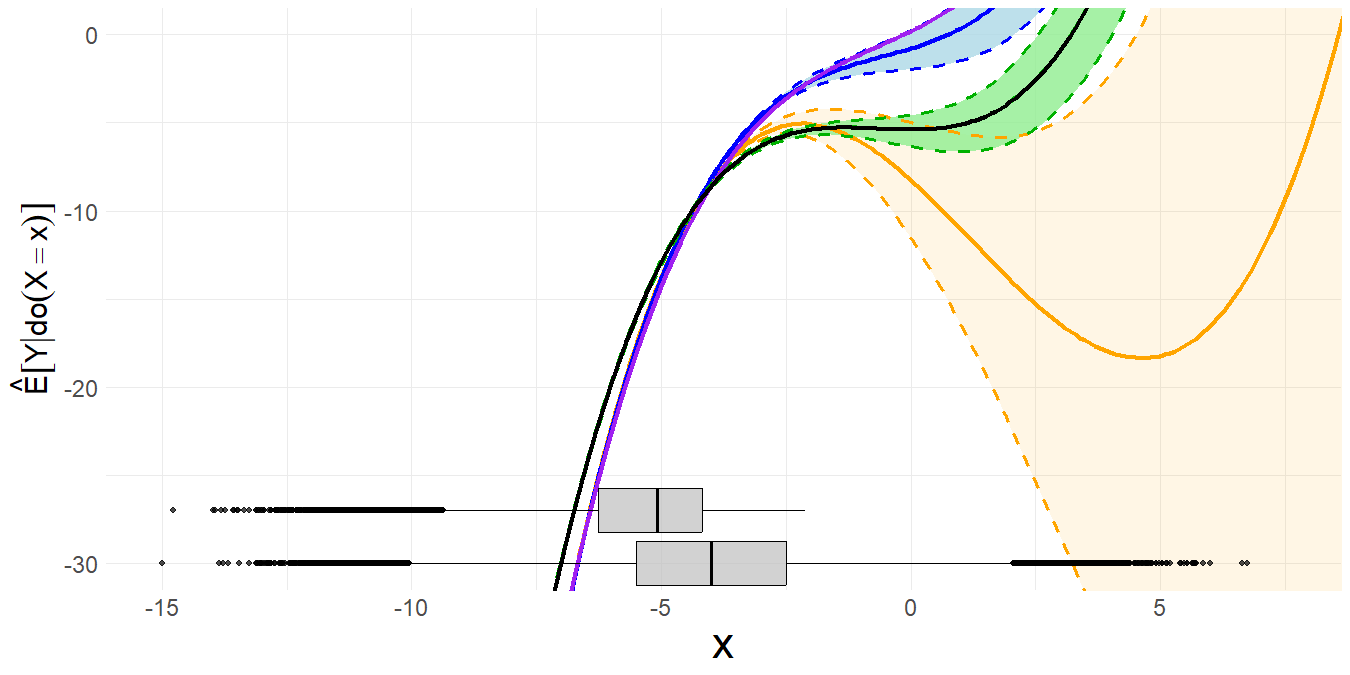}
      \caption*{$n=5000$}
   \end{minipage}
   \hspace{.01\linewidth}
   \begin{minipage}[b]{.07\linewidth} 
      \includegraphics[scale=0.11]{Fig4d_4h_10d_11d_12d_13d_14d_15d.png}
      \caption*{}
   \end{minipage}\\
   \begin{minipage}[b]{.26\linewidth}
      \includegraphics[width=1\linewidth]{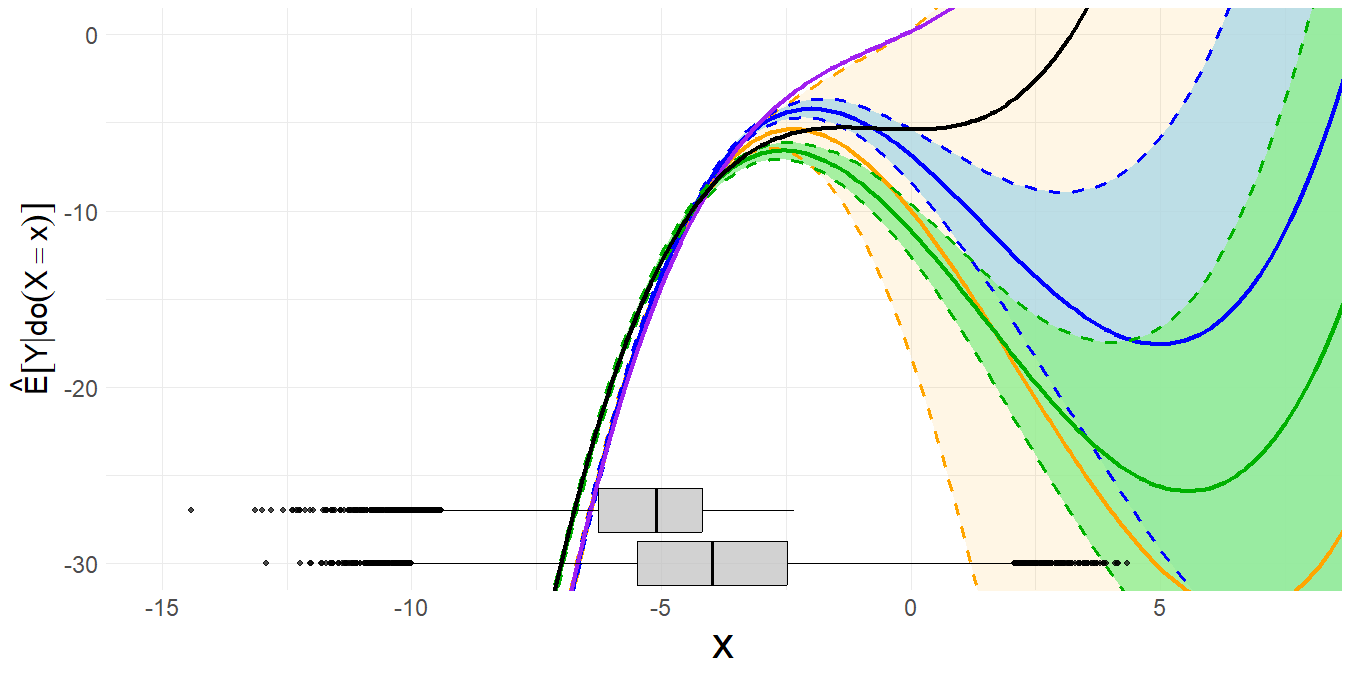}
      \caption*{$n=500$}
   \end{minipage}
   \hspace{.01\linewidth}
   \begin{minipage}[b]{.26\linewidth} 
      \includegraphics[width=1\linewidth]{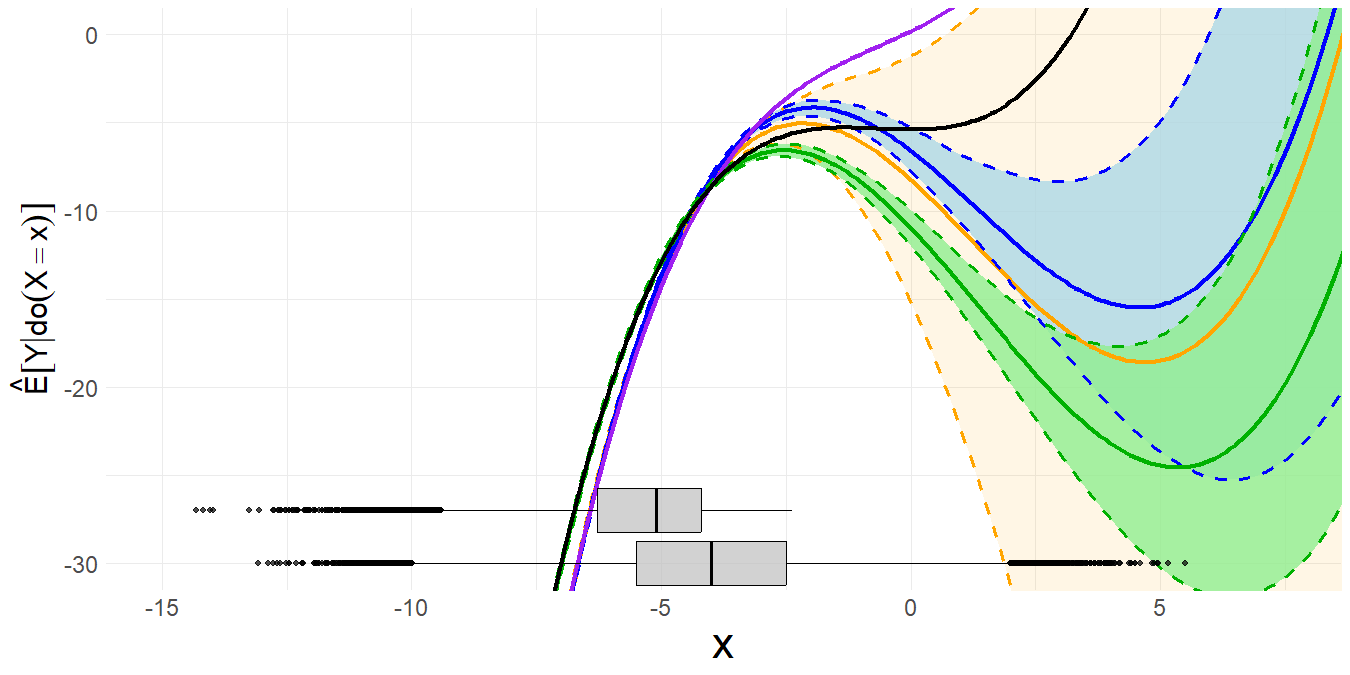}
      \caption*{$n=1000$}
   \end{minipage}
   \hspace{.01\linewidth}
   \begin{minipage}[b]{.26\linewidth} 
      \includegraphics[width=1\linewidth]{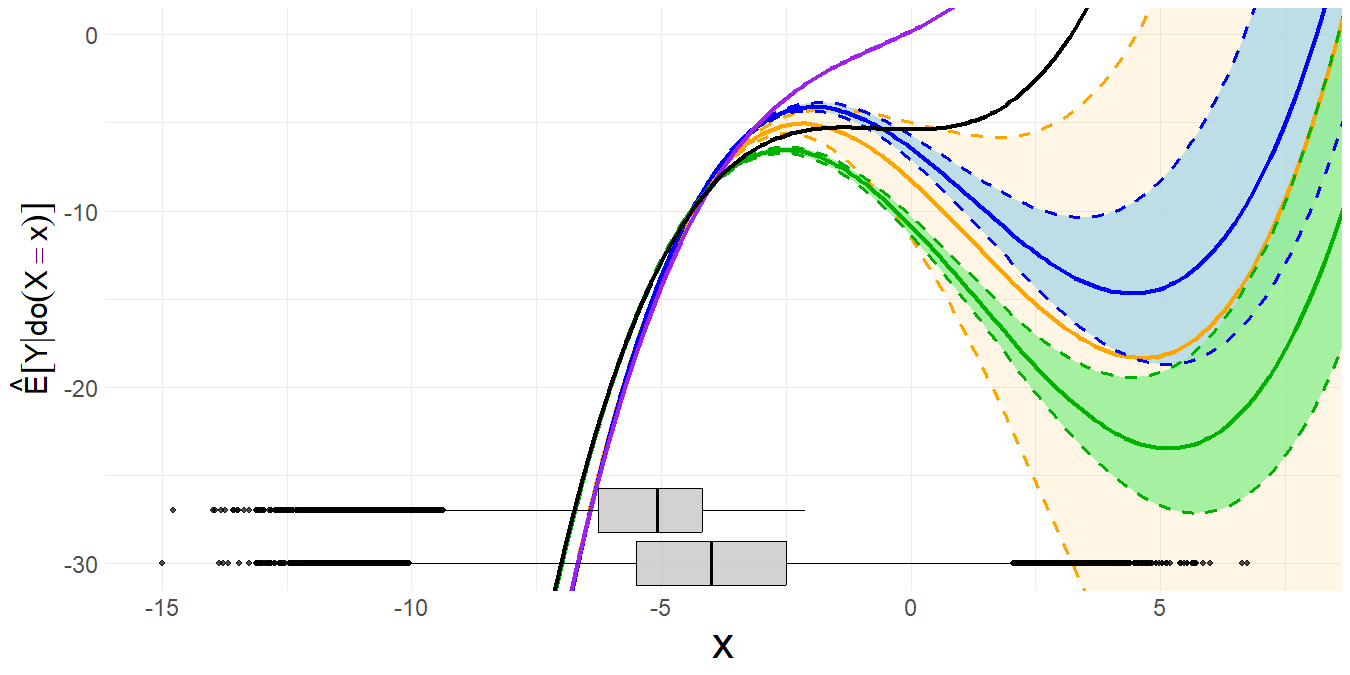}
      \caption*{$n=5000$}
   \end{minipage}
   \hspace{.01\linewidth}
   \begin{minipage}[b]{.07\linewidth} 
      \includegraphics[scale=0.11]{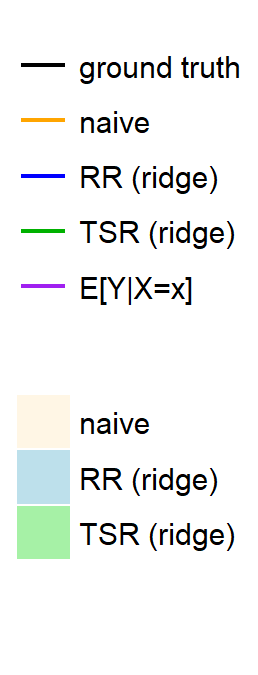}
      \caption*{}
   \end{minipage}
   \caption{\textbf{Example 1}: We show in black the ground truth value of $E[Y|do(X=x)]$, and colored the mean estimates and $95\%$-areas over all simulations for naive, RR, and TSR  for the DAG in Figure~\ref{fig:expvariousDAGs} (a) for $n\in\{500, 1000, 5000\}$ and with the coefficients fixed at their means (for $\mathcal{S}\cap\mathcal{D}=\emptyset$). The upper line shows the results for OLS and the lower for ridge.  As $E[Y\mid Y]\neq E[Y\mid do(X)]$ in these examples, we also show the true underlying $E[Y|X=x]$ in purple. We see that TSR recovers the ground truth $E[Y\mid do(X)]$, whereas RR recovers $E[Y\mid X]$. The boxplots on the x-axis show the distribution of $\Xv$ in $\mathcal{S}$ (upper) and $\mathcal{D}$ (lower).}
   \label{empty_ex1}
\end{figure}

\begin{figure}[h!]
   \begin{minipage}[b]{.26\linewidth}
      \includegraphics[width=1\linewidth]{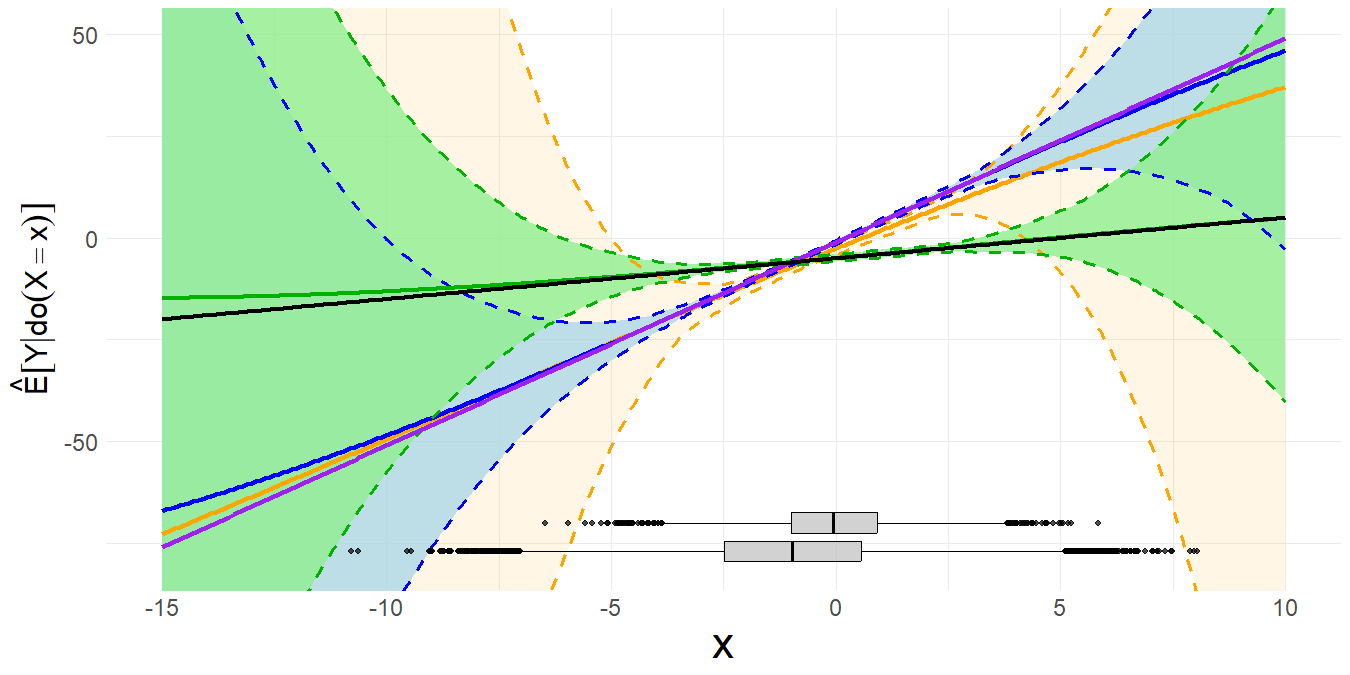}
      \caption*{$n=500$}
   \end{minipage}
   \hspace{.01\linewidth}
   \begin{minipage}[b]{.26\linewidth} 
      \includegraphics[width=1\linewidth]{Fig4b_11b.png}
      \caption*{$n=1000$}
   \end{minipage}
   \hspace{.01\linewidth}
   \begin{minipage}[b]{.26\linewidth} 
      \includegraphics[width=1\linewidth]{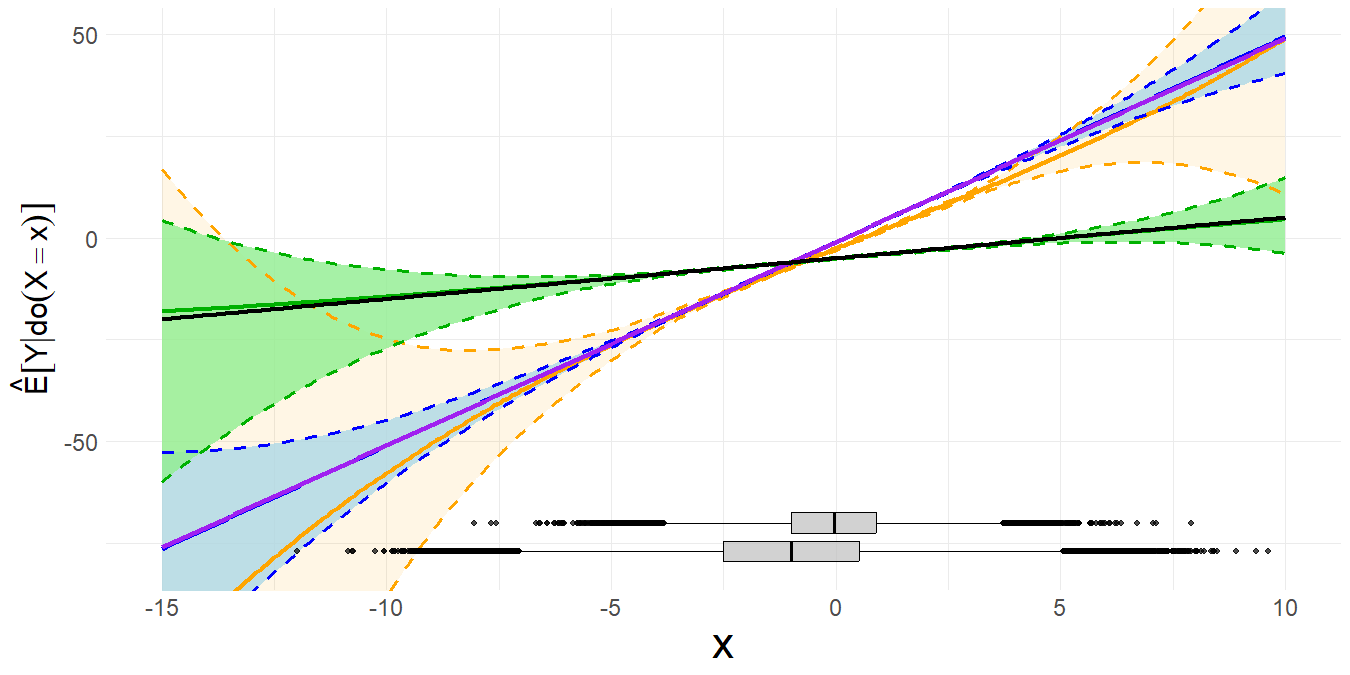}
      \caption*{$n=5000$}
   \end{minipage}
   \hspace{.01\linewidth}
   \begin{minipage}[b]{.07\linewidth} 
      \includegraphics[scale=0.11]{Fig4d_4h_10d_11d_12d_13d_14d_15d.png}
      \caption*{}
   \end{minipage}\\
   \begin{minipage}[b]{.26\linewidth}
      \includegraphics[width=1\linewidth]{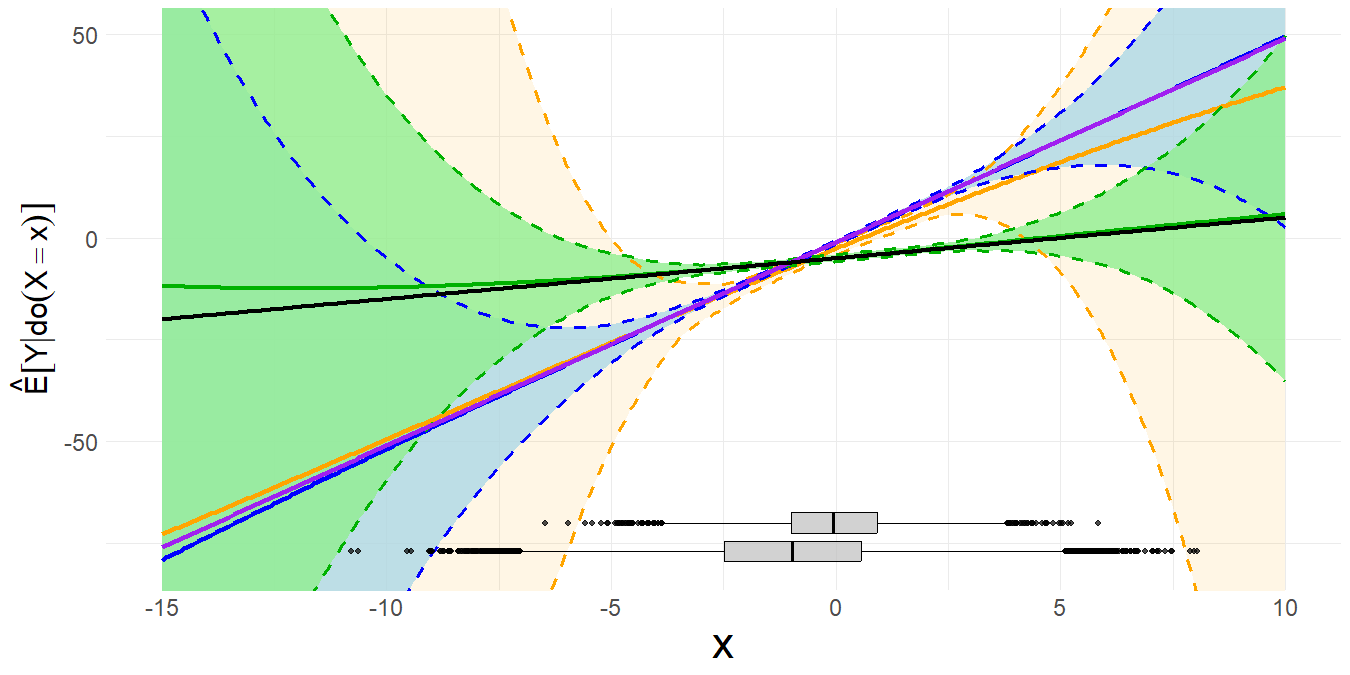}
      \caption*{$n=500$}
   \end{minipage}
   \hspace{.01\linewidth}
   \begin{minipage}[b]{.26\linewidth} 
      \includegraphics[width=1\linewidth]{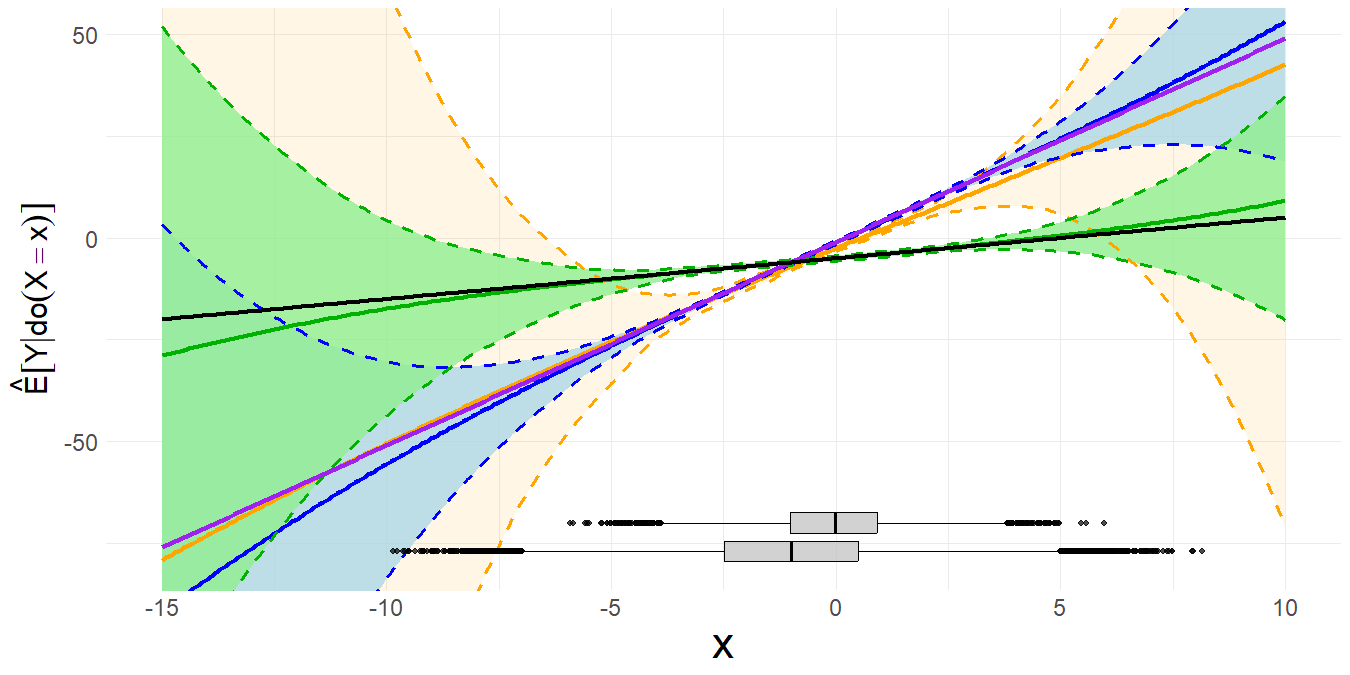}
      \caption*{$n=1000$}
   \end{minipage}
   \hspace{.01\linewidth}
   \begin{minipage}[b]{.26\linewidth} 
      \includegraphics[width=1\linewidth]{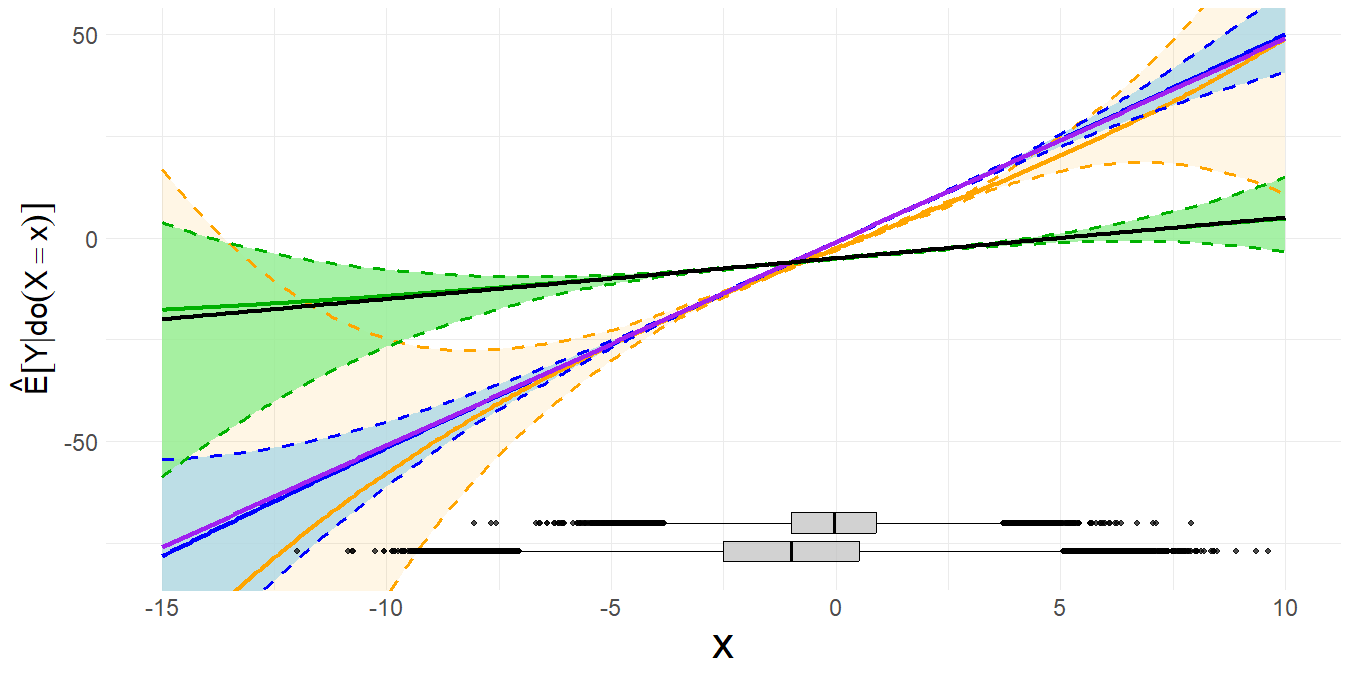}
      \caption*{$n=5000$}
   \end{minipage}
   \hspace{.01\linewidth}
   \begin{minipage}[b]{.07\linewidth} 
      \includegraphics[scale=0.11]{Fig10h_11h_12h_13h_14h_15h.png}
      \caption*{}
   \end{minipage}
 \caption{\textbf{Example 2}: We show in black the ground truth value of $E[Y|do(X=x)]$, and colored the mean estimates and $95\%$-areas over all simulations for naive, RR, and TSR  for the DAG in Figure~\ref{fig:expvariousDAGs} (a) for $n\in\{500, 1000, 5000\}$ and with the coefficients fixed at their means (for $\mathcal{S}\cap\mathcal{D}=\emptyset$). The upper line shows the results for OLS and the lower for ridge. As $E[Y\mid Y]\neq E[Y\mid do(X)]$ in these examples, we also show the true underlying $E[Y|X=x]$ in purple. We see that TSR recovers the ground truth $E[Y\mid do(X)]$, whereas RR recovers $E[Y\mid X]$. The boxplots on the x-axis show the distribution of $\Xv$ in $\mathcal{S}$ (upper) and $\mathcal{D}$ (lower).}   \label{empty_ex2}
\end{figure}

\begin{figure}[h!]
   \begin{minipage}[b]{.26\linewidth}
      \includegraphics[width=1\linewidth]{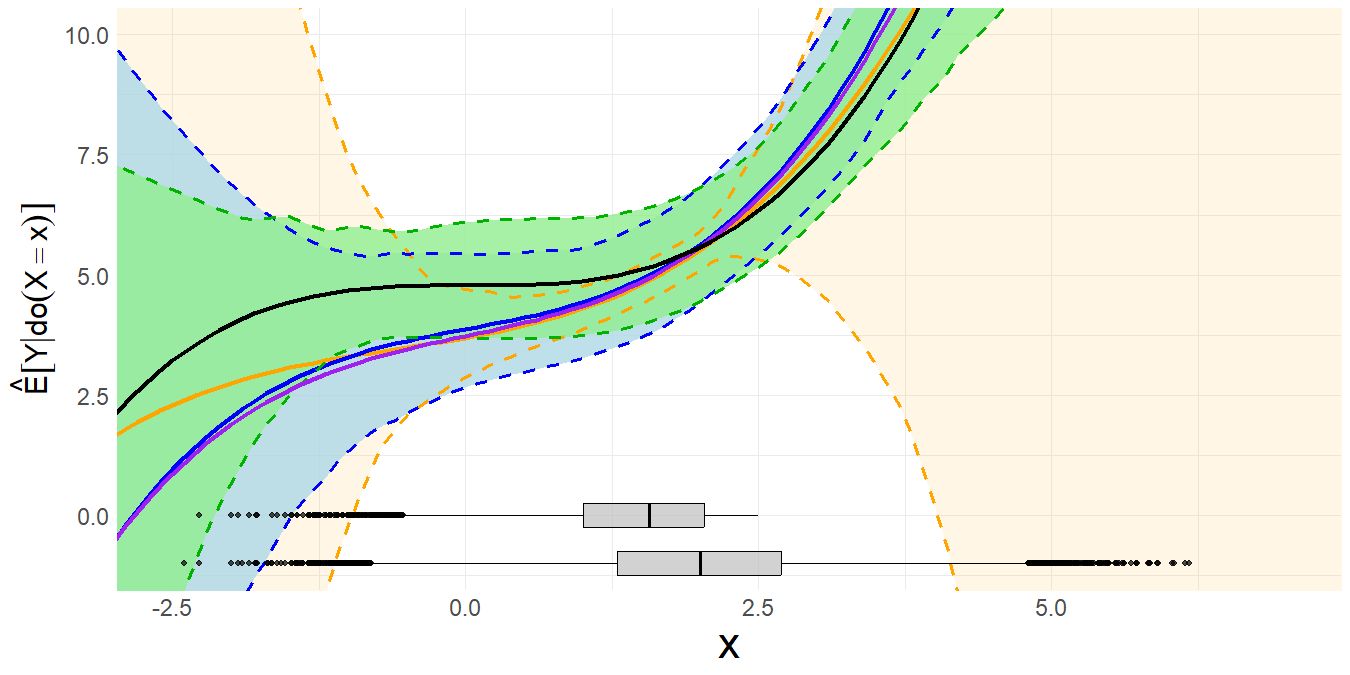}
      \caption*{$n=500$}
   \end{minipage}
   \hspace{.01\linewidth}
   \begin{minipage}[b]{.26\linewidth} 
      \includegraphics[width=1\linewidth]{Fig4c_12b.png}
      \caption*{$n=1000$}
   \end{minipage}
   \hspace{.01\linewidth}
   \begin{minipage}[b]{.26\linewidth} 
      \includegraphics[width=1\linewidth]{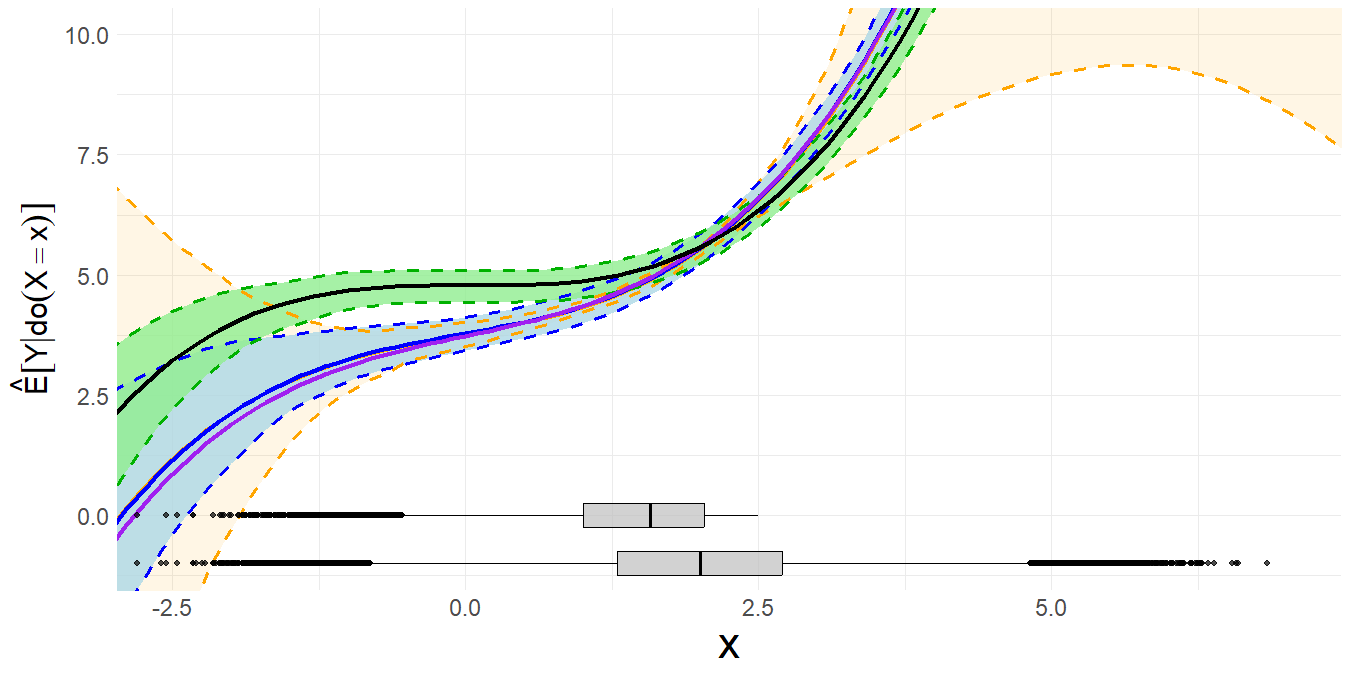}
      \caption*{$n=5000$}
   \end{minipage}
   \hspace{.01\linewidth}
   \begin{minipage}[b]{.07\linewidth} 
      \includegraphics[scale=0.11]{Fig4d_4h_10d_11d_12d_13d_14d_15d.png}
      \caption*{}
   \end{minipage}\\
   \begin{minipage}[b]{.26\linewidth}
      \includegraphics[width=1\linewidth]{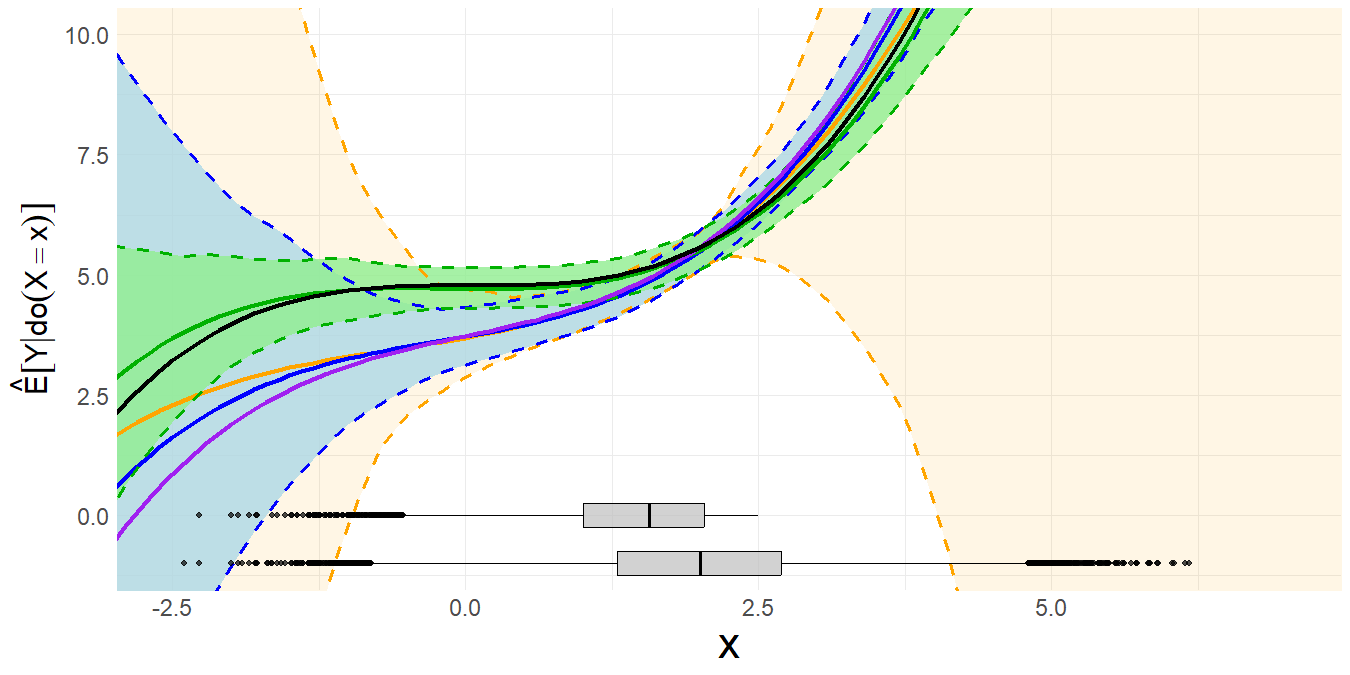}
      \caption*{$n=500$}
   \end{minipage}
   \hspace{.01\linewidth}
   \begin{minipage}[b]{.26\linewidth} 
      \includegraphics[width=1\linewidth]{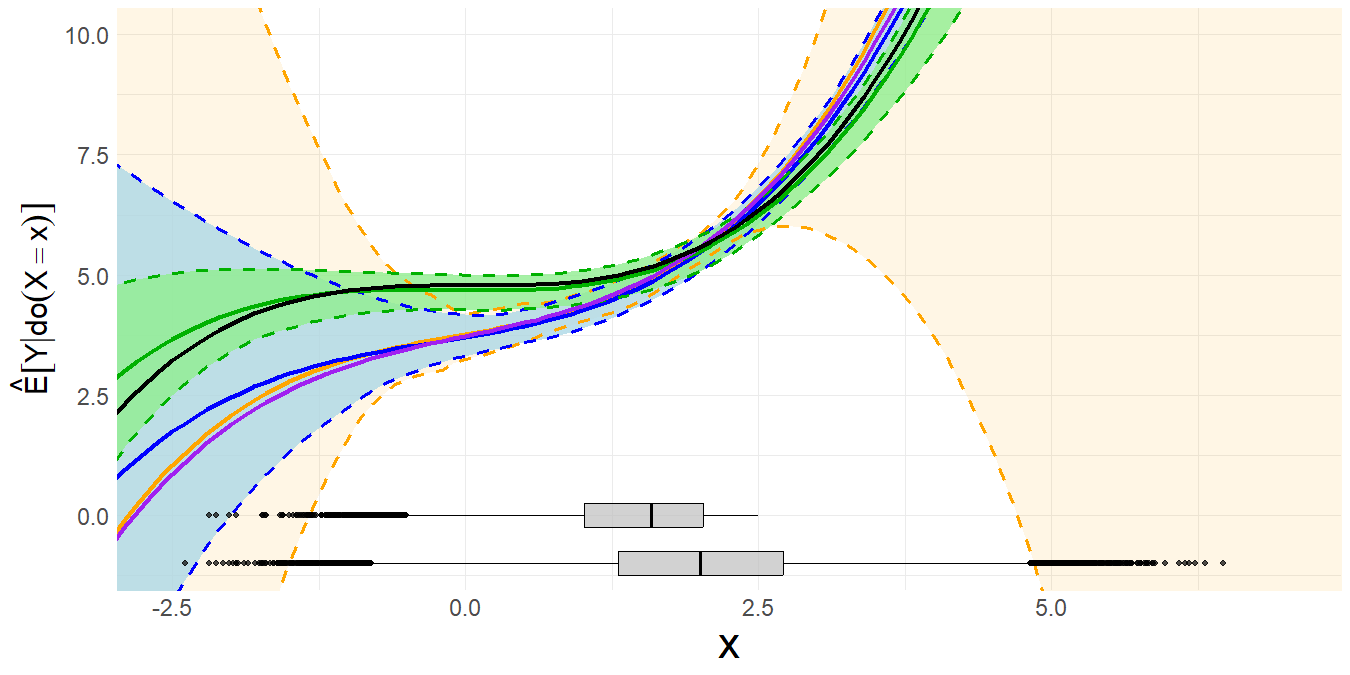}
      \caption*{$n=1000$}
   \end{minipage}
   \hspace{.01\linewidth}
   \begin{minipage}[b]{.26\linewidth} 
      \includegraphics[width=1\linewidth]{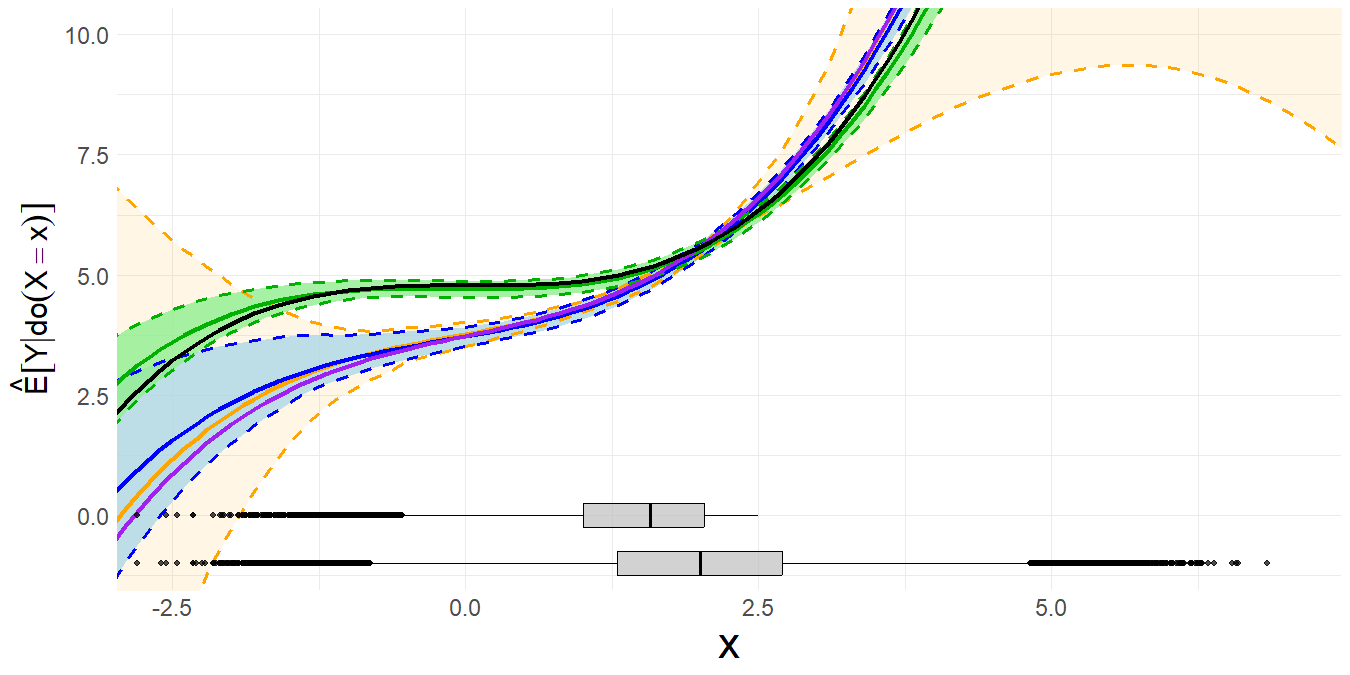}
      \caption*{$n=5000$}
   \end{minipage}
   \hspace{.01\linewidth}
   \begin{minipage}[b]{.07\linewidth} 
      \includegraphics[scale=0.11]{Fig10h_11h_12h_13h_14h_15h.png}
      \caption*{}
   \end{minipage}
 \caption{\textbf{Example 3}: We show in black the ground truth value of $E[Y|do(X=x)]$, and colored the mean estimates and $95\%$-areas over all simulations for naive, RR, and TSR  for the DAG in Figure~\ref{fig:expvariousDAGs} (b) for $n\in\{500, 1000, 5000\}$ and with the coefficients fixed at their means (for $\mathcal{S}\cap\mathcal{D}=\emptyset$). The upper line shows the results for OLS and the lower for ridge.  As $E[Y\mid Y]\neq E[Y\mid do(X)]$ in these examples, we also show the true underlying $E[Y|X=x]$ in purple. We see that TSR recovers the ground truth $E[Y\mid do(X)]$, whereas RR recovers $E[Y\mid X]$. The boxplots on the x-axis show the distribution of $\Xv$ in $\mathcal{S}$ (upper) and $\mathcal{D}$ (lower).}   \label{empty_ex3}
\end{figure}

\begin{figure}[h!]
   \begin{minipage}[b]{.26\linewidth}
      \includegraphics[width=1\linewidth]{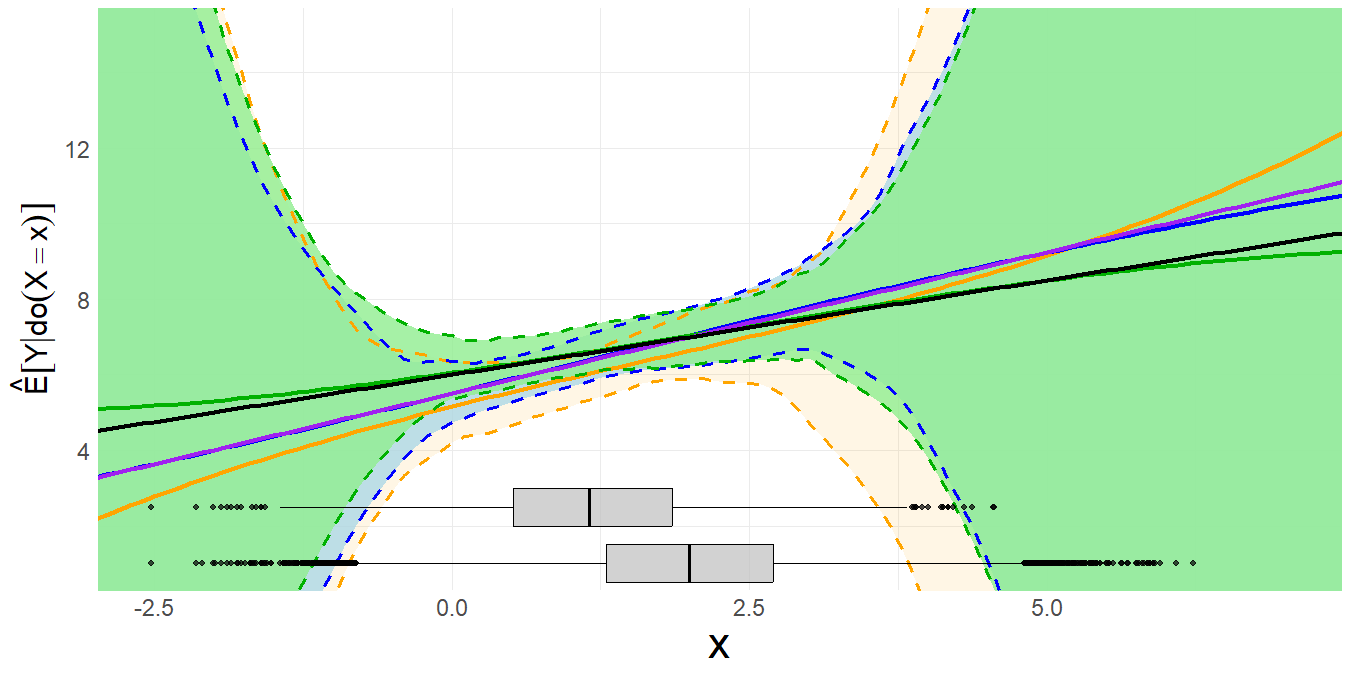}
      \caption*{$n=500$}
   \end{minipage}
   \hspace{.01\linewidth}
   \begin{minipage}[b]{.26\linewidth} 
      \includegraphics[width=1\linewidth]{Fig4e_13b.png}
      \caption*{$n=1000$}
   \end{minipage}
   \hspace{.01\linewidth}
   \begin{minipage}[b]{.26\linewidth} 
      \includegraphics[width=1\linewidth]{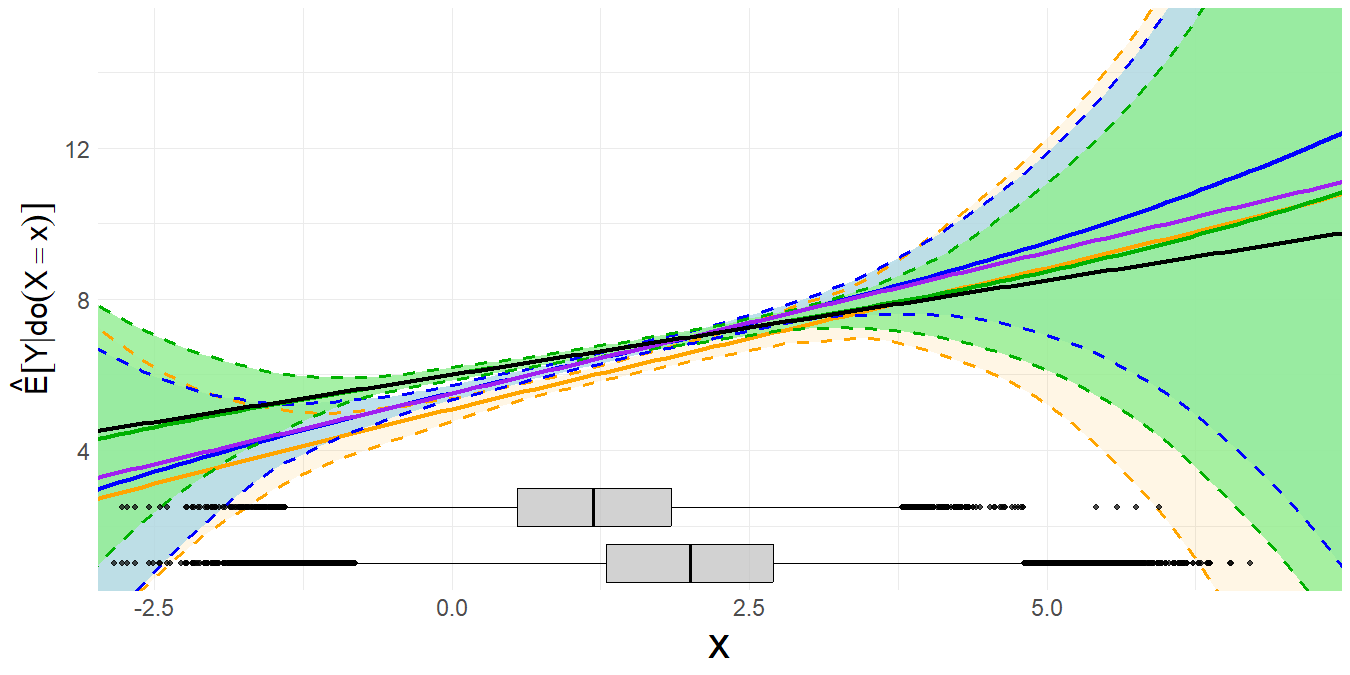}
      \caption*{$n=5000$}
   \end{minipage}
   \hspace{.01\linewidth}
   \begin{minipage}[b]{.07\linewidth} 
      \includegraphics[scale=0.11]{Fig4d_4h_10d_11d_12d_13d_14d_15d.png}
      \caption*{}
   \end{minipage}\\
   \begin{minipage}[b]{.26\linewidth}
      \includegraphics[width=1\linewidth]{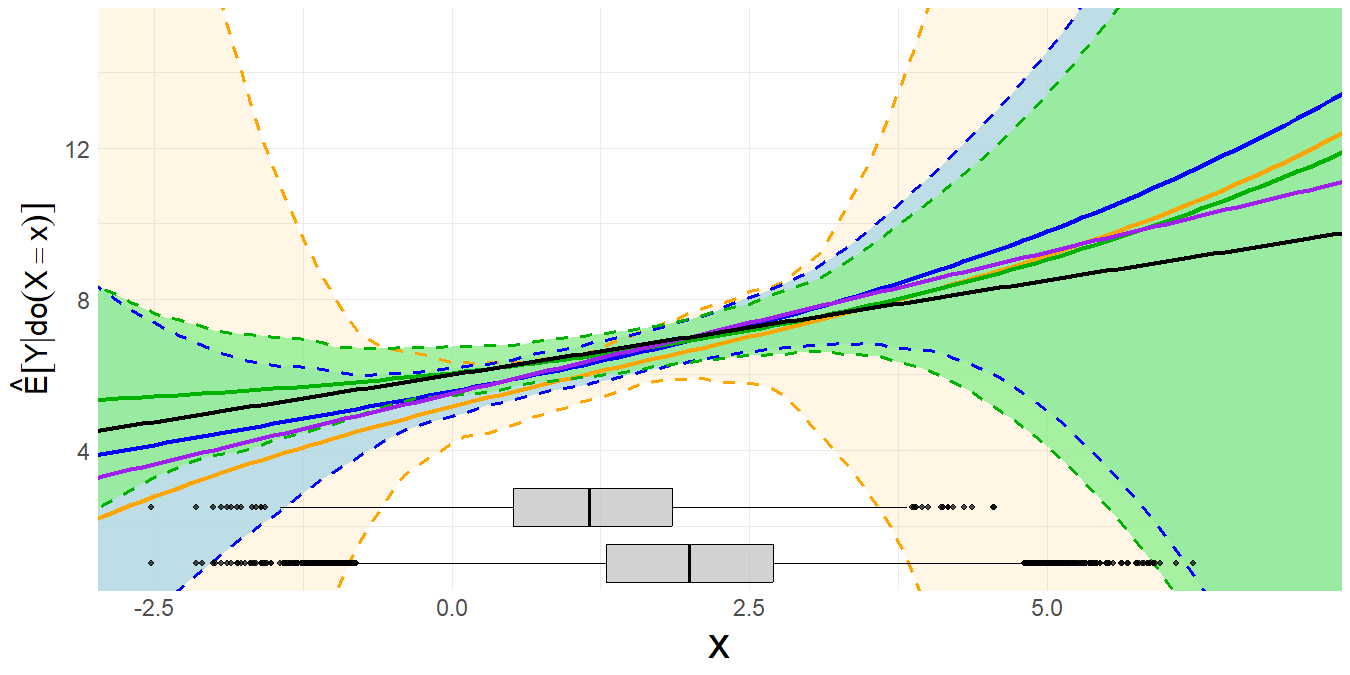}
      \caption*{$n=500$}
   \end{minipage}
   \hspace{.01\linewidth}
   \begin{minipage}[b]{.26\linewidth} 
      \includegraphics[width=1\linewidth]{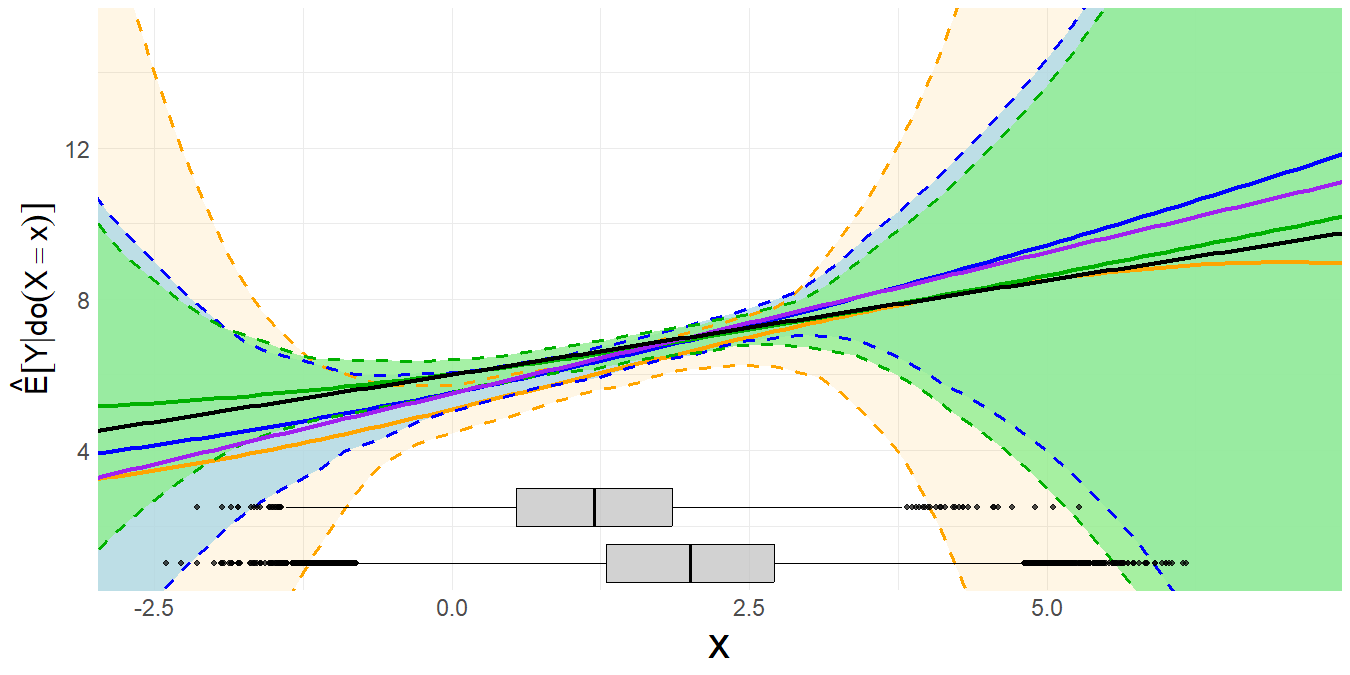}
      \caption*{$n=1000$}
   \end{minipage}
   \hspace{.01\linewidth}
   \begin{minipage}[b]{.26\linewidth} 
      \includegraphics[width=1\linewidth]{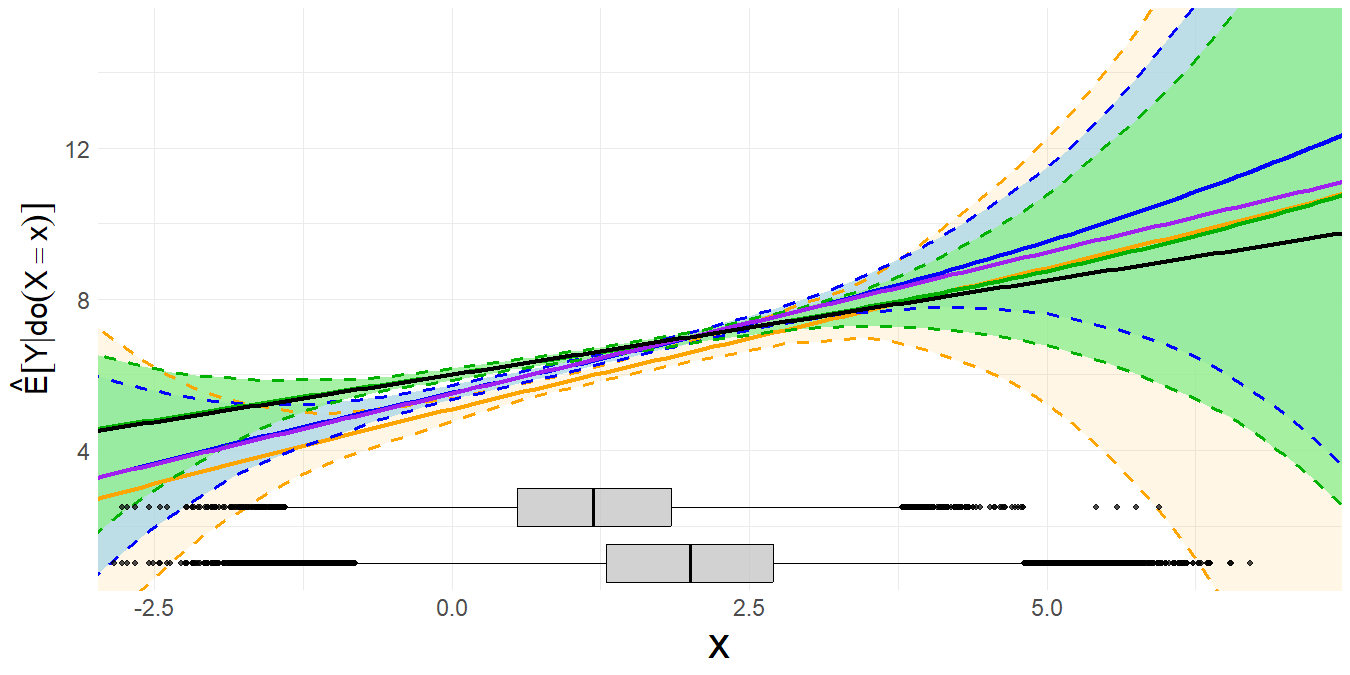}
      \caption*{$n=5000$}
   \end{minipage}
   \hspace{.01\linewidth}
   \begin{minipage}[b]{.07\linewidth} 
      \includegraphics[scale=0.11]{Fig10h_11h_12h_13h_14h_15h.png}
      \caption*{}
   \end{minipage}
 \caption{\textbf{Example 4}: We show in black the ground truth value of $E[Y|do(X=x)]$, and colored the mean estimates and $95\%$-areas over all simulations for naive, RR, and TSR  for the DAG in Figure~\ref{fig:expvariousDAGs} (b) for $n\in\{500, 1000, 5000\}$ and with the coefficients fixed at their means (for $\mathcal{S}\cap\mathcal{D}=\emptyset$). The upper line shows the results for OLS and the lower for ridge.  As $E[Y\mid Y]\neq E[Y\mid do(X)]$ in these examples, we also show the true underlying $E[Y|X=x]$ in purple. We see that TSR recovers the ground truth $E[Y\mid do(X)]$, whereas RR recovers $E[Y\mid X]$. The boxplots on the x-axis show the distribution of $\Xv$ in $\mathcal{S}$ (upper) and $\mathcal{D}$ (lower).}   \label{empty_ex4}
\end{figure}

\begin{figure}[h!]
   \begin{minipage}[b]{.26\linewidth}
      \includegraphics[width=1\linewidth]{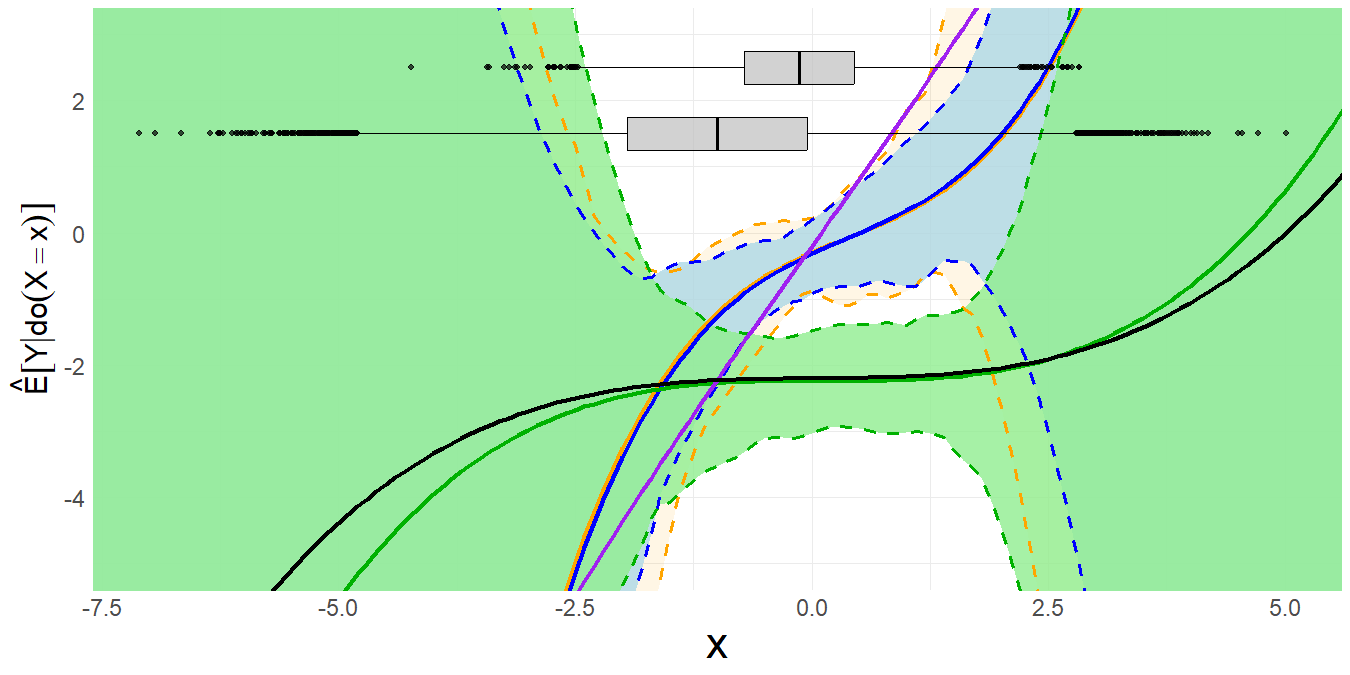}
      \caption*{$n=500$}
   \end{minipage}
   \hspace{.01\linewidth}
   \begin{minipage}[b]{.26\linewidth} 
      \includegraphics[width=1\linewidth]{Fig4f_14b.png}
      \caption*{$n=1000$}
   \end{minipage}
   \hspace{.01\linewidth}
   \begin{minipage}[b]{.26\linewidth} 
      \includegraphics[width=1\linewidth]{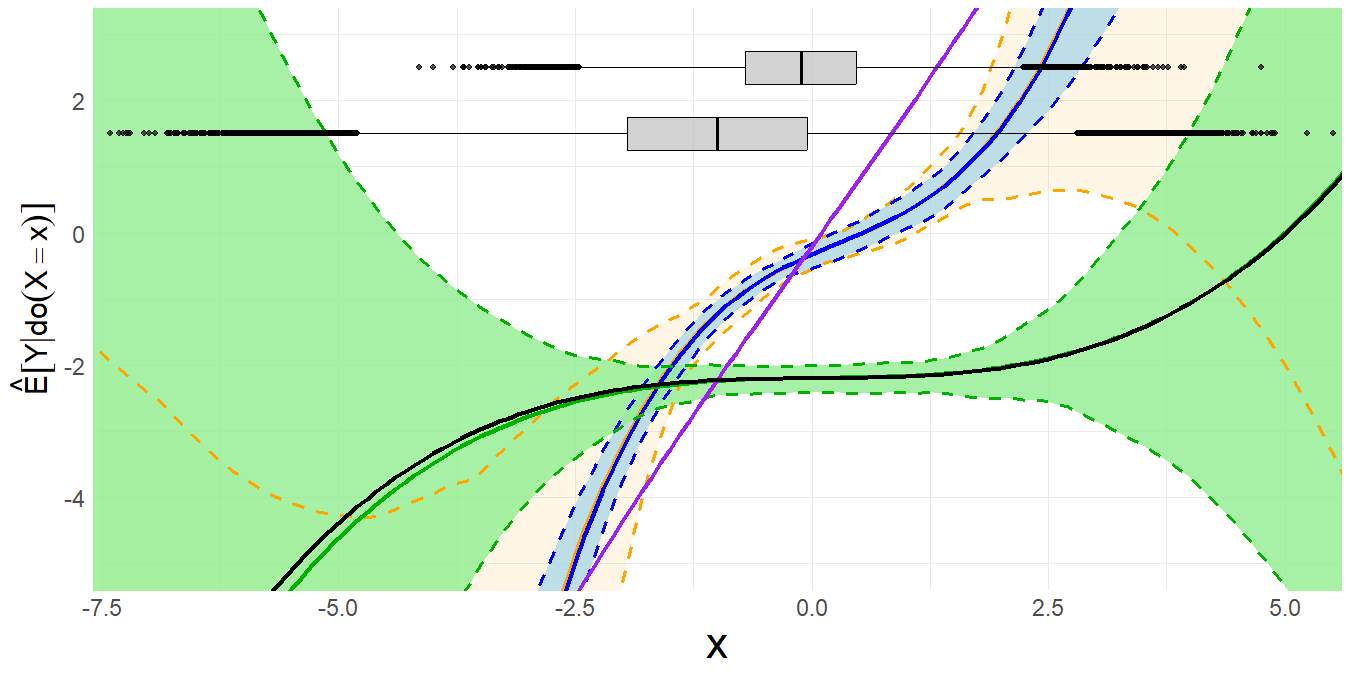}
      \caption*{$n=5000$}
   \end{minipage}
   \hspace{.01\linewidth}
   \begin{minipage}[b]{.07\linewidth} 
      \includegraphics[scale=0.11]{Fig4d_4h_10d_11d_12d_13d_14d_15d.png}
      \caption*{}
   \end{minipage}\\
   \begin{minipage}[b]{.26\linewidth}
      \includegraphics[width=1\linewidth]{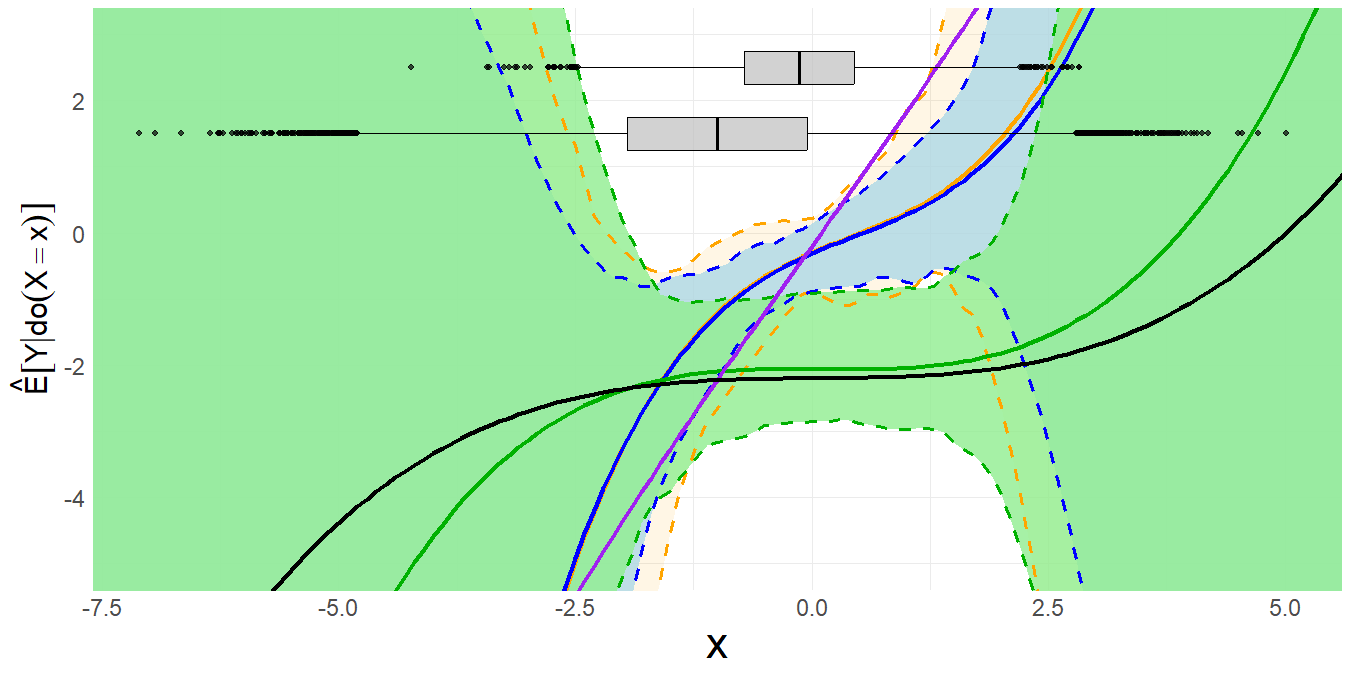}
      \caption*{$n=500$}
   \end{minipage}
   \hspace{.01\linewidth}
   \begin{minipage}[b]{.26\linewidth} 
      \includegraphics[width=1\linewidth]{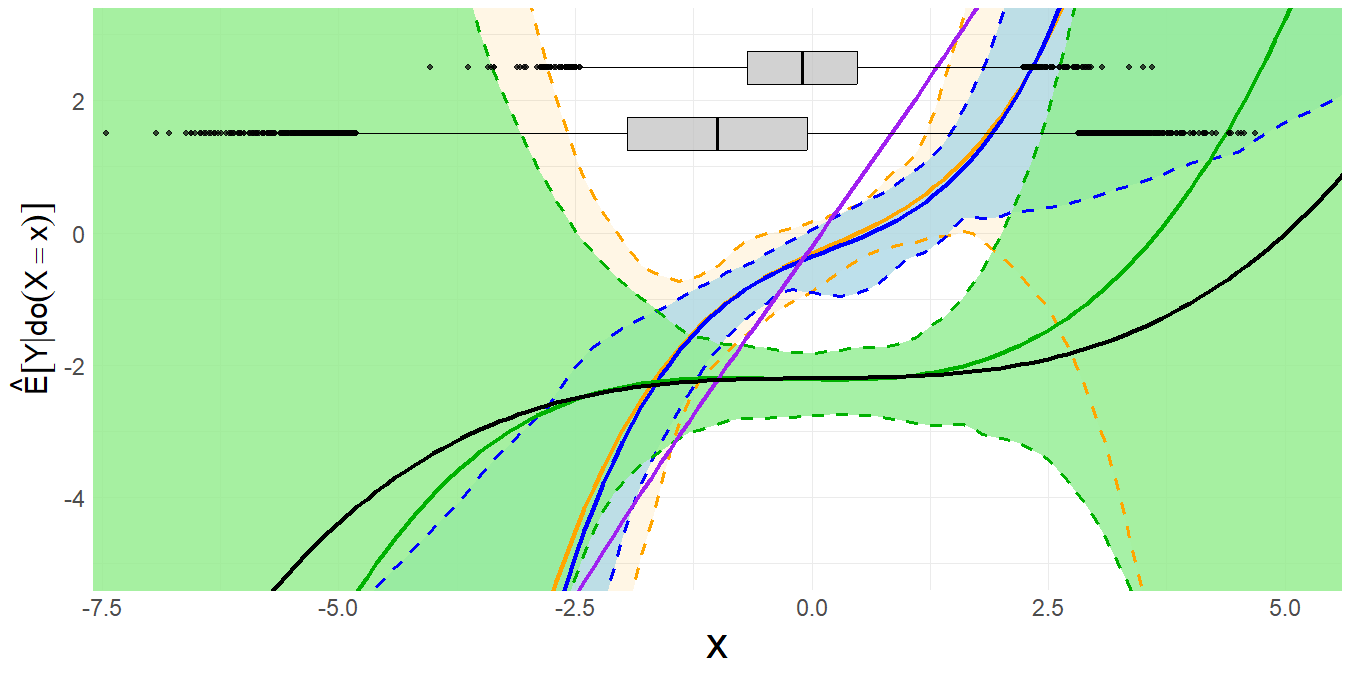}
      \caption*{$n=1000$}
   \end{minipage}
   \hspace{.01\linewidth}
   \begin{minipage}[b]{.26\linewidth} 
      \includegraphics[width=1\linewidth]{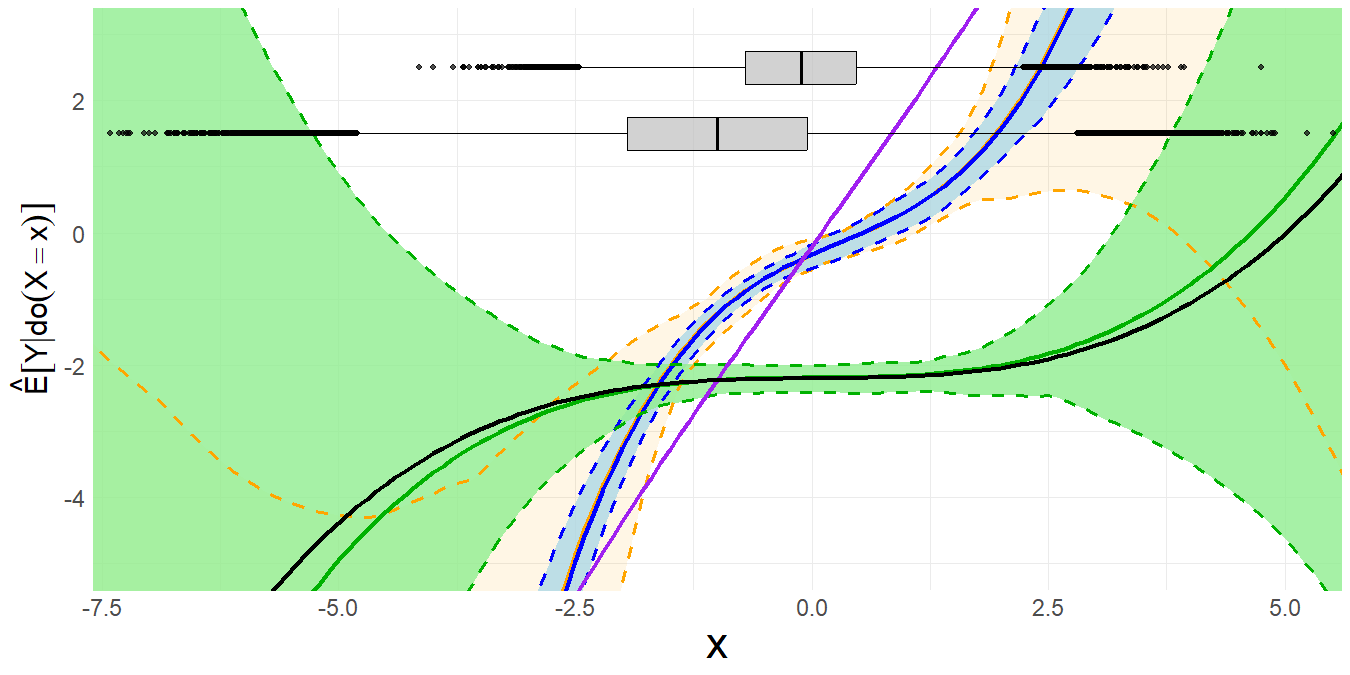}
      \caption*{$n=5000$}
   \end{minipage}
   \hspace{.01\linewidth}
   \begin{minipage}[b]{.07\linewidth} 
      \includegraphics[scale=0.11]{Fig10h_11h_12h_13h_14h_15h.png}
      \caption*{}
   \end{minipage}
 \caption{\textbf{Example 5}: We show in black the ground truth value of $E[Y|do(X=x)]$, and colored the mean estimates and $95\%$-areas over all simulations for naive, RR, and TSR  for the DAG in Figure~\ref{fig:expvariousDAGs} (c) for $n\in\{500, 1000, 5000\}$ and with the coefficients fixed at their means (for $\mathcal{S}\cap\mathcal{D}=\emptyset$). The upper line shows the results for OLS and the lower for ridge.  As $E[Y\mid Y]\neq E[Y\mid do(X)]$ in these examples, we also show the true underlying $E[Y|X=x]$ in purple. We see that TSR recovers the ground truth $E[Y\mid do(X)]$, whereas RR recovers $E[Y\mid X]$. The boxplots on the x-axis show the distribution of $\Xv$ in $\mathcal{S}$ (upper) and $\mathcal{D}$ (lower).}
\label{empty_ex5}
\end{figure}

\begin{figure}[h!]
   \begin{minipage}[b]{.26\linewidth}
      \includegraphics[width=1\linewidth]{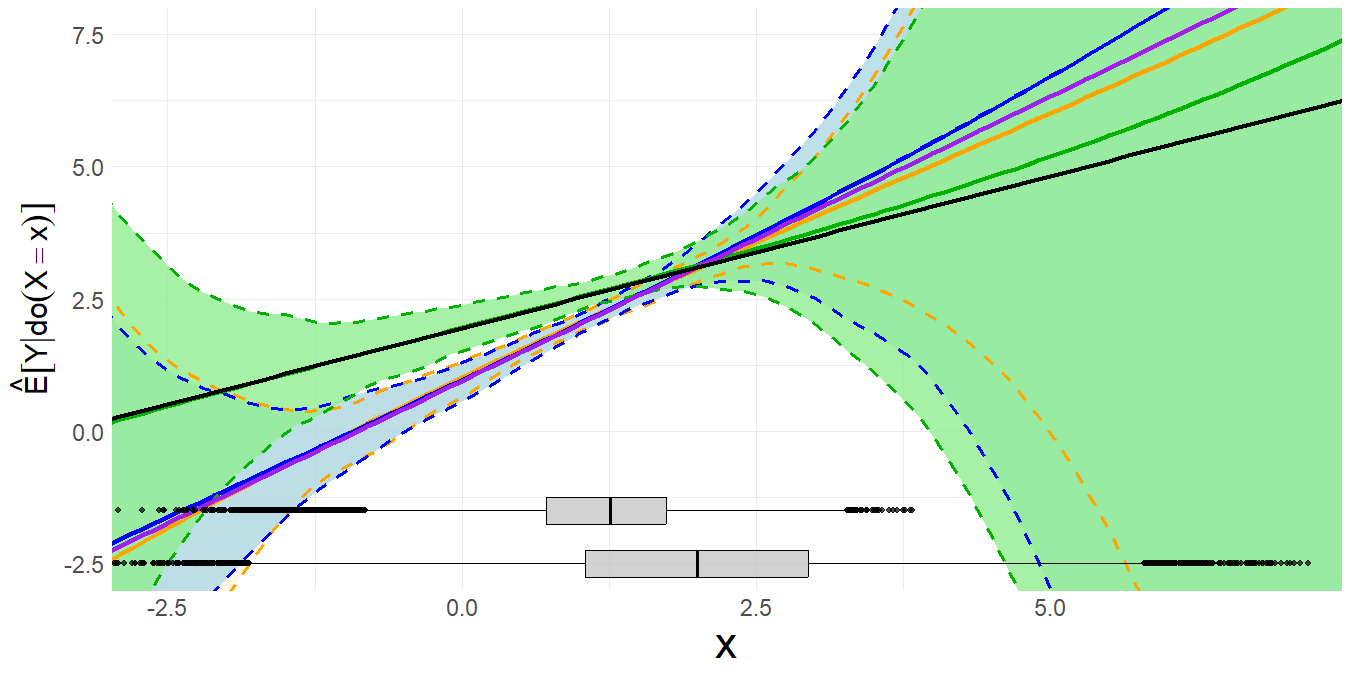}
      \caption*{$n=500$}
   \end{minipage}
   \hspace{.01\linewidth}
   \begin{minipage}[b]{.26\linewidth} 
      \includegraphics[width=1\linewidth]{Fig4g_15b.png}
      \caption*{$n=1000$}
   \end{minipage}
   \hspace{.01\linewidth}
   \begin{minipage}[b]{.26\linewidth} 
      \includegraphics[width=1\linewidth]{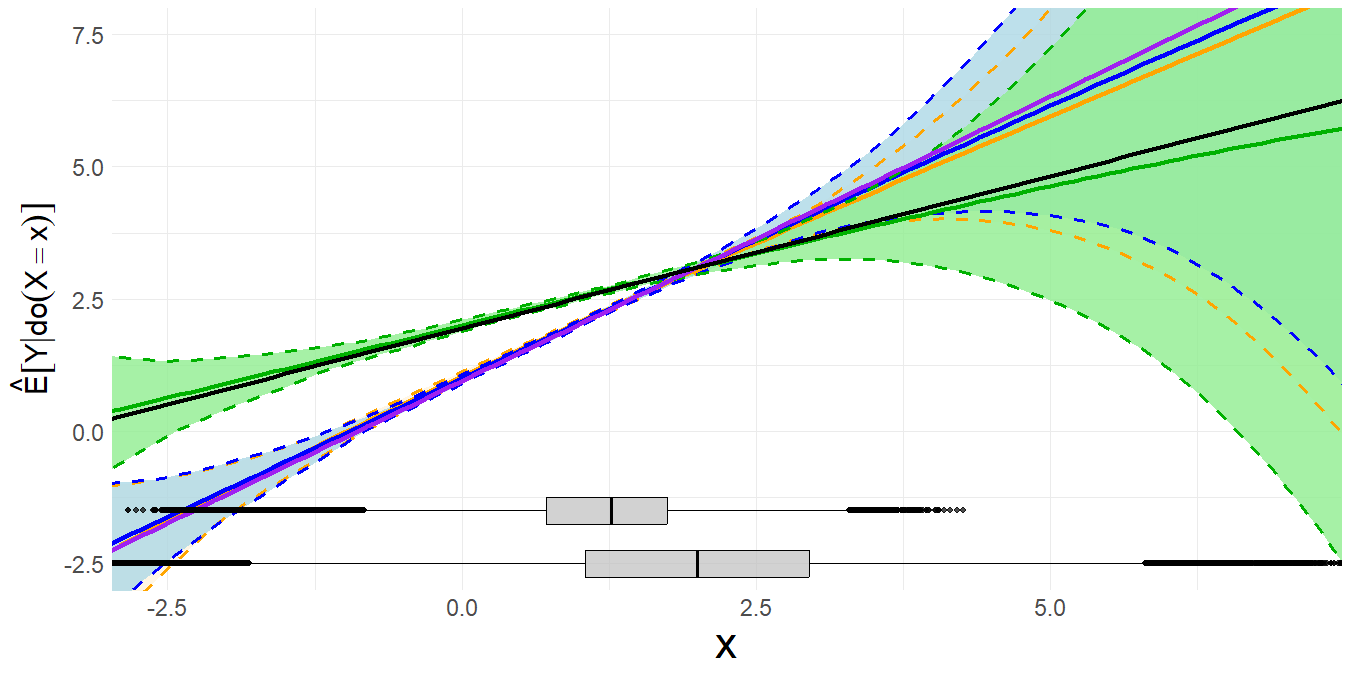}
      \caption*{$n=5000$}
   \end{minipage}
   \hspace{.01\linewidth}
   \begin{minipage}[b]{.07\linewidth} 
      \includegraphics[scale=0.11]{Fig4d_4h_10d_11d_12d_13d_14d_15d.png}
      \caption*{}
   \end{minipage}\\
   \begin{minipage}[b]{.26\linewidth}
      \includegraphics[width=1\linewidth]{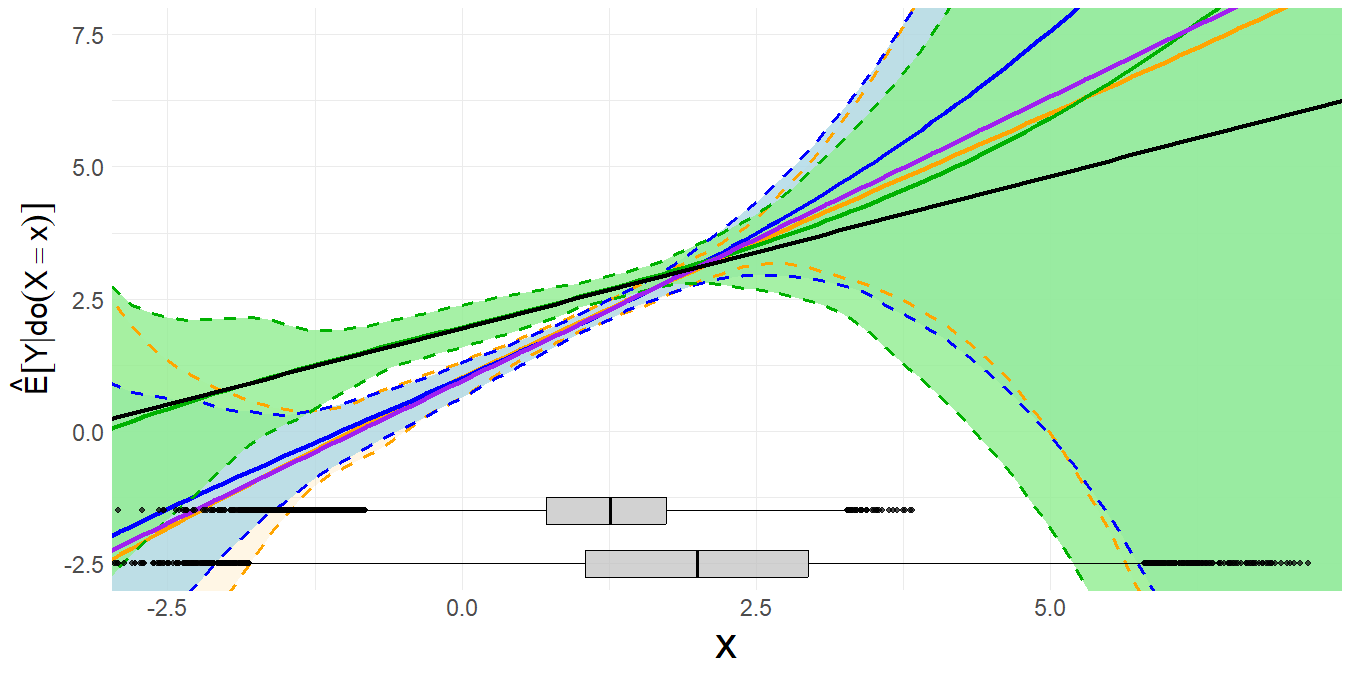}
      \caption*{$n=500$}
   \end{minipage}
   \hspace{.01\linewidth}
   \begin{minipage}[b]{.26\linewidth} 
      \includegraphics[width=1\linewidth]{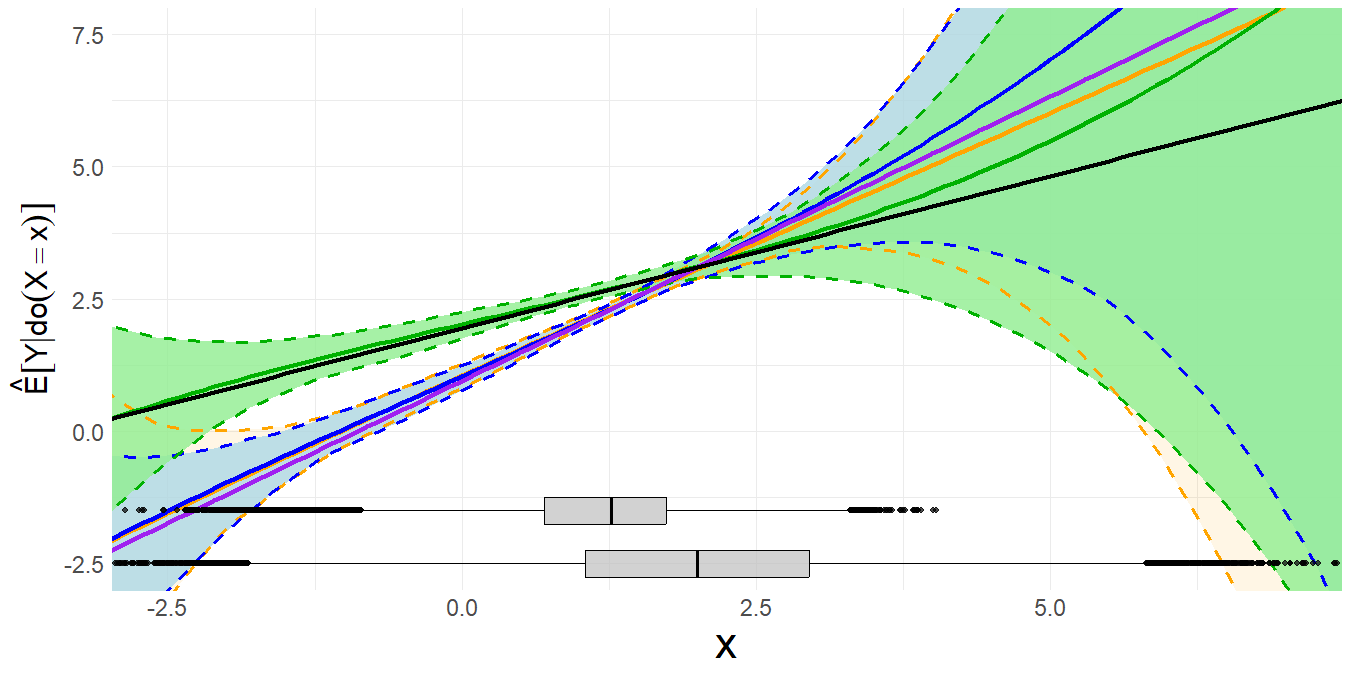}
      \caption*{$n=1000$}
   \end{minipage}
   \hspace{.01\linewidth}
   \begin{minipage}[b]{.26\linewidth} 
      \includegraphics[width=1\linewidth]{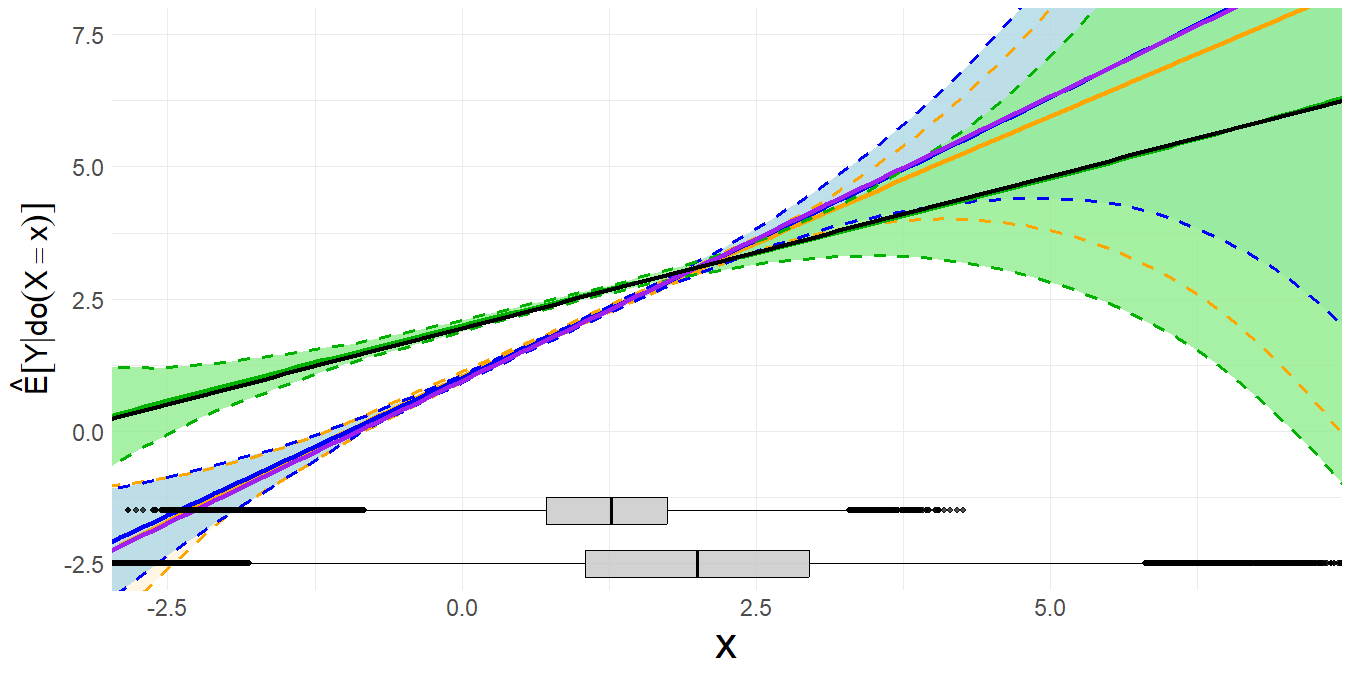}
      \caption*{$n=5000$}
   \end{minipage}
   \hspace{.01\linewidth}
   \begin{minipage}[b]{.07\linewidth} 
       \includegraphics[scale=0.11]{Fig10h_11h_12h_13h_14h_15h.png}
      \caption*{}
   \end{minipage}
 \caption{\textbf{Example 6}: We show in black the ground truth value of $E[Y|do(X=x)]$, and colored the mean estimates and $95\%$-areas over all simulations for naive, RR, and TSR  for the DAG in Figure~\ref{fig:expvariousDAGs} (c) for $n\in\{500, 1000, 5000\}$ and with the coefficients fixed at their means (for $\mathcal{S}\cap\mathcal{D}=\emptyset$). The upper line shows the results for OLS and the lower for ridge. As $E[Y\mid Y]\neq E[Y\mid do(X)]$ in these examples, we also show the true underlying $E[Y|X=x]$ in purple. We see that TSR recovers the ground truth $E[Y\mid do(X)]$, whereas RR recovers $E[Y\mid X]$. The boxplots on the x-axis show the distribution of $\Xv$ in $\mathcal{S}$ (upper) and $\mathcal{D}$ (lower).}   \label{empty_ex6}
\end{figure}

In this section, we will present further details concerning the six examples mentioned in \Cref{sec:expvariance}. First, we state the data generating processes. Then, for each example, we show six plots \mbox{(Figures~\ref{empty_ex1}--\ref{empty_ex6})}, which show the 95\%-areas and means of TSR and RR for varying $n\in\{500,1000,5000\}$ and the effect of adding a ridge penalty. Finally, we will present the mean and standard deviation of the MSE over all considered settings, evaluated on $\mathcal{S}$ as well as on $\mathcal{D}$ in Tables~\ref{tab:ex1data}--\ref{tab:ex6data}. The data generating processes are given in the following.

\textbf{Example 1:} We generate the data according to the SCM
\begin{align*}
X&:=2Z+\varepsilon_X\\
S&:=\mathbf{1}_{\{X+Z<-5\}}\\
Y&:=\mu_0+a_1X+a_2X^2+a_3X^3+b_1Z+b_2Z^2+b_3Z^3+\varepsilon_Y
\end{align*}
with $\varepsilon_X,\varepsilon_Y\sim\mathcal{N}(0,1)$, $Z\sim\mathcal{N}(-2,1)$, $a_2\sim \text{Unif}(-0.3,0.7)$, $a_3,b_3\sim \text{Unif}(-0.4,0.6)$, $b_1\sim \text{Unif}(1.5,2.5)$ and $\mu_0,a_1,b_2\sim \text{Unif}(-0.5,0.5)$. The ground truth expectations are as follows:
\begin{align*}
    E[Y\mid X=x]%
    &=\mu_0+a_1x+a_2x^2+a_3x^3+b_1(-\frac{2}{5}+\frac{2}{5}x)+b_2(5-\frac{8}{5}(x+4))+b_3(10+6x)\\
    E[Y\mid do(X:=x)]%
    &=\mu_0+a_1x+a_2x^2+a_3x^3+b_1(-2)+b_25+b_3(-14)
\end{align*}

\textbf{Example 2:} We generate the data according to the SCM
\begin{align*}
X&:=Z+\varepsilon_X\\
S&\sim \mathcal{B}ern\biggl(\frac{1}{(1+exp(-X))(1+exp(Z))},n\biggr)\\
Y&:=\mu_0+a_1X+a_2X^2+a_3X^3+b_1Z+b_2Z^2+b_3Z^3+\varepsilon_Y
\end{align*}
with $\varepsilon_X,\varepsilon_Y\sim\mathcal{N}(0,1)$, $
Z\sim\mathcal{N}(-1,4)$, $a_1\sim \text{Unif}(0.5,1.5)$, $
b_1\sim \text{Unif}(4.5,5.5)$ and $\mu_0,a_2,a_3,b_2,b_3\sim \text{Unif}(-0.5,0.5)$. The ground truth expectations are as follows:
\begin{align*}
    E[Y\mid X=x]%
    &=\mu_0+a_1x+a_2x^2+a_3x^3+b_1(-\frac{1}{5}+\frac{4}{5}x)+b_2(5-\frac{8}{5}(x+1))\\
    &\;\;\;\;\;+b_3(-13+\frac{60}{5}(x+1))\\
    E[Y\mid do(X:=x)]&=\mu_0+a_1x+a_2x^2+a_3x^3+b_1(-1)+b_2(5)+b_3(-13)
\end{align*}

\textbf{Example 3:} We generate the data according to the SCM
\begin{align*}
X&:=Z_B^++\varepsilon_X\\
S&:=\mathbf{1}_{\{Z_A^+>0,X<2.5\}}\\
Y&:=\mu_0+a_1X+a_2X^2+b_1Z_A^++b_2(Z_A^+)^2+b_3(Z_A^+)^3+c_1Z_B^++c_2(Z_B^+)^2+c_3(Z_B^+)^3+\varepsilon_Y
\end{align*}
with $\varepsilon_X\sim\mathcal{N}(0,1)$, $\varepsilon_Y\sim\mathcal{N}(0,0.3^2)$, $Z_B^+\sim \mathcal{N}(2,0.3^2)$, $Z_A^+\sim\mathcal{N}(-0.3,1)$, $a_3,b_2\sim \text{Unif}(-0.4,0.6)$, $c_1\sim \text{Unif}(-0.3,0.7)$, $c_3\sim \text{Unif}(0,1)$ and $\mu_0,a_1,a_2,b_1,b_3,c_2\sim \text{Unif}(-0.5,0.5)$. The ground truth expectations are as follows: 
\begin{align*}
    E[Y\mid X=x] &=\mu_0+a_1x+a_2x^2+a_3x^3+b_1(-0.3)+b_2(1.09)+b_3(-0.927)\\
    &\;\;\;\;\;+c_1(2+\frac{0.3^2}{0.3^2+1}(x-2))+c_2(4.09+\frac{8.54-8.18}{1.09}(x-2))\\
    &\;\;\;\;\;+c_3(8.54+\frac{18.1843-17.08}{1.09}(x-2)\\
    E[Y\mid do(X:=x)]&=\mu_0+a_1x+a_2x^2+a_3x^3+b_1(-0.3)+b_2(1.09)+b_3(-0.927)\\
    &\;\;\;\;\;+c_1(2)+c_2(4.09)+c_3(8.54)
\end{align*}

\textbf{Example 4:} We generate the data according to the SCM
\begin{align*}
X&:=Z_B^++\varepsilon_X\\
S&\sim\mathcal{B}ern\biggl(\frac{1}{(1+exp(X))(1+exp(Z_A^+))}\biggr)\\
Y&:=\mu_0+a_1X+a_2X^2+a_3x^3+b_1Z_A^++b_2(Z_A^+)^2+b_3(Z_A^+)^3+c_1Z_B^++c_2(Z_B^+)^2+c_3(Z_B^+)^3+\varepsilon_Y
\end{align*}
with $\varepsilon_X,\varepsilon_Y\sim\mathcal{N}(0,1)$, $
Z_B^+\sim\mathcal{N}(2,0.3^2)$, $Z_A^+\sim\mathcal{N}(0,1)$, $a_1\sim \text{Unif}(0,1)$, 
$b_1\sim \text{Unif}(0.5,1.5)$,
$c_1\sim \text{Unif}(2.5,3.5)$ and $
\mu_0,a_2,a_3,b_2,b_3,c_2,c_3\sim \text{Unif}(-0.5,0.5)$. The ground truth expectations are as follows:
\begin{align*}
    E[Y\mid X=x]&=\mu_0+a_1x+a_2x^2+a_3x^3+b_1(0)+b_2(1)+b_3(0)+c_1(2+\frac{0.3^2}{0.3^2+1}(x-2))\\
    &\;+c_2(4.09+\frac{8.54-8.18}{0.3^2+1}(x-2))+c_3(8.54+\frac{18.1843-17.08}{0.3^2+1}(x-2))\\
    E[Y\mid do(X:=x)]&=\mu_0+a_1x+a_2x^2+a_3x^3+b_1(0)+b_2(1)+b_3(0)+c_1(2)+c_2(4.09)+c_3(8.54)
\end{align*}

\textbf{Example 5:} We generate data according to the SCM
\begin{align*}
X&:=Z^++\varepsilon_X\\
Z^-&:=-2X+\varepsilon_{Z^-}\\
S&\sim \mathcal{B}ern\biggl(\frac{1}{(1+exp(X))(1+exp(Z^-))}\biggr)\\
Y&:=\mu_0+a_1X+a_2X^2+a_3x^3+b_1Z^-+b_2(Z^-)^2+b_3(Z^-)^3\\
&\;\;\;\;\;+c_1Z^++c_2(Z^+)^2+c_3(Z^+)^3+\varepsilon_Y
\end{align*}
with $\varepsilon_X,\varepsilon_{Z^-},\varepsilon_Y\sim\mathcal{N}(0,1)$, $
Z^+\sim\mathcal{N}(-1,1)$, $\mu_0\sim \text{Unif}(-0.7,0.3)$, $a_3\sim \text{Unif}(-0.499,0.501)$, $b_3\sim \text{Unif}(-0.502,0.498)$, $
c_2,c_3\sim \text{Unif}(1.5,2.5)$ and $\mu_0,a_1,a_2,b_1,b_2,c_1\sim \text{Unif}(-0.5,0.5)$. The ground truth expectation is as follows:
\begin{align*}
E[Y\mid X=x]&=\mu_0+a_1x+a_2x^2+a_3x^3+b_1(-2x)+b_2(4x^2+1)+b_3(-8x^3-6x)\\
&\;\;\;\;\;+c_1(-0.5+0.5x)+c_2(1-x)+c_3(-1+3x)\\
E[Y\mid do(X:=x)]&=\mu_0+a_1x+a_2x^2+a_3x^3+b_1(-2x)+b_2(4x^2+1)+b_3(-8x^3-6x)\\
&\;\;\;\;\;+c_1(-1)+c_2(2)+c_3(-4)\\
\end{align*}

\textbf{Example 6:} We generate the data according to the SCM
\begin{align*}
X&:=Z^++\varepsilon_X\\
Z^-&:=X+\varepsilon_{Z^-}\\
S&:=\mathbf{1}_{\{(Z^-X)<4,(Z^-X)^2+Z^->0\}}\\
Y&:=\mu_0+a_1X+a_2X^2+a_3X+b_1Z^-+b_2(Z^-)^2+b_3(Z^-)^3\\
&\;\;\;\;\;+c_1Z^++c_2(Z^+)^2+c_3(Z^+)^3+\varepsilon_Y\\
\end{align*}
with $\varepsilon_X,\varepsilon_{Z^-},\varepsilon_Y\sim\mathcal{N}(0,1)$, $
Z^+\sim\mathcal{N}(2,1)$, 
$b_1\sim \text{Unif}(-0.45,0.55)$,
$a_1\sim \text{Unif}(0,1)$, 
$c_1\sim \text{Unif}(0.5,1.5)$ and $
\mu_0,a_2,a_3,b_2,b_3,c_2,c_3\sim \text{Unif}(-0.5,0.5)$. The ground truth expectations are as follows:
\begin{align*}
E[Y\mid X=x]&=\mu_0+a_1x+a_2x^2+a_3x^3+b_1(1.5x-1)+b_2(x^2+1)+b_3(x^3+3x)\\
&\;\;\;\;\;+c_1(1+0.5x)+c_2(1+2x)+c_3(7.5x-1)\\
E[Y\mid do(X:=x)]&=\mu_0+a_1x+a_2x^2+a_3x^3+b_1(1.5x-1)+b_2(x^2+1)+b_3(x^3+3x)\\
&\;\;\;\;\;+c_1(2)+c_2(5)+c_3(14)\\
\end{align*}

The results visualized in Figures~\ref{empty_ex1}--\ref{empty_ex6}, and Tables~\ref{tab:ex1data}--\ref{tab:ex6data} show that the spread of RR and TSR gets reduced by adding ridge penalty. But we also recognize that adding a ridge penalty can go along with adding bias, which is evident for $n=500$. Especially, for Example 1, the ridge estimation deviates far from the true underlying causal effect, which is even outside $95\%$-area of TSR with ridge penalty. The same applies to RR with a ridge regression penalty for Example 1, which in turn does not include $E[Y\mid X]$, in its confidence interval, even though $E[Y\mid X]$ is the quantity RR aims to estimate. As expected, RR does not recover $E[Y\mid do(X)]$ (since it is misspecified in these settings), which is evident since the underlying causal effect is not covered by the $95\%$ of the RR estimator at least in a wide range of the distribution of $X$ in $\mathcal{D}$. This is exactly what we would expect to see when $E[Y\mid do(X)]$ and $ E[Y\mid X]$ differ significantly. Of course, the estimates vary stronger the further the particular values of $X$ are from the support of $X$ in $\mathcal{S}$. But again, the variation diminishes with increasing sample size. Just to notice, particularly for Example~6, it becomes visible how unreliable the naive estimation is. It suggests a quadratic relationship instead of the linear ground truth.

\begin{table}[h!]
 \caption{\textbf{Example 1}: mean(sd) of MSE over $\mathcal{S}$ and $\mathcal{D}$.}\vspace{0.4cm}
    \centering
    \tiny
    \begin{tabular}{lcccccc}
    &\multicolumn{6}{c}{$\mathcal{S}$}\\
    \cmidrule(rl){2-7}
    &\multicolumn{3}{c}{$\mathcal{S}\subset\mathcal{D}$}&\multicolumn{3}{c}{$\mathcal{S}\cap\mathcal{D}=\emptyset$}\\
&$n=500$&$n=1000$&$n=5000$&$n=500$&$n=1000$&$n=5000$\\
\cmidrule(rl){2-4} \cmidrule(rl){5-7}
         naive& 35.64 (41.54) & 31.15 (39.76)& 37.30 (37.76)&
         35.67 (44.17)& 31.41 (39.75)& 37.24 (37.70)\\
         RR& 36.62 (42.61)& 31.64 (40.51)& 38.03 (38.57)&
         36.54 (43.83)& 31.27 (38.91)& 37.76 (38.30)\\
         RR (ridge)& 36.75 (41.95)& 31.96 (40.86)& 38.26 (38.45)&
         36.69 (43.75)& 32.13 (39.32)& 37.86 (38.14)\\
         TSR& 0.12 (0.17)& 0.10 (0.25)& 0.01 (0.03)&
         0.17 (0.25)& 0.09 (0.13)& 0.02 (0.02)\\
         TSR (ridge)& 1.00 (1.62)& 0.61 (0.65)& 0.46 (0.68)&
         0.96 (1.48)& 0.91 (1.51)& 0.46 (0.65)\\
    \end{tabular}

\vspace{0.4cm}

    \begin{tabular}{lcccccc}
    &\multicolumn{6}{c}{$\mathcal{D}$}\\
    \cmidrule(rl){2-7}
    &\multicolumn{3}{c}{$\mathcal{S}\subset\mathcal{D}$}&\multicolumn{3}{c}{$\mathcal{S}\cap\mathcal{D}=\emptyset$}\\
&$n=500$&$n=1000$&$n=5000$&$n=500$&$n=1000$&$n=5000$\\
\cmidrule(rl){2-4} \cmidrule(rl){5-7}
         naive& 53.51 (80.57)& 34.66 (48.07)& 30.13 (30.03)&
         47.79 (63.77)& 34.74 (47.45)& 30.30 (30.33)\\
         RR& 31.65 (35.44)& 27.03 (33.77)&32.59 (32.57)&
         31.77 (36.69)& 26.76 (32.76)& 32.47 (32.50)\\
         RR (ridge)& 91.82 (82.69)& 77.97 (71.21)& 68.97 (62.87)&
         90.08 (80.88)& 74.12 (61.28)& 69.14 (63.63)\\
         TSR& 0.93 (1.14)& 0.47 (0.60)& 0.07 (0.08)&
         0.90 (1.06)& 0.47 (0.57)& 0.07 (0.08)\\
         TSR (ridge)& 71.84 (84.50)& 61.73 (74.93)& 47.66 (65.08)&
         69.91 (81.75)& 57.16 (61.17)& 47.85 (65.58)\\
    \end{tabular}
    \vspace{0.2cm}
 \label{tab:ex1data}
 \end{table}

\vspace{2cm}

\begin{table}[h!]
 \caption{\textbf{Example 2}: mean(sd) of MSE over $\mathcal{S}$ and $\mathcal{D}$.}\vspace{0.4cm}
    \centering
    \tiny
    \begin{tabular}{lcccccc}
     &\multicolumn{6}{c}{$\mathcal{S}$}\\
    \cmidrule(rl){2-7}
    &\multicolumn{3}{c}{$\mathcal{S}\subset\mathcal{D}$}&\multicolumn{3}{c}{$\mathcal{S}\cap\mathcal{D}=\emptyset$}\\
&$n=500$&$n=1000$&$n=5000$&$n=500$&$n=1000$&$n=5000$\\
\cmidrule(rl){2-4} \cmidrule(rl){5-7}
         naive& 44.62 (39.08)& 59.09 (55.41)& 49.16 (40.61)&
         43.88 (40.26)& 60.25 (50.04)& 48.07 (38.91)\\
         RR& 62.85 (55.63)& 82.69 (62.06)& 71.72 (58.78)&
         61.26 (54.98)& 85.49 (66.62)& 70.74 (59.00)\\
         RR (ridge)& 62.39 (55.13)& 82.44 (61.84)& 71.59 (58.67)&
         60.81 (54.56)& 85.20 (66.41)& 70.63 (58.91)\\
         TSR& 0.58 (0.77)& 0.35 (0.69)& 0.05 (0.06)&
         0.55 (0.89)& 0.39 (0.76)& 0.04 (0.06)\\
         TSR (ridge)& 0.58 (0.77)& 0.35 (0.70)& 0.05 (0.0.06)&
         0.56 (0.88)& 0.40 (0.76)& 0.04 (0.06)\\
    \end{tabular}

\vspace{0.4cm}

    \centering
    \begin{tabular}{lcccccc}
     &\multicolumn{6}{c}{$\mathcal{D}$}\\
    \cmidrule(rl){2-7}
    &\multicolumn{3}{c}{$\mathcal{S}\subset\mathcal{D}$}&\multicolumn{3}{c}{$\mathcal{S}\cap\mathcal{D}=\emptyset$}\\
&$n=500$&$n=1000$&$n=5000$&$n=500$&$n=1000$&$n=5000$\\
\cmidrule(rl){2-4} \cmidrule(rl){5-7}
         naive& 192.91 (277.92)& 232.67 (314.83)& 178.90 (231.55)&
         218.91 (362.72)& 205.79 (257.19)& 158.34 (167.97)\\
         RR& 180.61 (190.55)& 218.99 (181.33)& 183.83 (172.65)& 
         177.40 (180.90)& 228.01 (190.31)& 185.24 (178.39)\\
         RR (ridge)& 180.72 (190.17)& 219.30 (181.57)& 183.81 (172.79)&
         176.75 (179.73)& 228.17 (190.81)& 185.29 (178.51)\\
         TSR& 2.43 (3.86)& 0.81 (1.43)& 0.12 (0.11)&
         1.78 (2.12)& 0.99 (1.17)& 0.12 (0.11)\\
         TSR (ridge)& 2.43 (4.01)& 0.82 (1.37)& 0.12 (0.12)&
         1.87 (2.45)& 1.08 (1.23)& 0.12 (0.12)\\
    \end{tabular}
    \vspace{0.2cm}
 \label{tab:ex2data}
 \end{table}

\begin{table}[h!]
\caption{\textbf{Example 3}: mean(sd) of MSE over $\mathcal{S}$ and $\mathcal{D}$.}\vspace{0.4cm}
    \centering
    \tiny
    \begin{tabular}{lcccccc}
     &\multicolumn{6}{c}{$\mathcal{S}$}\\
    \cmidrule(rl){2-7}
    &\multicolumn{3}{c}{$\mathcal{S}\subset\mathcal{D}$}&\multicolumn{3}{c}{$\mathcal{S}\cap\mathcal{D}=\emptyset$}\\
&$n=500$&$n=1000$&$n=5000$&$n=500$&$n=1000$&$n=5000$\\
\cmidrule(rl){2-4} \cmidrule(rl){5-7}
         naive& 0.89 (0.84)& 0.85 (0.90)& 0.75 (0.79)&
         0.92 (0.84)& 0.85 (0.89)& 0.75 (0.79)\\
         RR& 0.75 (0.82)& 0.46 (0.44)& 0.29 (0.25)&
         0.79 (0.83)& 0.50 (0.50)& 0.29 (0.25)\\
         RR (ridge)& 0.58 (0.55)& 0.52 (0.40)& 0.39 (0.30)&
         0.62 (0.59)& 0.54 (0.45)& 0.38 (0.29)\\
         TSR& 0.37 (0.45)& 0.15 (0.20)& 0.03 (0.05)&
         0.40 (0.50)& 0.13 (0.18)& 0.03 (0.05)\\
         TSR (ridge)& 0.22 (0.25)& 0.19 (0.21)& 0.10 (0.14)&
         0.23 (0.25)& 0.17 (0.20)& 0.10 (0.14)\\
    \end{tabular}

\vspace{0.4cm}

    \centering
    \begin{tabular}{lcccccc}
     &\multicolumn{6}{c}{$\mathcal{D}$}\\
    \cmidrule(rl){2-7}
    &\multicolumn{3}{c}{$\mathcal{S}\subset\mathcal{D}$}&\multicolumn{3}{c}{$\mathcal{S}\cap\mathcal{D}=\emptyset$}\\
&$n=500$&$n=1000$&$n=5000$&$n=500$&$n=1000$&$n=5000$\\
\cmidrule(rl){2-4} \cmidrule(rl){5-7}
         naive& 7.91 (13.09)& 5.53 (10.13)& 1.25 (1.21)&
         7.58 (10.70)& 5.67 (11.53)& 1.23 (1.19)\\
         RR& 0.97 (0.81)& 0.66 (0.50)& 0.42 (0.33)&
         1.00 (0.86)& 0.67 (0.51)& 0.42 (0.33)\\
         RR (ridge)& 0.87 (0.70)& 0.77 (0.59)& 0.54 (0.33)&
         0.91 (0.74)& 0.76 (0.57)& 0.54 (0.32)\\
         TSR& 0.49 (0.49)& 0.22 (0.23)& 0.04 (0.05)&
         0.52 (0.52)& 0.21 (0.21)& 0.04 (0.05)\\
         TSR (ridge)& 0.51 (0.59)& 0.38 (0.35)& 0.20 (0.21)&
         0.51 (0.56)& 0.37 (0.37)& 0.20 (0.21)\\
    \end{tabular}
   \vspace{0.2cm} 
 \label{tab:ex3data}
 \end{table}

\begin{table}[h!]
 \caption{\textbf{Example 4}: mean(sd) of MSE over $\mathcal{S}$ and $\mathcal{D}$.}\vspace{0.4cm}
    \centering
    \tiny
   \begin{tabular}{lcccccc}
     &\multicolumn{6}{c}{$\mathcal{S}$}\\
    \cmidrule(rl){2-7}
    &\multicolumn{3}{c}{$\mathcal{S}\subset\mathcal{D}$}&\multicolumn{3}{c}{$\mathcal{S}\cap\mathcal{D}=\emptyset$}\\
&$n=500$&$n=1000$&$n=5000$&$n=500$&$n=1000$&$n=5000$\\
\cmidrule(rl){2-4} \cmidrule(rl){5-7}
         naive& 13.65 (16.52)& 14.64 (16.63)& 11.75 (14.17)&
         1.98 (4.68)& 1.24 (1.29)& 0.59 (0.58)\\
         RR& 11.90 (14.16)& 13.31 (15.08)& 11.69 (14.03)&
         0.80 (1.75)& 0.42 (0.46)& 0.22 (0.26)\\
         RR (ridge)& 11.75 (14.00)& 13.29 (15.07)& 11.69 (14.03)&
         0.58 (0.57)& 0.41 (0.42)& 0.23 (0.26)\\
         TSR& 13.01 (15.54)& 14.45 (16.88)& 13.10 (15.94)&
         0.47 (1.52)& 0.11 (0.19)& 0.01 (0.01)\\
         TSR (ridge)& 12.75 (15.23)& 14.42 (16.84)& 13.10 (15.93)&
         0.26 (0.34)& 0.11 (0.17)& 0.02 (0.01)\\
    \end{tabular}

\vspace{0.4cm}

    \centering
   \begin{tabular}{lcccccc}
     &\multicolumn{6}{c}{$\mathcal{D}$}\\
    \cmidrule(rl){2-7}
    &\multicolumn{3}{c}{$\mathcal{S}\subset\mathcal{D}$}&\multicolumn{3}{c}{$\mathcal{S}\cap\mathcal{D}=\emptyset$}\\
&$n=500$&$n=1000$&$n=5000$&$n=500$&$n=1000$&$n=5000$\\
\cmidrule(rl){2-4} \cmidrule(rl){5-7}
         naive& 17.74 (19.86)& 17.16 (18.87)& 13.31 (15.99)&
         6.23 (27.29)& 2.61 (4.70)& 0.52 (0.47)\\
         RR& 14.94 (18.84)& 14.84 (16.86)& 13.44 (16.17)&
         2.32 (11.34)& 0.73 (1.20)& 0.19 (0.20)\\
         RR (ridge)& 13.87 (15.50)& 15.08 (16.94)& 13.45 (16.15)&
         1.24 (1.69)& 0.74 (1.06)& 0.20 (0.20)\\
         TSR& 14.76 (19.41)& 14.57 (16.79)& 13.19 (16.05)&
         2.07 (10.69)& 0.45 (0.86)& 0.04 (0.05)\\
         TSR (ridge)& 13.61 (15.34)& 14.76 (16.86)& 13.20 (16.04)&
         1.03 (1.55)& 0.48 (0.75)& 0.05 (0.07)\\
    \end{tabular}
    \vspace{0.2cm}
 \label{tab:ex4data}
 \end{table}

\begin{table}[h!]
    \caption{\textbf{Example 5}: mean(sd) of MSE over $\mathcal{S}$ and $\mathcal{D}$.}\vspace{0.4cm}
    \centering
    \tiny
    \begin{tabular}{lcccccc}
     &\multicolumn{6}{c}{$\mathcal{S}$}\\
    \cmidrule(rl){2-7}
    &\multicolumn{3}{c}{$\mathcal{S}\subset\mathcal{D}$}&\multicolumn{3}{c}{$\mathcal{S}\cap\mathcal{D}=\emptyset$}\\
&$n=500$&$n=1000$&$n=5000$&$n=500$&$n=1000$&$n=5000$\\
\cmidrule(rl){2-4} \cmidrule(rl){5-7}
         naive& 29.25 (54.11)& 20.20 (20.62)& 15.53 (12.21)&
         26.44 (44.31)& 18.14 (16.38)& 16.29 (14.99)\\
         RR& 8.95 (8.36)& 7.35 (7.38)& 4.53 (3.95)&
         7.76 (7.95)& 6.38 (4.63)& 4.38 (3.76)\\
         RR (ridge)& 8.94 (8.34)& 7.32 (7.34)& 4.52 (3.94)&
         7.58 (7.13)& 6.38 (4.61)& 4.38 (3.76)\\
         TSR& 3.95 (6.97)& 1.86 (3.90)& 0.29 (0.42)&
         2.55 (5.46)& 1.07 (1.63)& 0.25 (0.31)\\
         TSR (ridge)& 3.89 (6.74)& 1.83 (3.86)& 0.29 (0.42)&
         2.40 (4.60)& 1.05 (1.60)& 0.25 (0.31)\\
    \end{tabular}

\vspace{0.4cm}

    \centering
   \begin{tabular}{lcccccc}
     &\multicolumn{6}{c}{$\mathcal{D}$}\\
    \cmidrule(rl){2-7}
    &\multicolumn{3}{c}{$\mathcal{S}\subset\mathcal{D}$}&\multicolumn{3}{c}{$\mathcal{S}\cap\mathcal{D}=\emptyset$}\\
&$n=500$&$n=1000$&$n=5000$&$n=500$&$n=1000$&$n=5000$\\
\cmidrule(rl){2-4} \cmidrule(rl){5-7}
         naive& 645.27 (846.81)& 458.25 (683.80)& 285.40 (349.38)&
         685.67 (957.58)& 447.97 (651.41)& 282.29 (344.48)\\
         RR& 41.68 (64.16)& 23.23 (26.25)& 13.13 (12.25)&
         41.07 (66.25)& 19.71 (20.32)& 12.89 (11.29)\\
         RR (ridge)& 42.44 (60.81)& 23.07 (25.11)& 13.07 (12.11)&
         40.48 (59.89)& 19.28 (20.05)& 12.79 (11.08)\\
         TSR& 24.91 (40.01)& 8.27 (12.43)& 1.63 (2.22)&
         25.68 (51.84)& 7.14 (11.49)& 1.43 (1.89)\\
         TSR (ridge)& 25.91 (40.30)& 7.99 (11.58)& 1.65 (2.26)&
         25.38 (46.43)& 6.81 (11.30)& 1.41 (1.91)\\
    \end{tabular}
    \vspace{0.2cm}
 \label{tab:ex5data}
 \end{table}

\vspace{2cm}
\begin{table}[h!]
\caption{\textbf{Example 6}: mean(sd) of MSE over $\mathcal{S}$ and $\mathcal{D}$.}\vspace{0.4cm}
    \centering
    \tiny
   \begin{tabular}{lcccccc}
     &\multicolumn{6}{c}{$\mathcal{S}$}\\
    \cmidrule(rl){2-7}
    &\multicolumn{3}{c}{$\mathcal{S}\subset\mathcal{D}$}&\multicolumn{3}{c}{$\mathcal{S}\cap\mathcal{D}=\emptyset$}\\
&$n=500$&$n=1000$&$n=5000$&$n=500$&$n=1000$&$n=5000$\\
\cmidrule(rl){2-4} \cmidrule(rl){5-7}
         naive& 13.16 (10.06)& 13.11 (8.62)& 10.78 (8.52)&
         12.85 (9.30)& 13.44 (8.76)& 10.63 (8.31)\\
         RR& 6.35 (6.38)& 6.32 (6.11)& 4.50 (4.81)&
         6.61 (6.38)& 6.33 (6.09)& 4.55 (4.88)\\
         RR (ridge)& 6.33 (6.37)& 6.31 (6.11)& 4.49 (4.81)&
         6.59 (6.37)& 6.32 (6.08)& 4.55 (4.88)\\
         TSR& 0.46 (0.51)& 0.29 (0.35)& 0.08 (0.16)&
         0.62 (1.07)& 0.25 (0.21)& 0.06 (0.06)\\
         TSR (ridge)& 0.44 (0.47)& 0.28 (0.35)& 0.08 (0.15)&
         0.60 (1.03)& 0.24 (0.21)& 0.06 (0.06)\\
    \end{tabular}

\vspace{0.4cm}

    \centering
    \begin{tabular}{lcccccc}
     &\multicolumn{6}{c}{$\mathcal{D}$}\\
    \cmidrule(rl){2-7}
    &\multicolumn{3}{c}{$\mathcal{S}\subset\mathcal{D}$}&\multicolumn{3}{c}{$\mathcal{S}\cap\mathcal{D}=\emptyset$}\\
&$n=500$&$n=1000$&$n=5000$&$n=500$&$n=1000$&$n=5000$\\
\cmidrule(rl){2-4} \cmidrule(rl){5-7}
         naive& 261.50 (304.63)& 252.64 (258.01)& 216.26 (231.02)&
         245.66 (259.98)& 264.85 (267.89)& 214.43 (228.39)\\
         RR& 17.59 (18.28)& 14.23 (14.10)& 9.68 (10.10)&
         17.02 (17.84)& 14.59 (15.44)& 9.56 (10.67)\\
         RR (ridge)& 17.36 (19.08)& 14.01 (13.76)& 9.70 (10.14)&
         16.79 (18.68)& 14.36 (14.89)& 9.61 (10.66)\\
         TSR& 4.12 (5.70)& 2.50 (3.77)& 0.42 (0.55)&
         4.05 (5.40)& 2.17 (3.19)& 0.037 (0.60)\\
         TSR (ridge)& 3.88 (5.87)& 2.29 (3.41)& 0.44 (0.52)&
         4.24 (5.68)& 2.06 (3.06)& 0.40 (0.64)\\
    \end{tabular}
    \vspace{0.2cm}
 \label{tab:ex6data}
 \end{table}

\subsection{Supplementary Information to \Cref{sec:expfigure1d}}
\label{sec:appexpfigure1d}

\begin{figure}[t!]
   \begin{minipage}[b]{.5\linewidth}
       \includegraphics[width=\linewidth]{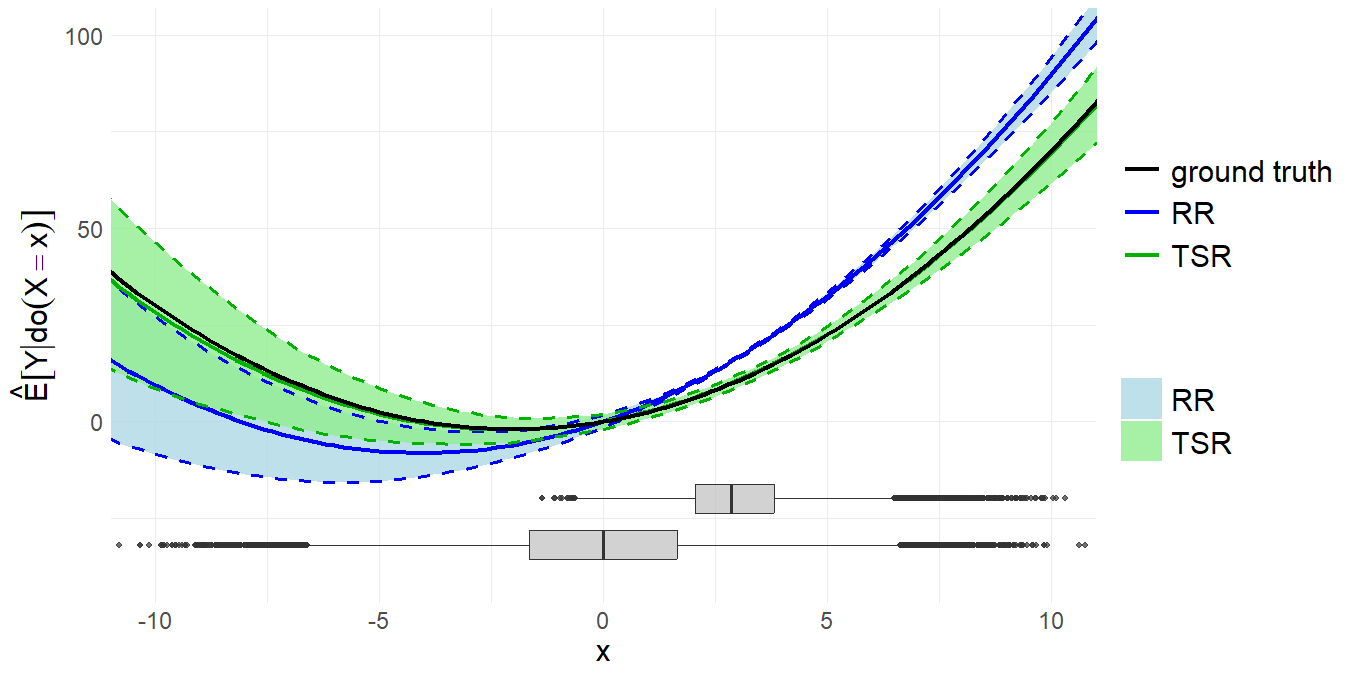}
   \end{minipage}
   \hspace{.01\linewidth}
   \begin{minipage}[b]{.5\linewidth} 
      \includegraphics[width=\linewidth]{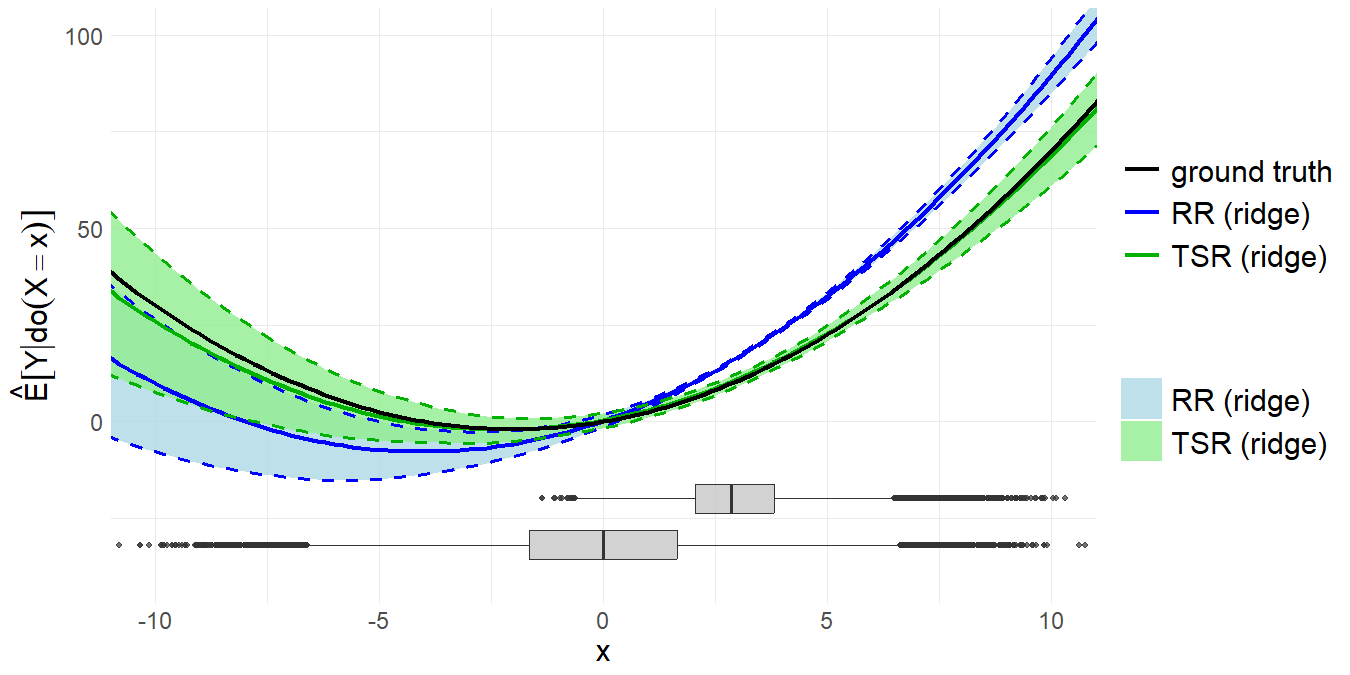}
   \end{minipage}
\caption{We show in black the ground truth value of $E[Y\mid do(X)]$ and colored the mean estimates and $95\%$-areas over all simulations for RR and TSR via OLS (left) and ridge regression (right) for $n=2000$ (for $\mathcal{S}\cap\mathcal{D}=\emptyset$). We see that TSR recovers the ground truth whereas RR does not. The boxplots on the x-axis show the distribution of $\Xv$ in $\mathcal{S}$ (upper) and $\mathcal{D}$ (lower).}
\label{fig:integraln2000}
\end{figure}

In comparison to the example with $n=500$ in the main part of the paper (\Cref{sec:expfigure1d}), Figure~\ref{fig:integraln2000} shows the results for a larger sample size ($n=2000$). We see that when increasing the sample size $n$, the estimations spread less. Further, adding a ridge penalty affects the estimation less than for a smaller sample size. 

\subsection{Supplementary Information to \Cref{sec:exprobust}}
\label{sec:appexprobust}

We generate the data according to the SCM
\begin{align*}
S&:=\mathbf{1}_{\{X+Z<-5\}}\\
Y&:=0.2X^2+5Z+\varepsilon_Y
\end{align*}
with $\varepsilon_Y\sim\mathcal{N}(0,1)$, $Z\sim\mathcal{N}(-2,1)$ and $
X\sim\mathcal{N}(-4,5)$. The ground truth expectations are as follows: 
\begin{align*}
    E[Y\mid X=x]&=0.2x^2-2+2x\\
    E[Y\mid do(X:=x)]&=0.2x^2-10
\end{align*}

\begin{figure}[t!]
\centering
\hspace{-1cm}without $U$\hspace{5cm}$U\rightarrow Y$\\
   \begin{minipage}[b]{.21\linewidth}
      \includegraphics[width=0.94\linewidth]{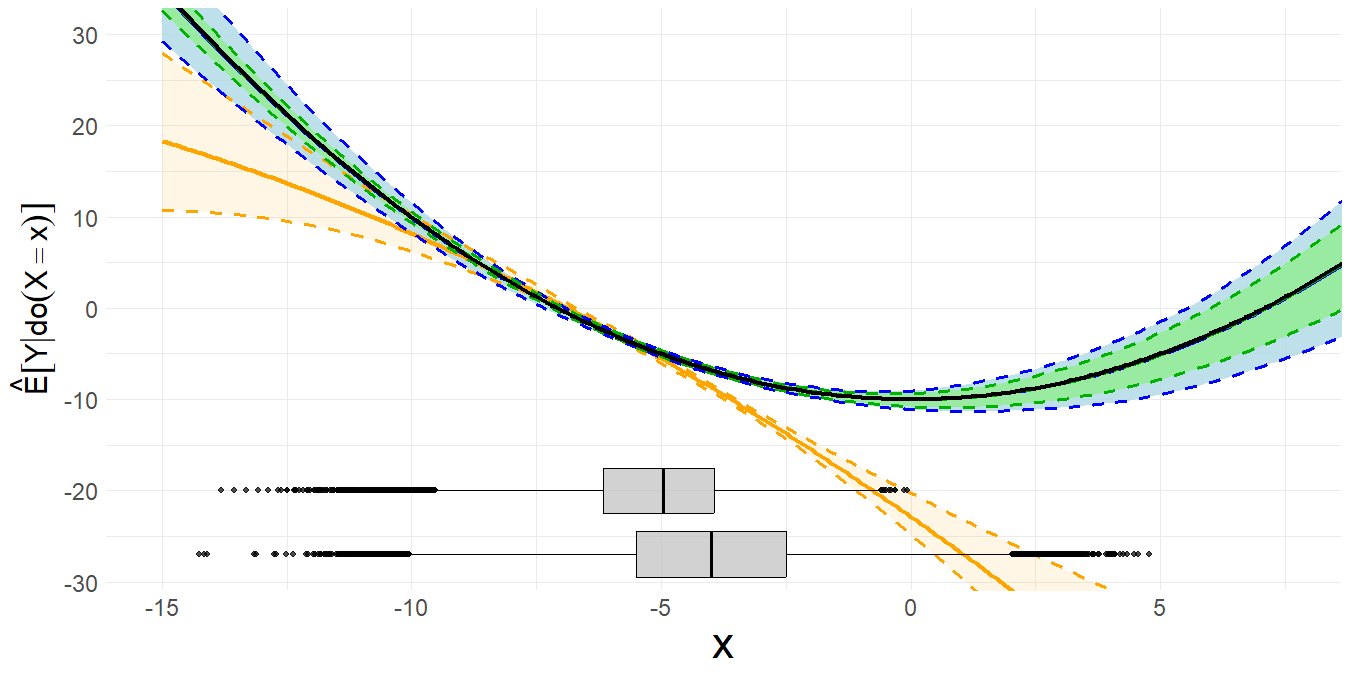}
    \end{minipage}
   \begin{minipage}[b]{.21\linewidth} 
\includegraphics[width=0.94\linewidth]{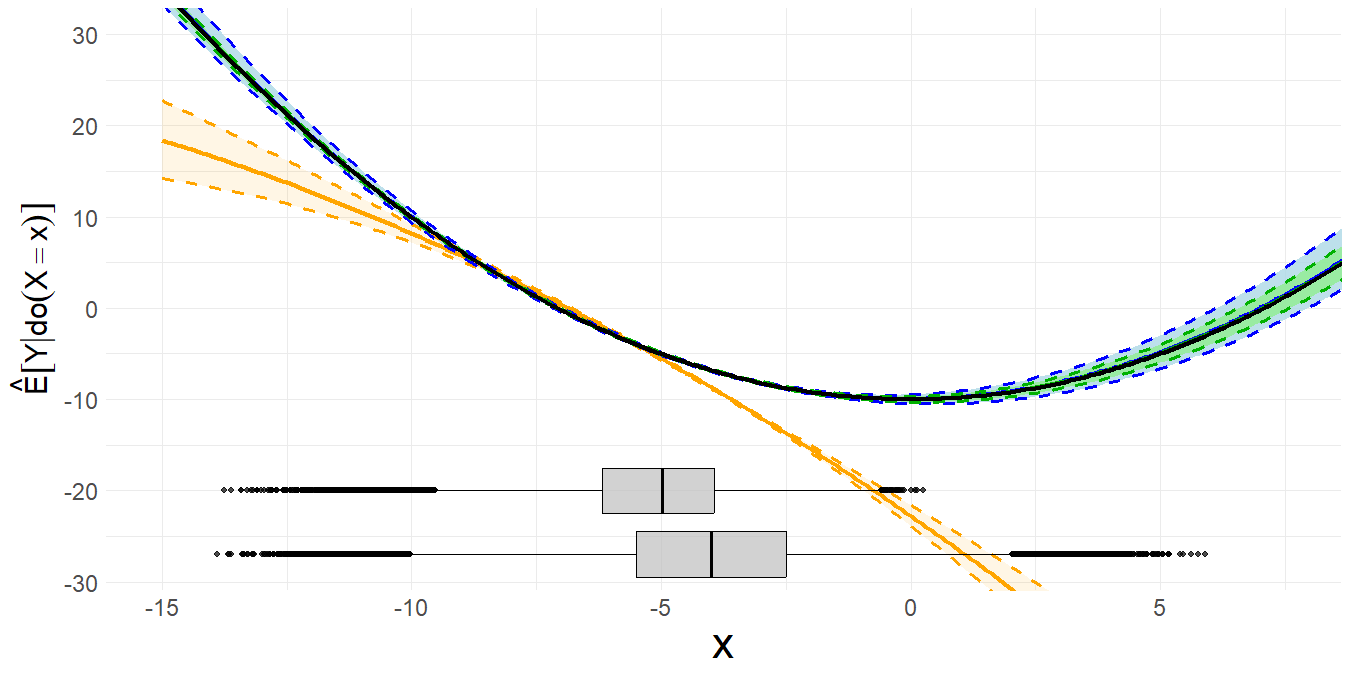}
    \end{minipage}
   \begin{minipage}[b]{.21\linewidth}
\includegraphics[width=0.94\linewidth]{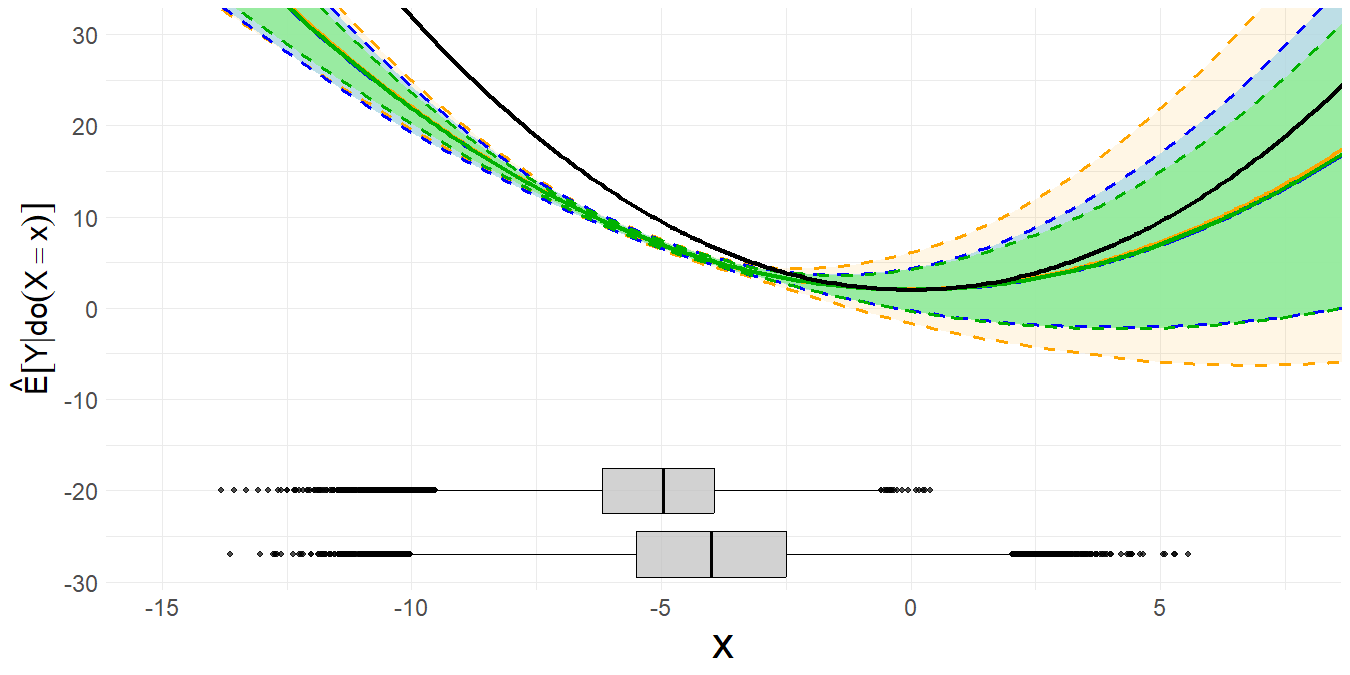}
   \end{minipage}
   \begin{minipage}[b]{.21\linewidth} 
\includegraphics[width=0.94\linewidth]{Fig6a_17d}
   \end{minipage}
   \begin{minipage}[b]{.07\linewidth} 
\includegraphics[scale=0.11]{Fig6d_17e_17j_17o.png}
   \end{minipage}\\

$S\leftarrow U$ (type 1) \hspace{3cm} $S\leftarrow U\rightarrow Y$ (type 1)\\
   \begin{minipage}[b]{.21\linewidth}
\includegraphics[width=0.94\linewidth]{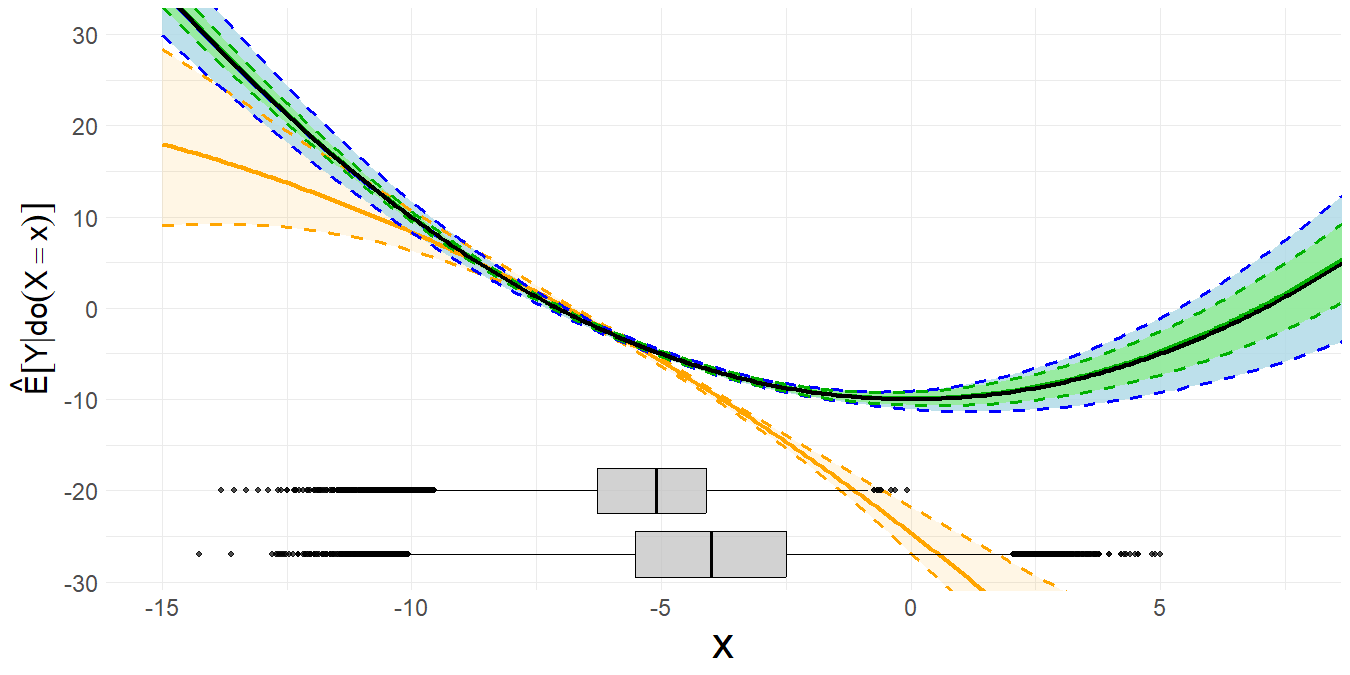}
   \end{minipage}
   \begin{minipage}[b]{.21\linewidth} 
\includegraphics[width=0.94\linewidth]{Fig6b_17g}
   \end{minipage}
   \begin{minipage}[b]{.21\linewidth}
\includegraphics[width=0.94\linewidth]{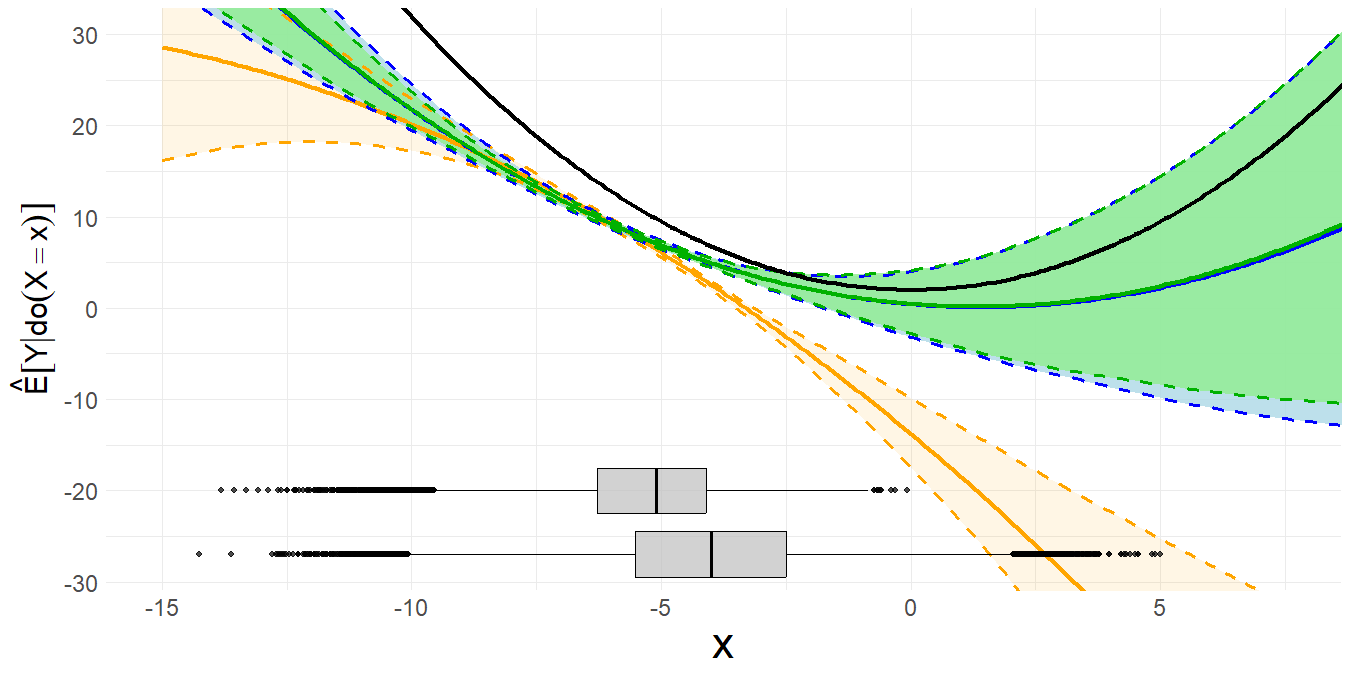}
   \end{minipage}
   \begin{minipage}[b]{.21\linewidth} 
\includegraphics[width=0.94\linewidth]{Fig6c_17i}
   \end{minipage}
   \begin{minipage}[b]{.07\linewidth}
\includegraphics[scale=0.11]{Fig6d_17e_17j_17o.png}
   \end{minipage}\\

$S\leftarrow U$ (type 2) \hspace{3cm} $S\leftarrow U\rightarrow Y$ (type 2)\\
   \begin{minipage}[b]{.21\linewidth}
\includegraphics[width=0.94\linewidth]{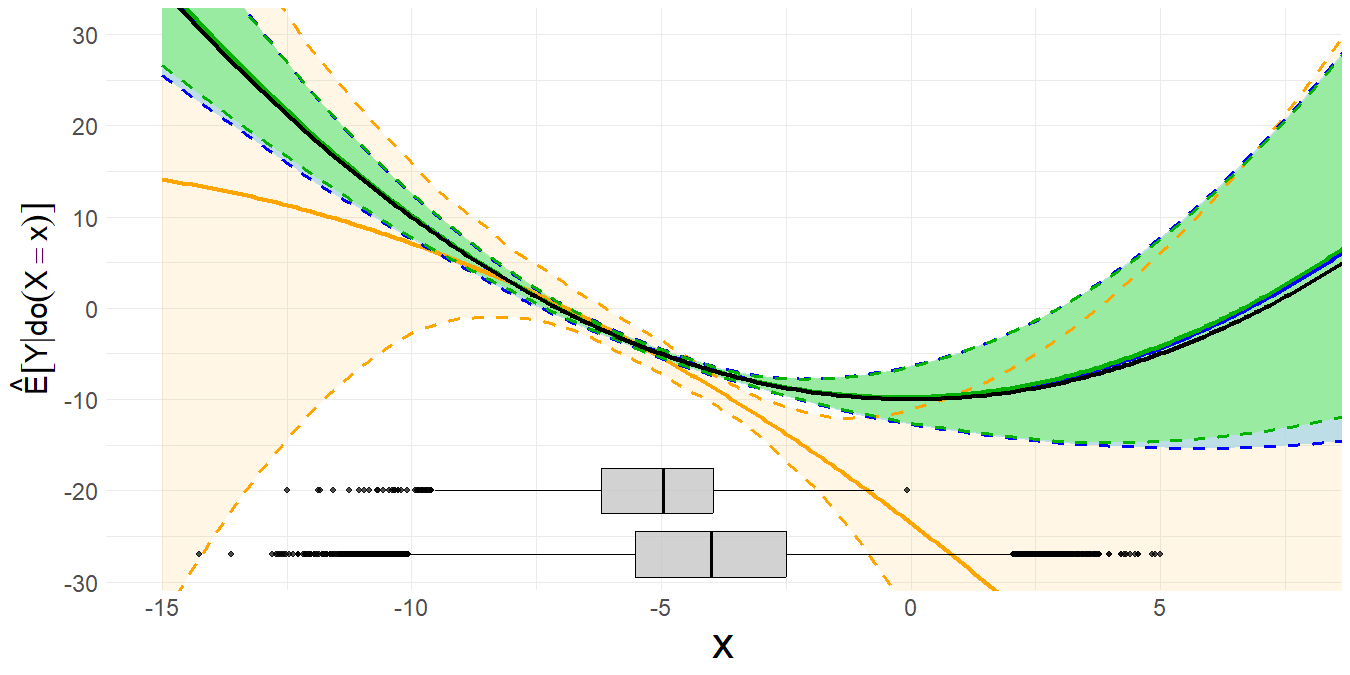}
      \caption*{$n=1000$}
   \end{minipage}
   \begin{minipage}[b]{.21\linewidth} 
\includegraphics[width=0.94\linewidth]{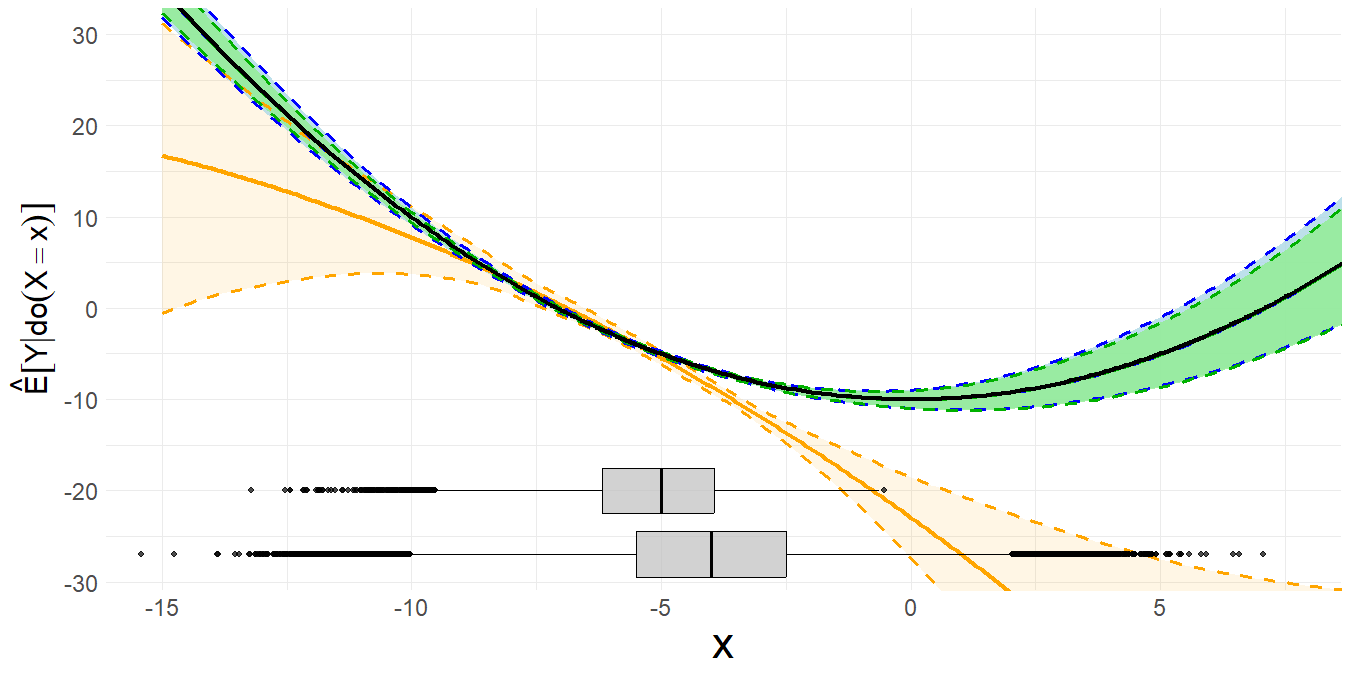}
      \caption*{$n=5000$}
   \end{minipage}
   \begin{minipage}[b]{.21\linewidth}
\includegraphics[width=0.94\linewidth]{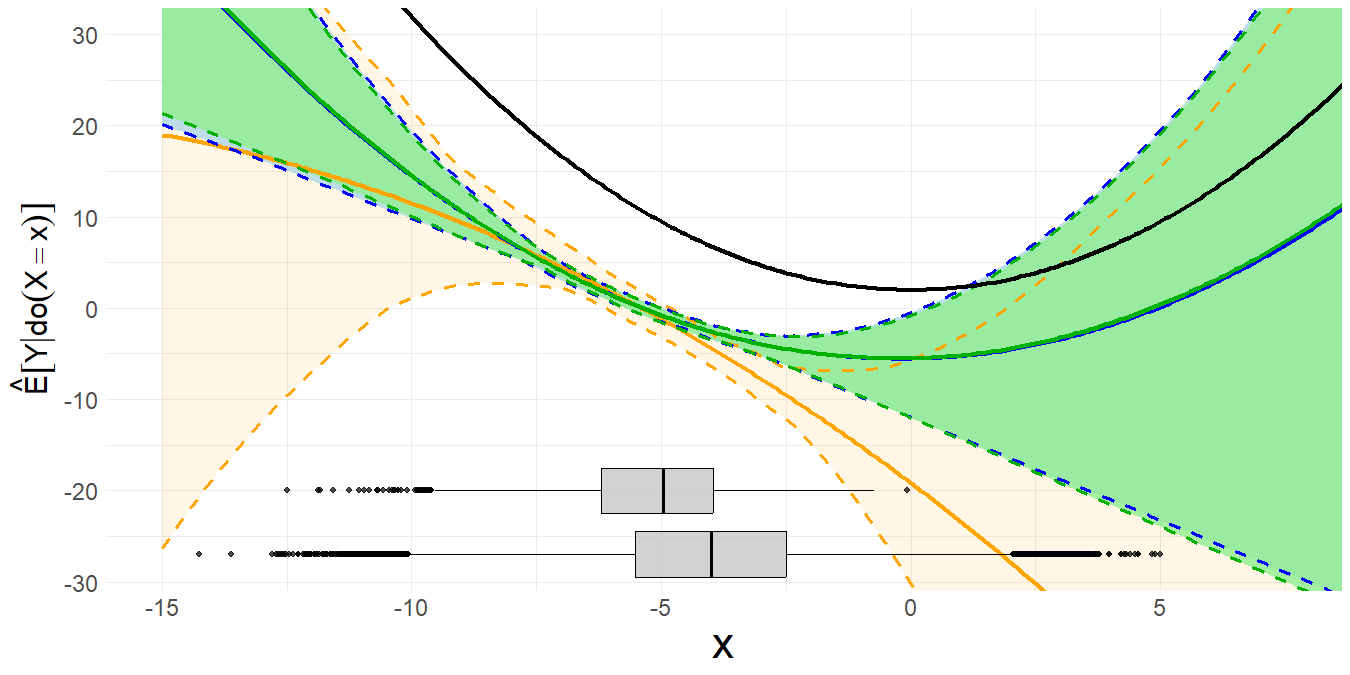}
      \caption*{$n=1000$}
   \end{minipage}
   \begin{minipage}[b]{.21\linewidth} 
\includegraphics[width=0.94\linewidth]{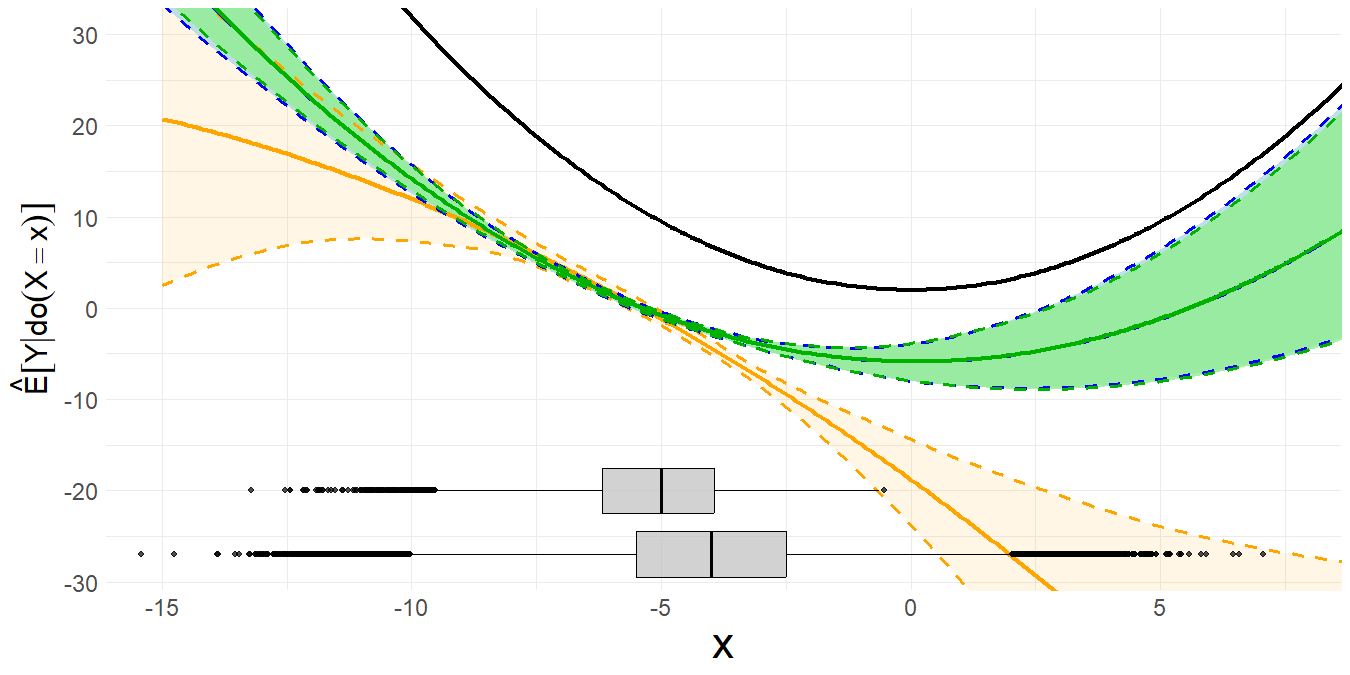}
      \caption*{$n=5000$}
   \end{minipage}
   \begin{minipage}[b]{.07\linewidth}
   \includegraphics[scale=0.11]{Fig6d_17e_17j_17o.png}
   \vspace{0.05cm}\\
   \end{minipage}
 \caption{We show in black the ground truth value of $E[Y|do(X=x)]$, and colored the mean estimates and $95\%$-areas over all simulations for naive, RR, and TSR in absence of $U$, introducing $U\rightarrow Y$, $S\leftarrow U$ or $S\leftarrow U\rightarrow Y$ for $n\in\{1000,5000\}$ (for $\mathcal{S}\cap\mathcal{D}=\emptyset$). The boxplots on the x-axis show the distribution of $\Xv$ in $\mathcal{S}$ (upper) and $\mathcal{D}$ (lower).}   
\label{fig:rob_2}
\end{figure}

\subsection{Supplementary Information to \Cref{sec:exprealworld}}
\label{sec:appexprealworld}

Our simulations are based on data from the Infant Health and Development Program (IHDP), a randomized experiment with sample size $n= 985$ exploring the effect of home visits on cognitive test scores for infants. We use the original data,\!\footnote{The data is available at \href{https://www.tandfonline.com/doi/suppl/10.1198/jcgs.2010.08162}{https://www.tandfonline.com/doi/suppl/10.1198/jcgs.2010.08162}} but generate the target and the treatment based on simulations inspired by \citet{hill:11:IHDP}.

\citet{hill:11:IHDP} models the potential outcomes separately as normal distributions, where the mean is a linear combination of 25 covariates, which include information on both the infant (for instance the birthweight, gender and neonatal health) and the mother (for example her age at birth, education and indicators for smoking boozing or working during pregnancy). The coefficients are randomly sampled values $(0, 1, 2, 3, 4)$ with probabilities $(0.5, 0.2, 0.15, 0.1, 0.05)$. In contrast to \citet{hill:11:IHDP}, we generate a continuous instead of a binary treatment and accordingly use only one model for the target variable instead of two seperate models for the potential outcomes. We generate the outcome $Y$ following a normal distribution, where the mean is a linear combination of the 25 covariates (serve as our proxies) and the continuous treatment $X$ plus an intercept and variance equal to $1$, where we use the non-binary covariates in standardized form. The---in total 27---coefficients are randomly sampled values $(0, 1, 2, 3, 4)$ with probabilities $(0.5, 0.2, 0.15, 0.1, 0.05)$. The treatment $X$ is uniformly distributed on $[-1,1]$. Specifically, the following variables were used:\\

\textbf{(Quasi-) continuous covariates}
\begin{itemize}
\item child: birthweight (grams)
\item child: head circumference (cm) at birth
\item child: weeks pre-term that the child was born
\item child: birth order
\item child: neo-natal health index
\item mother: age when child was born
\end{itemize}
\textbf{Binary covariates}
\begin{itemize}
\item child: gender 
\item child: twin
\item mother: married when child was born
\item mother: education level when child was born\\
    (did not graduate from high school, graduated from high school, attended some college but did not graduate)
\item mother: cigarettes during pregnancy
\item child: first born (recoded from $\{1,2\}$ to $\{0,1\})$
\item mother: alcohol during pregnancy
\item mother: drugs during pregnancy
\item mother: work during pregnancy
\item mother: received prenatal care
\item site in
which the family resided at the start of the intervention\\
(ark, ein, har, mia, pen, tex, was)
\end{itemize}

 Then, we insert a selection process $S=\mathbf{1}_{\{X+Z_3>0.7\}}$ based on the treatment and proxy variable $Z_3$, which denotes the number of weeks pre-term that the child was born, constructing $\mathcal{S}\subset\mathcal{D}$. By proceeding in this way, we construct the setting shown in \Cref{fig:vierDAGs} (a). We fit linear regressions by OLS and instantiated with a ridge penalty analogously to the previous sections and then evaluate the coefficients for the treatment $X$. \Cref{fig:real-world} shows the results from $1000$ simulation runs confirming that RR and TSR recover the causal effect under selection bias (satisfying PMAR) when confounding is absent, whereas the naive estimator is biased. 

 Last, we insert a confounder $C\sim \mathcal{N}(2,2)$
 affecting treatment and target, which serves also as a proxy in TSR and RR. Previously, we had $E[Y\mid X,Z]=q(X,Z)$. We construct the confounded Variables $X_{\text{conf}}$ and $Y_{\text{conf}}$ as $X_{\text{conf}}:=X+C$ and $E[Y_{conf}|X_{conf},Z,C]:=q(X_{conf},Z)+C$ such that the mean of the target is a linear combination of the 25 covariates (serve as our proxies), the continuous treatment $X$ plus the confounder $C$. By adding the confounder, we obtain the setting shown in \Cref{fig:expvariousDAGs} (b), defining $Z_A^+:=Z$ and $Z_B^+:=C$. The results from 1000 simulation runs show that TSR recovers the causal effect, whereas RR and the naive estimator are biased, when confounding is introduced. Those results complement the fully synthetic data experiments that we conducted. As in the previous sections, we also add a ridge penalty for the regressions for RR and TSR, obtaining similar results.

\end{appendices}

\end{document}